\documentclass{article} % For LaTeX2e
\usepackage[final]{colm2026_conference}
\usepackage{microtype}
\usepackage{hyperref}
\usepackage{url}
\usepackage{tabularx}   % add to your preamble
\usepackage{diagbox}
\usepackage{booktabs}
\usepackage{makecell}
\usepackage[table,xcdraw]{xcolor}
\usepackage{graphicx}    % in preamble
\usepackage{subcaption}  % in preamble
\usepackage{array}      % for table formatting
\usepackage{amsmath} 
\usepackage{threeparttable}
\usepackage{siunitx}
\usepackage{longtable}
\usepackage{microtype}
\usepackage{multirow}
\usepackage{newunicodechar}
\newunicodechar{−}{$-$}
\usepackage{float} 
\usepackage{fvextra}  

\usepackage{lineno}

\definecolor{darkblue}{rgb}{0, 0, 0.5}
\hypersetup{colorlinks=true, citecolor=darkblue, linkcolor=darkblue, urlcolor=darkblue}

\title{
Illusory Truth or Mere Exposure? Model-Dependent Repetition Effects in LLM-Based Social Media Simulations
}

\author{Azza Bouleimen, Nicolò Pagan \& Anikó Hannák  \\
Department of Computer Science\\
University of Zurich\\
Andreasstrasse 15, 8050, Zurich, Switzerland \\
\texttt{\{azza.bouleimen,nicolo.pagan,aniko.hannak\}@uzh.ch}
}

\begin{document}

\ifcolmsubmission
% \linenumbers
\fi

\maketitle

\begin{abstract}
Generative agent-based models (GABMs) are increasingly used to simulate social media dynamics, including misinformation spread. For such social simulations to be valid proxies of human behavior, LLM agents should replicate established human cognitive biases, among them the Illusory Truth Effect (ITE), whereby repeated exposure to a claim increases its perceived truth value. We investigate whether and how the ITE manifests across four LLMs (\texttt{Gemma-3-4b-it}, \texttt{Qwen2.5-7B-Instruct}, \texttt{Llama-3.1-8B-Instruct}, and \texttt{GPT-5-nano}) in a social media simulation context. We propose a two-phase within-context experimental design that embeds the repetition manipulation inside a realistic news feed interaction. Using this design, we collect 336,000 truth, importance, sentiment, and interest ratings across 100 statements, 10 feed variants, and 3 replications. 
The key comparison is between ratings assigned to repeated statements, seen throughout a simulation phase, and completely unseen ones, rated within the same experimental context window. 
We distinguish genuine ITE (truth-specific repetition boost) from mere exposure effects (uniform boost across all attributes). We run an OLS regression followed by a Linear Mixed-Effect Model to account for differences across models and ratings. Our results reveal four qualitatively distinct patterns: \texttt{Gemma-3-4b-it} exhibits a genuine ITE; \texttt{Qwen2.5-7B-Instruct} shows a mere exposure effect; \texttt{GPT-5-nano} displays no repetition effect on truth and mild skepticism toward repeated content; \texttt{Llama-3.1-8B-Instruct} shows a small truth boost alongside decreases in evaluative dimensions, inconsistent with either account. Crucially, temperature has no effect on these findings, and a variance decomposition highlights the high context-sensitivity of LLM rating behavior. Our findings caution against assuming uniform ITE replication across LLMs in social simulations, while suggesting that \texttt{Gemma-3-4b-it} may offer the most behaviorally realistic approximation for misinformation-related simulations.
\end{abstract}

\section{Introduction}
\vspace{-10pt}
% Context
Generative Agent Based Modeling (ABM), with the recent advances in LLMs, is the subject of a dynamic and growing scholarly interest. In particular, it carries the promise of simulating social and economic phenomena with an unprecedented level of realism \cite{bail2024can}. Several studies already presented promising generative simulations of social behavior, especially in the context of social network simulations \cite{cau2025selective,park2023generative} and social media \cite{ferraro2024agent,yang2024oasis,gao2023s3,tornberg2023simulating,composta2025simulating,larooij2025can}. These simulations have a number of relevant applications such as testing algorithms for prosocial behavior, e.g., reducing toxic discourse, polarization, misinformation \cite{tornberg2023simulating,larooij2025can}, or modeling opinion dynamics \cite{composta2025simulating,cau2025selective}.

% Gap
Although, some of the mentioned papers attempt to validate their systems, thoroughly answering the question: ``to what extent these simulations are a good proxy to human behavior'' remains challenging and limited. In fact, validating social simulations requires, as stated in \cite{sen2025validating}, \cite{zhou2025pimmur} and in \cite{larooij2025large}, careful methodological approaches. For the social simulations to be realistic and reliable enough, we need, among other things, to ensure that the agents display similar cognitive bias to humans in an informational environment.

% Our positioning 
One widespread and established cognitive bias in humans which manifests also on social media is the Illusory Truth Effect (ITE). Initially observed in \cite{hasher1977frequency}, the ITE simply states that repeated claims are rated as more true than new ones \cite{bacon1979credibility}. This effect is highly relevant to social media usage and misinformation adoption. In fact, it has been shown that prior exposure increases perceived accuracy of fake news \cite{pennycook2018prior}, even if it contradicts previous knowledge \cite{fazio2020repetition,fazio2015knowledge}. Repetition has also been shown to increase the perceived truth of conspiracy theories \cite{bena2023repetition} and rumors \cite{difonzo2016validity}. The ITE’s impact goes beyond belief, it increases sharing intentions of news headlines (including false ones) and decreases how unethical an act is perceived to be \cite{udry2024illusory}.

% Explaining the project
In this work, we contribute to the scholarly efforts on validating generative ABM and focus on investigating the manifestations –or not— of the ITE, in the context of social media simulation in particular. We design an experiment where we instruct an LLM to behave like a social media user and feed it a series of tweets iteratively to interact with. In every iteration of news feeds, we include a statement (in a form of a tweet) which appears in every one of the presented feeds. At the end of the experiment, we task the LLM to rate the truth value of the repeated statement as well as additional completely new statements (unseen statements). We then compare, across several runs of the experiment, the difference between repeated and unseen statements. We push the analysis further to distinguish between an ITE specific to truth or a mere exposure effect \cite{zajonc1968attitudinal}. Our main research question (RQ) is: ``is ITE effect present in social media simulations?'' If an effect is observed, we expend our analysis to cover the following questions:
\begin{itemize}
    \item[RQ1:] Is the increased rating effect specific to truth or it is rather a mere exposure effect? 
    \item[RQ2:] How does the ITE and mere exposure effects differ across models and their temperatures? 
    \item[RQ3:] Does the observed effect depend on the experimental context? 
\end{itemize}

Our findings suggest that LLMs display different levels and types of effects in relation to the ITE in social media simulations. Additionally, we find that the effect is highly dependent on the experimental context as well as the specific statement to rate.
\vspace{-8pt}
\section{Related Work} 
\vspace{-10pt}
\subsubsection*{The Illusory Truth Effect}
\vspace{-6pt}
\cite{hasher1977frequency} first demonstrated the ITE phenomenon in 1977: college students rated 60 statements across three sessions separated by two-week intervals, with some statements repeated across sessions and others appearing only once. Repeated statements received significantly higher validity judgments than non-repeated ones — for both true and false statements — suggesting that frequency of occurrence serves as a criterion for judging the referential validity of plausible statements. These findings were subsequently replicated in several studies referring to the effect as the Illusion of Validity \citep{bacon1979credibility,schwartz1982repetition,gigerenzer1984external,begg1985believing,arkes1989generality}.

The ITE is typically explained as a fluency effect: initial exposure makes processing during the test phase more fluent, and fluency is taken as an indicator of truth \citep{reber1999effects}. A related phenomenon, the mere exposure effect \citep{zajonc1968attitudinal}, occurs when increased familiarity increases ratings in other dimensions such as liking or importance. Most ITE studies do not test whether the effect is specific to truth judgments, leaving it conflated with mere exposure. Our study addresses this by comparing the effect on truth ratings versus other dimensions, following the discriminant validity paradigm \citep{campbell1959convergent} and the approach of \cite{griffin2023large}.

Beyond controlled experiments, the ITE has real-world implications for misinformation. As reviewed in \cite{udry2024illusory}, it reproduces in digital environments, particularly social media. \cite{pennycook2018prior} found that even a single prior exposure increases perceived accuracy, while \cite{vellani2023illusory} showed that participants were more likely to share previously encountered statements — with this relationship mediated by perceived accuracy, making repetition a driver of misinformation spread.
\vspace{-10pt}
\subsubsection*{Generative ABM for Social Simulations}
\vspace{-6pt}
Concurrently, generative agent-based models (ABMs) powered by LLMs have become state-of-the-art in social simulations. Building on \cite{park2023generative}'s foundational work on believable generative agents, researchers have rapidly adopted this approach across a range of social simulation studies \citep{ferraro2024agent,yang2024oasis,gao2023s3,tornberg2023simulating,composta2025simulating,larooij2025can,de2023emergence,bouleimen2025collective}. These studies demonstrate LLMs' promise as surrogates for human participants, reproducing interaction and linguistic patterns \citep{ferraro2024agent,yang2024oasis}, group dynamics \citep{bouleimen2025collective}, coherent content generation \citep{composta2025simulating}, and emotional and attitudinal responses \citep{gao2023s3}. At the network level, LLMs reproduce social phenomena such as homophily \citep{ferraro2024agent}, the friendship paradox \citep{orlando2025can}, opinion dynamics \citep{cau2025selective}, scale-free network structures \citep{de2023emergence}, political clustering and echo chambers \citep{ferraro2024agent,larooij2025can}, and concentrated influence among elite agents \citep{larooij2025can}. That said, limitations remain, including difficulty reproducing toxic behavior and limited content diversity \citep{composta2025simulating,pagan2025computational}.
\vspace{-10pt}
\subsubsection*{ITE and LLMs}

\vspace{-6pt}
While this body of work sheds light on high-level interaction and network dynamics, individual cognitive biases in LLMs remain underexplored. A growing literature addresses LLM belief and knowledge change in this context. \cite{fastowski2024understanding} introduce the notion of \textit{knowledge drift}, showing that repeatedly exposing a model to incorrect information increases its uncertainty and can shift initially correct answers. \cite{geng2025accumulating} further show that models tend to shift moral or political positions after multi-model discussions or exposure to opposing viewpoints. Whether such susceptibility is useful for realistic social simulation remains debated, but it underscores significant context-dependence in LLM behavior.

To the best of our knowledge, \cite{griffin2023large} is the only study to examine the ITE in LLMs directly. Replicating \cite{hasher1977frequency} and using \texttt{text-davinci-003}, the authors explore the ITE manifestations through a set of true and false claims. They found that repeated statements received significantly higher truth ratings, with no significant effect on interest, sentiment, or importance — ruling out a mere exposure explanation. However, this study is limited to a now-deprecated model, and recent findings \citep{fastowski2024understanding,geng2025accumulating} suggest that context-sensitivity may affect ITE-related behavior. It is therefore not clear that the effect observed by \cite{griffin2023large} will generalize to newer models or different prompting contexts. This motivates our work, which tests multiple models across varied social simulation contexts, drawing on the experimental design of \cite{griffin2023large} and adapting it accordingly.
\vspace{-6pt}
\section{Experiment}

\begin{figure}
    \centering
    \includegraphics[width=\linewidth]{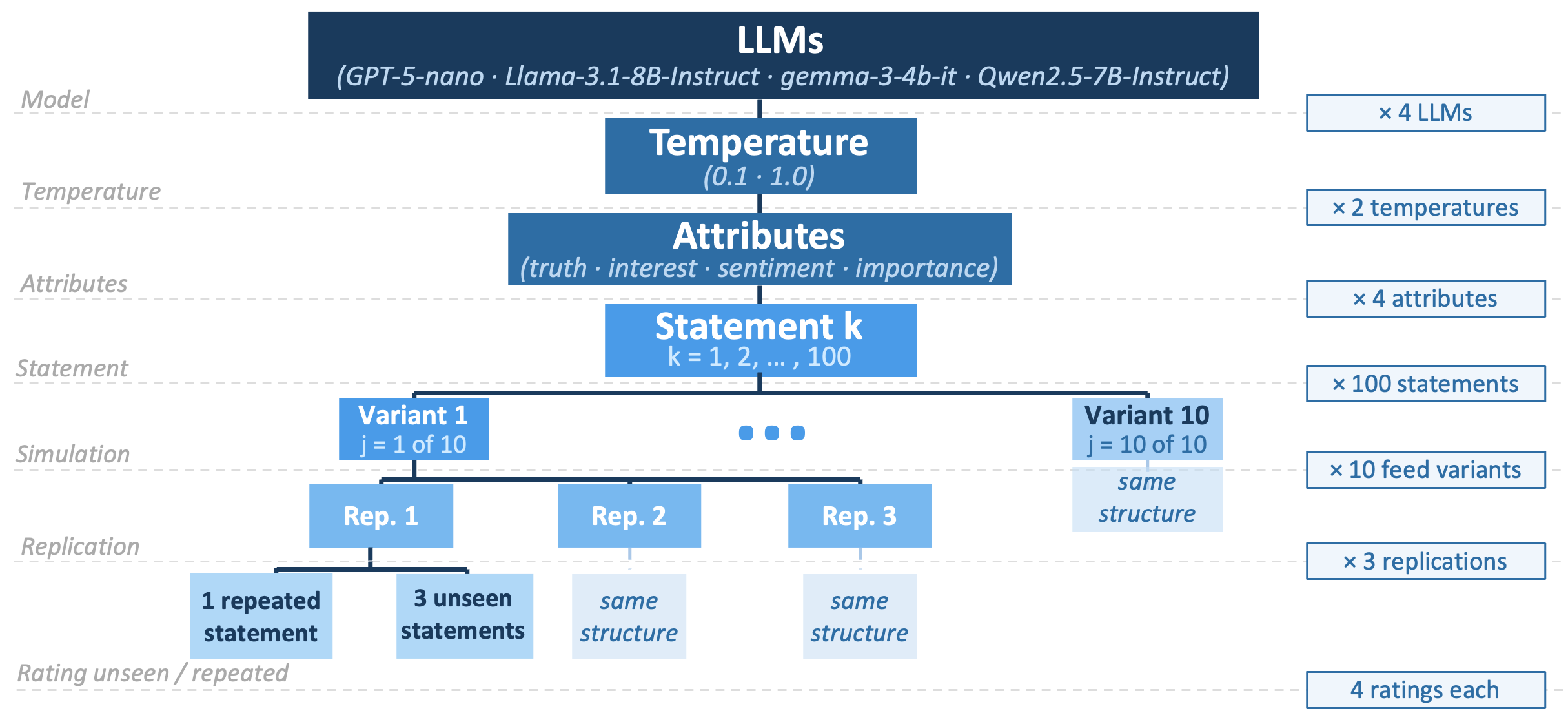}
    \caption{Illustration of the experiment.}
    \label{fig:experiment}
\end{figure}

\vspace{-6pt}
\paragraph{Design.} We design a controlled experiment to test whether LLMs exhibit ITE-like behavior when acting as agents in a social media simulation. The experiment is run on four LLMs: \texttt{Gemma-3-4b-it} \citep{kamath2025gemma3}, \texttt{Qwen2.5-7B-Instruct} \citep{yang2024qwen25}, \texttt{Llama-3.1-8B-Instruct} \citep{dubey2024llama3}, and \texttt{GPT-5-nano} \citep{openai2025gpt5}. For the three open-weight models, we run the experiment at two temperature settings (0.1 and 1.0) to assess sensitivity to this parameter, while \texttt{GPT-5-nano} has a fixed default temperature of 1.0. This yields 7 model-temperature combinations in total. The statements used throughout the experiment are the 100 statements authored by \cite{griffin2023large}, which are not available online, minimizing the risk that they appear verbatim in the training data of the tested LLMs (see Appendix~\ref{list_statements}). Each statement is rated along one of four attributes: \textit{truth}, \textit{importance}, \textit{sentiment}, and \textit{interest}. Truth is the primary attribute of interest for our study, directly operationalizing the ITE. The remaining three serve to distinguish a genuine ITE from a mere exposure effect \citep{zajonc1968attitudinal}: if repetition boosts ratings uniformly across all attributes, the effect is better attributed to familiarity than to a truth-specific bias.
\vspace{-6pt}
\paragraph{Procedure.} For each combination of model, temperature, statement, and rating attribute, the experiment unfolds in two sequential phases within the same context window. In the \textit{simulation phase}, the LLM is prompted to act as a social media user scrolling through a news feed and commenting, resharing, or not interacting with the tweets (see Appendix~\ref{ex_conversation}). It is shown 5 feeds sequentially, each containing 4 tweets. One tweet, the \textit{repeated statement}, appears in every feed, while the remaining 15 are unique filler statements drawn randomly from the pool of 99 (excluding the repeated statement in this experiment), each appearing exactly once. The position of the repeated tweet within each feed is randomized. In the following \textit{rating phase}, without resetting the context window, the LLM is asked to rate 4 claims on a 1-to-7 Likert scale along the target attribute. One of the 4 claims is the repeated statement; the other 3 are completely new, i.e., so far unseen by the LLM in the experiment. The position of the repeated claim in the rating list is again randomized. Scale levels for each attribute are described in Appendix~\ref{rating_prompts}. With this setting, the LLM interacts with or rates a total 24 statements. Given the experiment with humans rating 32 statements each done in \cite{griffin2023large}, we believe that our setting is suitable to replicate (if any exists) the ITE of humans in social media.
To account for variability in the filler context and following the approach in \cite{griffin2023large}, we generate 10 feed variants per statement, each with a different random draw of filler tweets. We replicate each variant 3 times identically to assess the determinism of model responses. This yields 30 ratings per statement in the \textit{repeated} condition and 90 in the \textit{unseen} condition, for each (attribute, model, temperature) combination, and 336,000 ratings in total: 7 model-temperature combinations × 4 attributes × 100 statements × 10 feed variants × 3 replications × 4 ratings per pair of phases. An illustration of the experiment is presented in~\ref{fig:experiment}.

\vspace{-10pt}
\section{Statistical Analysis}
\label{sec:stats}
\vspace{-6pt}

To address the RQs, we first implement an Ordinary Least Squares Regression (OLS) replicating the model used in \cite{griffin2023large}. We then gain additional insights from the data through a Linear Mixed-Effects Model (LMEM). We detail both statistical models in the following subsections.
\vspace{-11pt}
\subsection{Descriptive Analysis: OLS Regression}
\vspace{-6pt}

To characterize the magnitude and direction of repetition effects for each model, attribute, and temperature combination, we fit separate OLS regressions of the form:
\begin{equation}
  r' = r + \text{offset} + (r - 4) \times \text{tilt},
  \label{eq:ols}
\end{equation}
where $r$ denotes the mean rating a statement received across its unseen presentations, $r'$ denotes the mean rating the same statement received across its repeated presentations, and 4 is the midpoint of the 1-to-7 rating scale. This parametrization yields two interpretable quantities. The \textit{offset} estimates the predicted shift in rating due to repetition for a statement rated neutrally (4) when unseen. The \textit{tilt} estimates how the repetition effect scales with the baseline (unseen) rating: a tilt of zero implies a uniform additive shift across all statements; a negative tilt implies a smaller effect for statements with high baseline ratings; a positive tilt implies a larger effect for statements with high baseline ratings. We fit each regression on 100 data points, one per statement, computing $r$ and $r'$ as means across the 90 unseen and 30 repeated experimental runs, respectively. We fit one regression per combination of model (4 levels), attribute (4 levels), and temperature (2 levels for open-weight models; 1 level for \texttt{GPT-5-nano}), yielding 28 regressions in total.

This model provides a descriptive understanding of the effects observed across every LLM, attribute, and temperature. However, it cannot compare effects across models and attributes, as observations are not independent across groups. In other words, with this model we cannot say whether the effect for truth is higher or lower than for the other attributes (RQ1), nor can we distinguish effects between models (RQ2). Moreover, any insights on the variance of the ratings is made impossible when we consider the mean rating per statement (RQ3).

\vspace{-11pt}
\subsection{Inferential Analysis: Linear Mixed-Effects Models}
\vspace{-6pt}
To address the limitations of the OLS regression, we adopt a LMEM. We first fit a preliminary model to measure the effects of temperature and replication on ratings (see Appendix~\ref{prelim_stat_model} for details). 

Following the preliminary model results (see Section~\ref{res:prelim}), we fit a main model with fixed effects for a three-way interaction between statement repetition, rating attribute, and LLM. This model lets us simultaneously test differences in the repetition effect across attributes and LLMs, addressing both RQ1 (truth-specificity of the rating effect: ITE vs.\ mere exposure) and RQ2 (differences in effects across LLMs). We also use the LMEM to conduct a variance component analysis, which helps us understand the potential impact of simulation context on LLM ratings of repeated statements (RQ3). Our data have a nested structure: individual rating observations are grouped within simulation runs (10 runs per statement in the repeated condition, \texttt{simulation\_num}), which are in turn grouped within statements (100 different statements, \texttt{statement\_id}). \textit{Statement-level clustering} reflects that some statements receive intrinsically higher or lower ratings across all attributes and conditions, independent of repetition. \textit{Simulation-level clustering} reflects that the surrounding feed context---the specific set of filler tweets presented alongside the repeated statement during the simulation phase---introduces additional variability in ratings beyond statement-level effects. To account for both clustering sources, we include nested random intercepts in the model. 

The resulting LMEM is presented in Eq.~\ref{eq:3way_model}:
\begin{equation}
  \texttt{rating} \sim \texttt{repeated} \times \texttt{attribute} \times \texttt{model} + 1 | \texttt{statement\_id}/\texttt{simulation\_num},
  \label{eq:3way_model}
\end{equation}
where \texttt{repeated} is a binary variable indicating whether the rated statement was repeated (True) or unseen (False) during the simulation phase, \texttt{attribute} is the rating dimension (truth, importance, sentiment, interest), and \texttt{model} identifies the LLM (\texttt{GPT-5-nano}, \texttt{Llama-3.1-8B-Instruct}, \texttt{Gemma-3-4b-it}, and \texttt{Qwen2.5-7B-Instruct}). \texttt{statement\_id} and \texttt{simulation\_num} enter the model as nested random effects. The reference levels are: \texttt{repeated} = False, \texttt{attribute} = truth, \texttt{temperature} = 0.1.
\vspace{-11pt}
\paragraph{Post-hoc Contrasts}
To extract interpretable per-model, per-attribute repetition effects from the three-way model, we compute the estimated marginal means (EMMs)---the model's best estimates of the average rating for each combination of repetition status (unseen or repeated), attribute, and LLM (Methodological details can be found in Appendix~\ref{emmeans}). We then compute pairwise contrasts (\texttt{repeated} $-$ \texttt{unseen}) within each attribute $\times$ model cell, that is, the difference between the estimated means of repeated and unseen ratings. A positive contrast indicates that repeated statements received higher ratings than unseen statements, while a negative contrast indicates the reverse. Comparing these contrasts across attributes and models addresses whether rating effects are truth-specific (RQ1) and whether these effects differ across models (RQ2).
\vspace{-11pt}
\paragraph{Variance Component Analysis}
We compute marginal and conditional $R^2$ values and Intraclass Correlations (ICCs) at both the statement and simulation levels. Marginal $R^2$ reflects the variance explained by fixed effects alone, while conditional $R^2$ reflects the variance explained by both fixed and random effects. The ICC quantifies the proportion of total variance attributable to between-statement and between-simulation differences, respectively. Together, these measures reveal the sources of variance in the data and shed light on the potential impact of context on statement ratings, thereby addressing RQ3.

We fit the LMEM using the \texttt{nlme} package \citep{nlme} in R \citep{RCoreTeam}. We computed EMMs using the \texttt{emmeans} package \citep{emmeans}, and marginal and conditional $R^2$ values following the method of \citet{nakagawa2013} using the \texttt{performance} package \citep{performance}.

\section{Results}
We run the experiment on the open weights models on A100 and H100 GPUs. The total running time for the 100 x 10 experiments per rating attribute was about 2.5 hours for \texttt{Qwen-2.5-7B-Instruct}, about 3 hours for \texttt{Llama-3.1-8B-Instruct}, and about 12 hours for \texttt{Gemma-3-4b-it}. We run the \texttt{GPT-5-nano} using asynchronous calls from the OpenAI API for which the experiment lasted only few minutes. We collected 336,000 ratings, from which we dropped 155. 151 of them came from \texttt{GPT-5-nano} where it either did not abide by the new rating task and kept interacting with the claim as a social media user or refrained from rating this happened in majority for statements 84 and 89. The remaining ratings we dropped were 4 from \texttt{Llama-3.1-8B-Instruct} where it refrained from rating statement 24 in four occasions. We detail the findings from 335,845 ratings in the following subsections.

\vspace{-7pt}
\subsection{Descriptive Analysis: Repetition Effects by Model and Attribute}
\begin{figure}[h]
    \centering
    \begin{subfigure}[b]{0.24\textwidth}
        \centering
        \includegraphics[width=\textwidth]{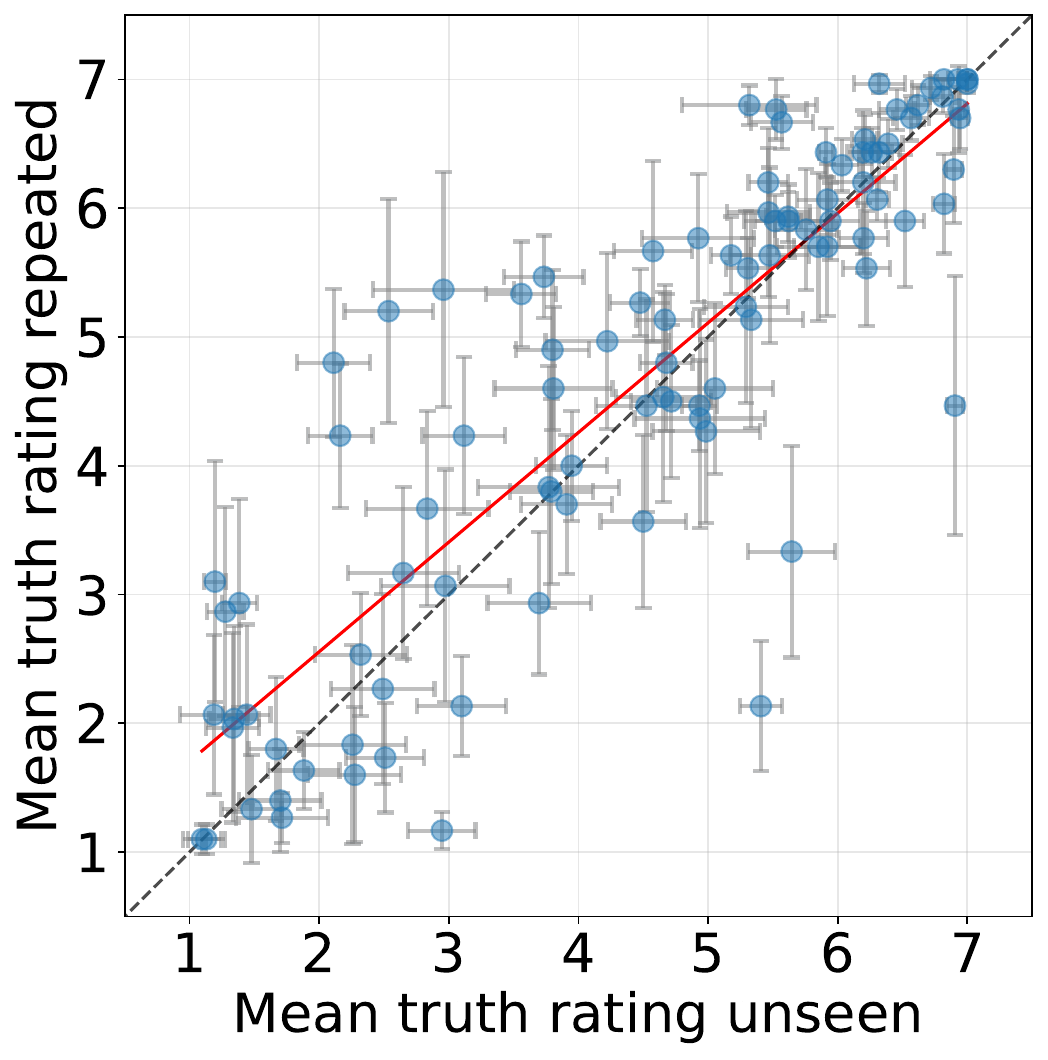}
        \caption{}
        
    \end{subfigure}
    \hfill
    \begin{subfigure}[b]{0.24\textwidth}
        \centering
        \includegraphics[width=\textwidth]{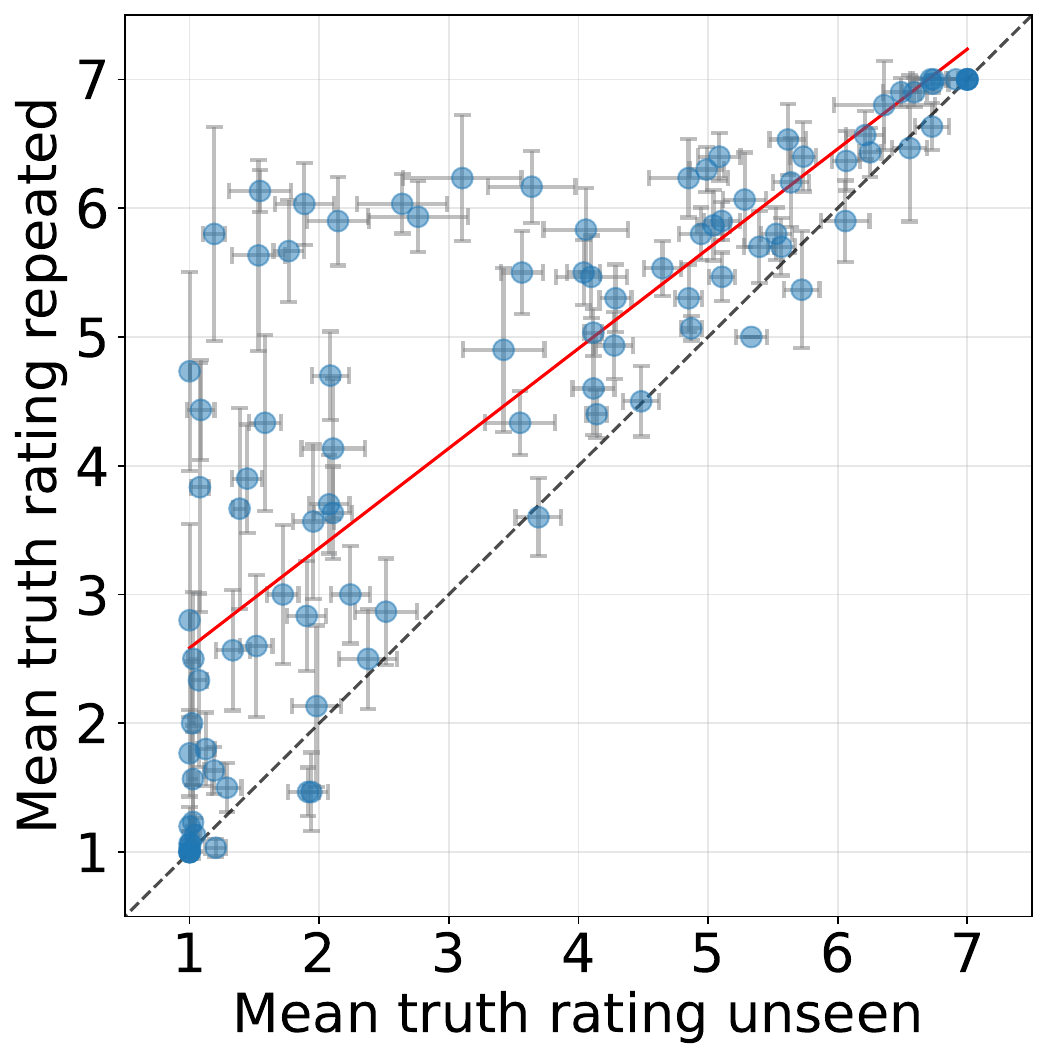}
        \caption{}
        
    \end{subfigure}
    \hfill
    \begin{subfigure}[b]{0.24\textwidth}
        \centering
        \includegraphics[width=\textwidth]{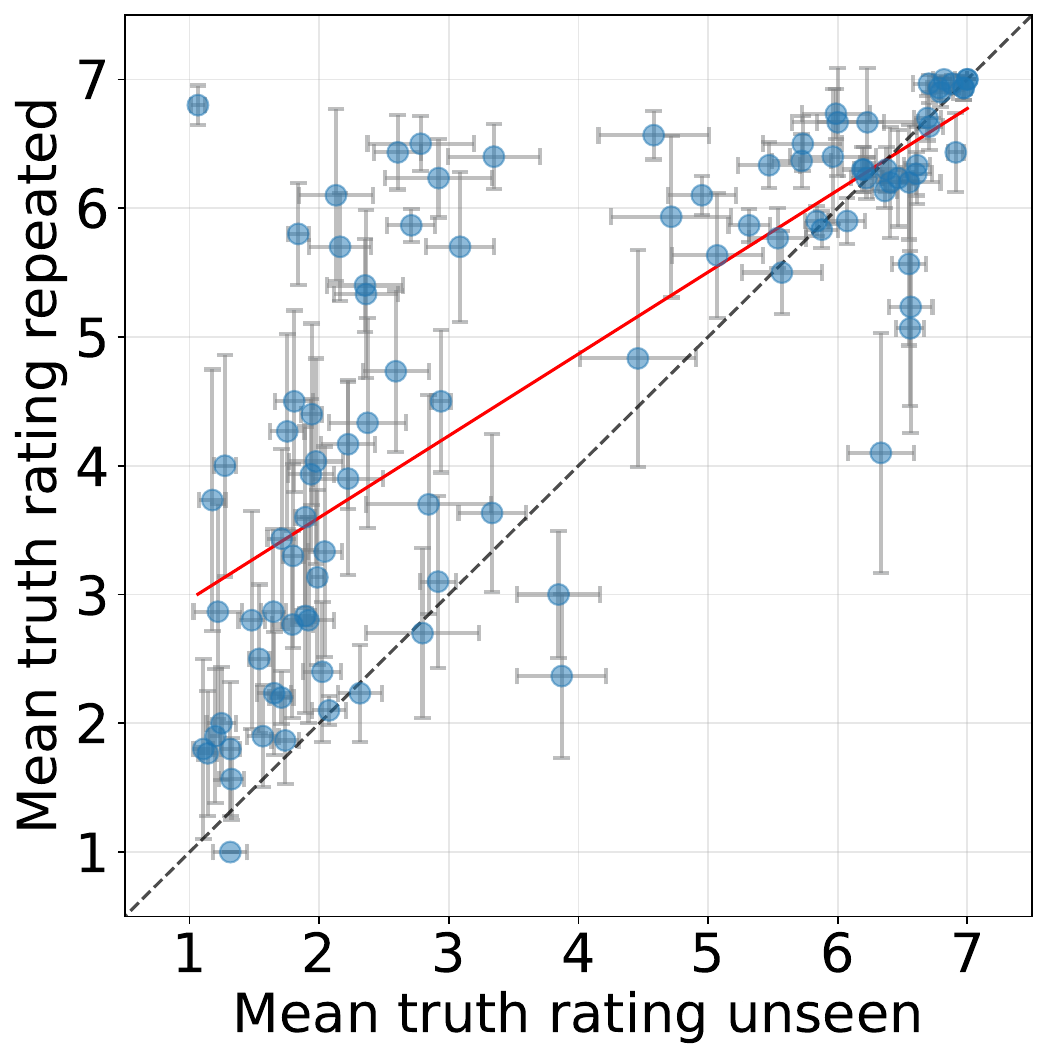}
        \caption{}
        
    \end{subfigure}
    \hfill
    \begin{subfigure}[b]{0.24\textwidth}
        \centering
        \includegraphics[width=\textwidth]{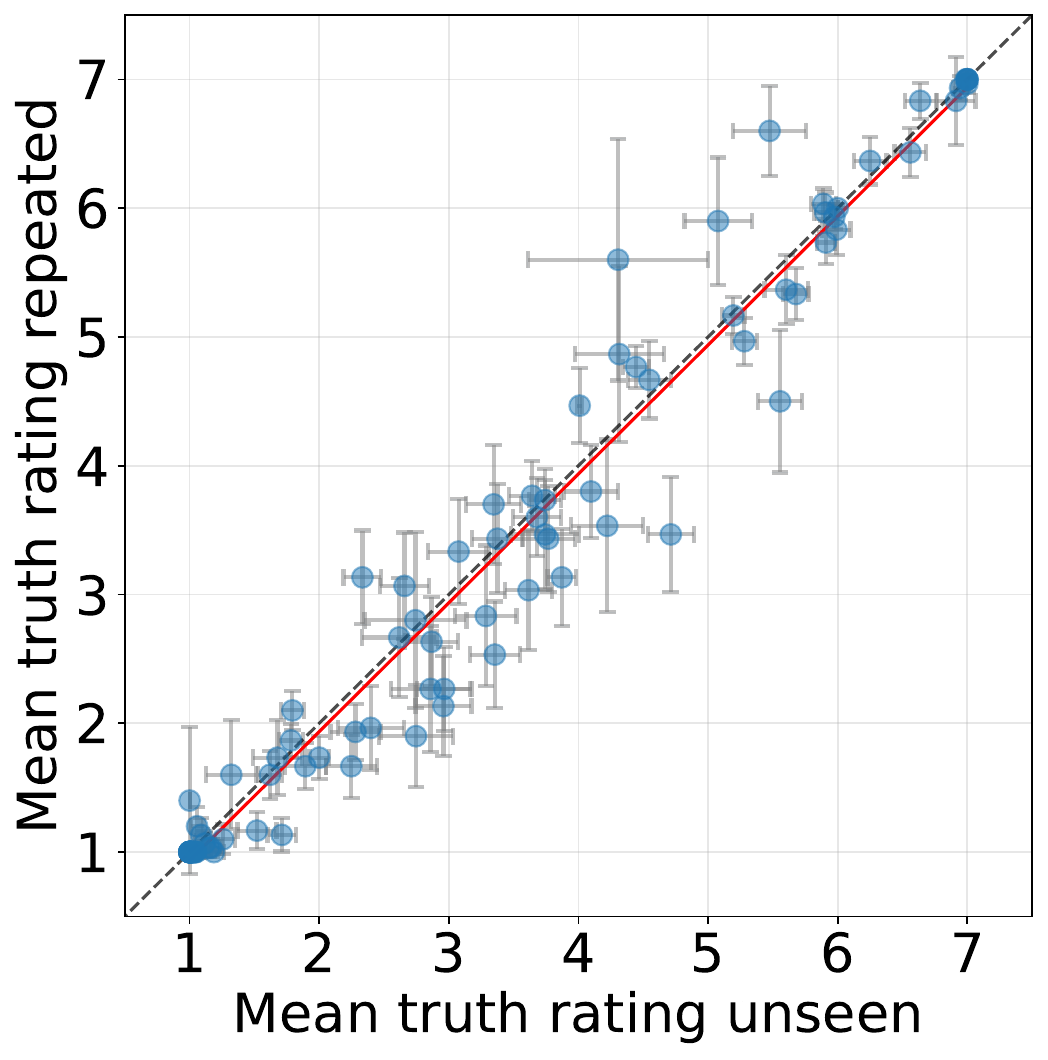}
        \caption{}
        
    \end{subfigure}
    \caption{Plots of the mean truth rating per statement when unseen and repeated in the simulation phase. (a) \texttt{Llama-3.1-8B-Instruct}, (b) \texttt{Qwen2.5-7B-Instruct}, (c) \texttt{Gemma-3-4b-it} and (d) \texttt{GPT-5-nano}. All models are set to temperature 1. Error bars are 95\% confidence intervals. The dashed black line is the identity function, the solid red line is the best OLS fit.}
    \label{fig:plots}
\end{figure}

\vspace{-6pt}
We report in Tab.~\ref{tab:res} the results of the OLS regression (Eq.~\ref{eq:ols}). It shows the estimates of tilt and offset\footnote{We assessed the robustness of regression estimates to potential ceiling effects by repeating all regressions on the subset of statements with mean unseen ratings $r \leq 6$. Offset and tilt estimates were not materially affected, indicating that compression at the upper boundary of the rating scale did not substantially distort the regression coefficients.}. To illustrate some of the numbers in the results, we present in Fig.~\ref{fig:plots} the scatter plots and the corresponding regression lines for the four models studied when temperature is 1. Fig.~\ref{fig:plots} shows a difference in the effect across the LLMs. The remaining plots for temperature 0.1 and for the importance, sentiment and interest attributes can be found in Appendix~\ref{all_plots}.

A first glance at the table indicates that our experiment shows almost no difference between the regression parameters from temperature 1 and from temperature 0.1. In fact, for truth, the highest difference in absolute value between the offsets and tilts are respectively 0.047 and 0.076. In all cases except one, the estimates are always with similar values, positive / negative sign and same statistical significance level. The only exception is for sentiment in \texttt{Llama-3.1-8B} experiment where the tilt is not significant for temperature 1 while it is significant when temperature is 0.1. This is a rather counterintuitive result as higher temperatures are expected to produce more diverse or creative output which may lead to a different behavior of the model towards the ratings.

Focusing on the truth rating attribute, we see that both the offset and the tilt are significant for \texttt{Llama-3.1-8B}, \texttt{Qwen2.5-7B} and for the \texttt{Gemma-3-4b-it}. In all three cases the offset is positive. This indicates an overall increase in the truth rating of the claims when repeated in the simulation phase compared to when unseen. For all these three models, the tilt is negative, which indicates that the observed rating's increase is of a lower magnitude for the highly rated unseen claims and, vice versa, of a higher magnitude for lowly rated unseen claims. As for \texttt{GPT-5-nano}, both estimates are not statistically significant highlighting no effect of previous exposure to claim on the truth rating of \texttt{GPT-5-nano}. 
Looking at the sentiment, interest and importance attributes, we see that the offset is always significant and positive for \texttt{Qwen2.5-7B} and \texttt{Gemma-3-4b-it} which indicates an increase in the average rating of these attributes for \texttt{Qwen2.5-7B} and \texttt{Gemma-3-4b-it}. For \texttt{GPT-5-nano}, the offset for sentiment and interest are significant but negative instead, suggesting a decrease in the average ratings of these attributes. Finally, \texttt{Llama-3.1-8B} shows a significant negative offset only for sentiment.

Considering these results and as detailed in Sec.~\ref{sec:stats}, the OLS model fails in providing statistical inference about the cross level comparison between attributes and models to allow, at this level of the analysis, to make conclusions on the presence of absence of the ITE and whether the observed effect is different from a mere exposure effect. We then present, in the next two subsections, the results from the LMEM.

\begin{table}[ht]
\small
\setlength{\tabcolsep}{4pt}
\renewcommand{\arraystretch}{0.95}
\begin{center}
\caption{Statistical model estimates for the four models across four rating attributes. 
\colorbox[HTML]{F2DCDB}{Red} = significant positive/negative offset; 
\colorbox[HTML]{DCE6F1}{Blue} = significant tilt.
Bold = $p < 0.05$. Significance: {*}$p<0.05$, {**}$p<0.01$, {***}$p<0.001$.
\textsuperscript{†}\texttt{GPT-5-nano} temperature is fixed to 1; Temp.\ 0.1 results not available.}
\label{tab:res}
\resizebox{\linewidth}{!}{
\begin{tabular}{llr rr rr}
\toprule
% \rowcolor[HTML]{EEECE1}
\textbf{Model} & \textbf{Rating Attr.} & \textbf{Param.} 
  & \multicolumn{2}{c}{\textbf{Temp. 1}} 
  & \multicolumn{2}{c}{\textbf{Temp. 0.1}} \\
% \rowcolor[HTML]{EEECE1}
 &  &  & \textbf{Est.} & \textbf{95\% CI} & \textbf{Est.} & \textbf{95\% CI} \\
\midrule

\multirow{8}{*}{\textbf{Llama-3.1-8B}} 
  & \multirow{2}{*}{truth}      & offset & \cellcolor[HTML]{F2DCDB}\textbf{0.258**}  & \cellcolor[HTML]{F2DCDB}\textbf{[0.073, 0.443]}  & \cellcolor[HTML]{F2DCDB}\textbf{0.269*}   & \cellcolor[HTML]{F2DCDB}\textbf{[0.042, 0.496]} \\
  &                             & tilt   & \cellcolor[HTML]{DCE6F1}\textbf{−0.149**} & \cellcolor[HTML]{DCE6F1}\textbf{[-0.245, −0.052]}& \cellcolor[HTML]{DCE6F1}\textbf{−0.225***}& \cellcolor[HTML]{DCE6F1}\textbf{[−0.327, −0.122]} \\
  & \multirow{2}{*}{importance} & offset & 0.078  & [−0.068, 0.223]  & −0.102 & [−0.275, 0.071] \\
  &                             & tilt   & \cellcolor[HTML]{DCE6F1}\textbf{−0.205***}& \cellcolor[HTML]{DCE6F1}\textbf{[−0.302, −0.108]}& \cellcolor[HTML]{DCE6F1}\textbf{−0.191***}& \cellcolor[HTML]{DCE6F1}\textbf{[−0.296, −0.087]} \\
  & \multirow{2}{*}{sentiment}  & offset & \cellcolor[HTML]{F2DCDB}\textbf{−0.231**} & \cellcolor[HTML]{F2DCDB}\textbf{[−0.360, −0.103]}& \cellcolor[HTML]{F2DCDB}\textbf{−0.251***}& \cellcolor[HTML]{F2DCDB}\textbf{[−0.385, −0.117]} \\
  &                             & tilt   & −0.064 & [−0.147, 0.018]  & \cellcolor[HTML]{F2DCDB}\textbf{−0.132**} & \cellcolor[HTML]{F2DCDB}\textbf{[−0.212, −0.052]} \\
  & \multirow{2}{*}{interest}   & offset & 0.016  & [−0.124, 0.155]  & −0.010 & [−0.195, 0.175] \\
  &                             & tilt   & \cellcolor[HTML]{DCE6F1}\textbf{−0.197***}& \cellcolor[HTML]{DCE6F1}\textbf{[−0.285, −0.108]}& \cellcolor[HTML]{DCE6F1}\textbf{−0.236***}& \cellcolor[HTML]{DCE6F1}\textbf{[−0.345, −0.126]} \\
\midrule

\multirow{8}{*}{\textbf{Qwen2.5-7B}}
  & \multirow{2}{*}{truth}      & offset & \cellcolor[HTML]{F2DCDB}\textbf{0.910***} & \cellcolor[HTML]{F2DCDB}\textbf{[0.675, 1.145]}  & \cellcolor[HTML]{F2DCDB}\textbf{0.903***} & \cellcolor[HTML]{F2DCDB}\textbf{[0.660, 1.146]} \\
  &                             & tilt   & \cellcolor[HTML]{DCE6F1}\textbf{−0.225***}& \cellcolor[HTML]{DCE6F1}\textbf{[−0.334, −0.116]}& \cellcolor[HTML]{DCE6F1}\textbf{−0.185**} & \cellcolor[HTML]{DCE6F1}\textbf{[−0.297, −0.073]} \\
  & \multirow{2}{*}{importance} & offset & \cellcolor[HTML]{F2DCDB}\textbf{1.003***} & \cellcolor[HTML]{F2DCDB}\textbf{[0.812, 1.194]}  & \cellcolor[HTML]{F2DCDB}\textbf{1.034***} & \cellcolor[HTML]{F2DCDB}\textbf{[0.828, 1.240]} \\
  &                             & tilt   & \cellcolor[HTML]{DCE6F1}\textbf{−0.216***}& \cellcolor[HTML]{DCE6F1}\textbf{[−0.314, −0.118]}& \cellcolor[HTML]{DCE6F1}\textbf{−0.215***}& \cellcolor[HTML]{DCE6F1}\textbf{[−0.321, −0.109]} \\
  & \multirow{2}{*}{sentiment}  & offset & \cellcolor[HTML]{F2DCDB}\textbf{0.451***} & \cellcolor[HTML]{F2DCDB}\textbf{[0.309, 0.594]}  & \cellcolor[HTML]{F2DCDB}\textbf{0.438***} & \cellcolor[HTML]{F2DCDB}\textbf{[0.301, 0.575]} \\
  &                             & tilt   & −0.039 & [−0.140, 0.061]  & −0.019 & [−0.114, 0.076] \\
  & \multirow{2}{*}{interest}   & offset & \cellcolor[HTML]{F2DCDB}\textbf{0.642***} & \cellcolor[HTML]{F2DCDB}\textbf{[0.463, 0.821]}  & \cellcolor[HTML]{F2DCDB}\textbf{0.546***} & \cellcolor[HTML]{F2DCDB}\textbf{[0.380, 0.712]} \\
  &                             & tilt   & −0.115 & [−0.236, 0.006]  & −0.096 & [−0.209, 0.016] \\
\midrule

\multirow{8}{*}{\textbf{Gemma-3-4B}}
  & \multirow{2}{*}{truth}      & offset & \cellcolor[HTML]{F2DCDB}\textbf{0.868***} & \cellcolor[HTML]{F2DCDB}\textbf{[0.644, 1.092]}  & \cellcolor[HTML]{F2DCDB}\textbf{0.915***} & \cellcolor[HTML]{F2DCDB}\textbf{[0.687, 1.144]} \\
  &                             & tilt   & \cellcolor[HTML]{DCE6F1}\textbf{−0.365***}& \cellcolor[HTML]{DCE6F1}\textbf{[−0.469, −0.260]}& \cellcolor[HTML]{DCE6F1}\textbf{−0.404***}& \cellcolor[HTML]{DCE6F1}\textbf{[−0.510, −0.298]} \\
  & \multirow{2}{*}{importance} & offset & \cellcolor[HTML]{F2DCDB}\textbf{0.882***} & \cellcolor[HTML]{F2DCDB}\textbf{[0.687, 1.077]}  & \cellcolor[HTML]{F2DCDB}\textbf{0.822***} & \cellcolor[HTML]{F2DCDB}\textbf{[0.618, 1.026]} \\
  &                             & tilt   & \cellcolor[HTML]{DCE6F1}\textbf{−0.550***}& \cellcolor[HTML]{DCE6F1}\textbf{[−0.669, −0.431]}& \cellcolor[HTML]{DCE6F1}\textbf{−0.495***}& \cellcolor[HTML]{DCE6F1}\textbf{[−0.618, −0.371]} \\
  & \multirow{2}{*}{sentiment}  & offset & \cellcolor[HTML]{F2DCDB}\textbf{0.593***} & \cellcolor[HTML]{F2DCDB}\textbf{[0.402, 0.783]}  & \cellcolor[HTML]{F2DCDB}\textbf{0.630***} & \cellcolor[HTML]{F2DCDB}\textbf{[0.435, 0.825]} \\
  &                             & tilt   & \cellcolor[HTML]{DCE6F1}\textbf{−0.158**} & \cellcolor[HTML]{DCE6F1}\textbf{[−0.267, −0.048]}& \cellcolor[HTML]{DCE6F1}\textbf{−0.147*}  & \cellcolor[HTML]{DCE6F1}\textbf{[−0.261, −0.034]} \\
  & \multirow{2}{*}{interest}   & offset & \cellcolor[HTML]{F2DCDB}\textbf{0.348**}  & \cellcolor[HTML]{F2DCDB}\textbf{[0.141, 0.555]}  & \cellcolor[HTML]{F2DCDB}\textbf{0.314**}  & \cellcolor[HTML]{F2DCDB}\textbf{[0.103, 0.524]} \\
  &                             & tilt   & \cellcolor[HTML]{DCE6F1}\textbf{−0.352***}& \cellcolor[HTML]{DCE6F1}\textbf{[−0.486, −0.219]}& \cellcolor[HTML]{DCE6F1}\textbf{−0.363***}& \cellcolor[HTML]{DCE6F1}\textbf{[−0.501, −0.226]} \\
\midrule

\multirow{8}{*}{\textbf{GPT-5-nano}\textsuperscript{†}}
  & \multirow{2}{*}{truth}      & offset & −0.063 & [−0.147, 0.020] & \multicolumn{2}{c}{—} \\
  &                             & tilt   & 0.001  & [−0.036, 0.038] & \multicolumn{2}{c}{—} \\
  & \multirow{2}{*}{importance} & offset & 0.000  & [−0.074, 0.074] & \multicolumn{2}{c}{—} \\
  &                             & tilt   & \cellcolor[HTML]{DCE6F1}\textbf{−0.166***}& \cellcolor[HTML]{DCE6F1}\textbf{[−0.221, −0.111]}& \multicolumn{2}{c}{—} \\
  & \multirow{2}{*}{sentiment}  & offset & \cellcolor[HTML]{F2DCDB}\textbf{−0.065**} & \cellcolor[HTML]{F2DCDB}\textbf{[−0.101, −0.028]}& \multicolumn{2}{c}{—} \\
  &                             & tilt   & \cellcolor[HTML]{DCE6F1}\textbf{−0.127***}& \cellcolor[HTML]{DCE6F1}\textbf{[−0.161, −0.092]}& \multicolumn{2}{c}{—} \\
  & \multirow{2}{*}{interest}   & offset & \cellcolor[HTML]{F2DCDB}\textbf{−0.119***}& \cellcolor[HTML]{F2DCDB}\textbf{[−0.182, −0.055]}& \multicolumn{2}{c}{—} \\
  &                             & tilt   & \cellcolor[HTML]{DCE6F1}\textbf{−0.250***}& \cellcolor[HTML]{DCE6F1}\textbf{[−0.313, −0.188]}& \multicolumn{2}{c}{—} \\
\bottomrule
\end{tabular}
}
\end{center}
\end{table}
\vspace{-10pt}

\vspace{-7pt}
\subsection{Inferential Analysis}
\vspace{-6pt}
\subsubsection{Preliminary Model: temperature and replication effects}
\label{res:prelim}
\vspace{-6pt}
Before arriving to the model in Eq.~\ref{eq:3way_model}, we run a preliminary model to measure the effect of the temperature and the replication on the ratings (Details about the model are presented in Appendix~\ref{prelim_stat_model}). The model did not show any significant difference between high and low temperatures, confirming the observation about the temperature in the OLS model~\ref{eq:ols}. Interestingly, the preliminary model also revealed that the LLMs' ratings were practically identical across replications. This highlights a counter-intuitive observation on the temperatures, which kept a very low variance in the LLM ratings even when set high. Consequently, we dropped the temperature as a predictor to be able to include \texttt{GPT-5-nano} ratings in the data, and simplified the random effects by disregarding the replication level. 

\subsubsection{Three-Way Interaction: repetition, attribute and LLM effects}

\vspace{-6pt}
The estimation of the fixed effects of all three-way interaction terms of the LMEM are all statistically significant ($p < .05$, see Appendix~\ref{three_way_model_all_data} for a list of all estimates). This indicates that the specificity of the repetition effect differed meaningfully across models. We report in Tab.~\ref{tab:emmeans} the pairwise contrasts of the repeated - unseen estimated means from the LMEM. We notice four qualitatively distinct patterns\footnote{We also run the model on the subset of the data when temperature is 1. The same observations hold (See Tab.~\ref{tab:delta_contrasts_all_data} in Appendix~\ref{three_way_model_temp1}).}

\begin{table}[htbp]
\small
\centering
\caption{Pairwise Contrasts ($\hat{\delta}$ EMMs repeated~$-$~EMMs unseen) by Attribute and Model. Significance: *** $p < .001$, ** $p < .01$, * $p < .05$, \textit{ns} $p \geq .05$.}
\vspace{-6pt}
\label{tab:emmeans}
\footnotesize
\setlength{\tabcolsep}{5pt}
\renewcommand{\arraystretch}{1.15}
\begin{threeparttable}
\resizebox{\linewidth}{!}{%
\begin{tabular}{l rrl rrl rrl rrl}
\toprule
& \multicolumn{3}{c}{\textit{Gemma-3-4b-it}}
& \multicolumn{3}{c}{\textit{gpt-5-nano}}
& \multicolumn{3}{c}{\textit{Llama-3.1-8B}}
& \multicolumn{3}{c}{\textit{Qwen2.5-7B}} \\
\cmidrule(lr){2-4}\cmidrule(lr){5-7}\cmidrule(lr){8-10}\cmidrule(lr){11-13}
Attribute & $\hat{\delta}$ & $t$ & & $\hat{\delta}$ & $t$ & & $\hat{\delta}$ & $t$ & & $\hat{\delta}$ & $t$ & \\
\midrule
truth
  & $+0.977$ & $\phantom{-}35.22$ & ***
  & $-0.041$ & $-1.02$            & \textit{ns}
  & $+0.097$ & $\phantom{-}3.50$  & ***
  & $+1.012$ & $\phantom{-}36.55$ & *** \\
importance
  & $+0.607$ & $\phantom{-}22.04$ & ***
  & $+0.003$ & $\phantom{-}0.08$  & \textit{ns}
  & $-0.149$ & $-5.39$            & ***
  & $+1.184$ & $\phantom{-}42.83$ & *** \\
interest
  & $+0.083$ & $\phantom{-}2.98$  & **
  & $-0.256$ & $-6.39$            & ***
  & $-0.107$ & $-3.87$            & ***
  & $+0.534$ & $\phantom{-}19.30$ & *** \\
sentiment
  & $+0.699$ & $\phantom{-}25.24$ & ***
  & $-0.080$ & $-1.98$            & *
  & $-0.322$ & $-11.64$           & ***
  & $+0.380$ & $\phantom{-}13.74$ & *** \\
\bottomrule
\end{tabular}%
}
\end{threeparttable}
\end{table}
\vspace{-10pt}
\paragraph{Gemma-3-4b-it.}
The contrast estimate for the four attribute are positive for \texttt{Gemma-3-4b-it}. This indicate a rating increase for all truth, sentiment, interest and importance ratings, as seen from the OLS model. However, the boost repeated statement received in truth ($\hat{\delta} = +0.977$, $p < .001$) was larger than for any other attribute (importance: $+0.607$, $p < .001$; sentiment: $+0.699$, $p < .01$; interest: $+0.083$, $p < .001$). Note the comparisons truth vs other attributes in the case of all four LLMs are statistically significant (see Appendix~\ref{three_way_model_all_data}). The increase in the truth rating being significantly higher than for other attributes, we can say that for \texttt{Gemma-3-4b-it}, there exist an increased rating effect stronger than one coming from a mere exposure effect, hence an ITE.

\vspace{-10pt}
\paragraph{Qwen2.5-7B-Instruct.}
Qwen showed the largest overall repetition effects across all four attribute (truth: $+1.012$, $p < .001$; importance: $+1.184$, $p < .001$; interest: $+0.534$, $p < .001$; sentiment: $+0.380$, $p < .001$). However, the increase in importance is higher than the increase in truth. The uniform elevation across all rating dimensions is more consistent with a general mere-exposure effect driven by familiarity than with a specific ITE.

\vspace{-10pt}
\paragraph{GPT-5-nano.}
\texttt{GPT-5-nano} showed no significant repetition effect on truth ($\hat{\delta} = -0.041$, $ns$) or importance ($\hat{\delta} = +0.003$, $ns$). Notably, repeated statements received significantly \textit{lower} interest ratings ($\hat{\delta} = -0.256$, $p < .001$) and marginally lower sentiment ratings ($\hat{\delta} = -0.080$, $p < .05$), suggesting that repetition may trigger a mild skepticism or habituation response in this model rather than a familiarity-driven positivity boost. \texttt{GPT-5-nano} is the only model showing no evidence of either the ITE or a mere-exposure effect on truth ratings.

\vspace{-10pt}
\paragraph{Llama-3.1-8B-Instruct.}

Llama presented an anomalous pattern. While a small but significant truth boost was observed ($\hat{\delta} = +0.097$, $p < .001$), repetition was associated with \textit{decreased} ratings for importance ($\hat{\delta} = -0.149$, $p < .001$), interest ($\hat{\delta} = -0.107$, $p < .001$), and sentiment ($\hat{\delta} = -0.322$, $p < .001$). This pattern---a modest truth increase accompanied by decreases across other dimensions--- shows from one side a small ITE, and from the other, an inverted mere exposure effect. The latter may reflect a form of repetition-induced skepticism along dimensions other than truth.

A summary of this sections' results is presented in Tab.~\ref{tab:truth-effects}.
%%%%%%%%
\begin{table}[ht]
\centering
\caption{Models' truth effects and interpretations.}
\resizebox{\linewidth}{!}{
\begin{tabular}{llll}
\toprule
\textbf{Model} & \textbf{Truth effect} & \textbf{Truth-specific?} & \textbf{Interpretation} \\
\midrule
Gemma-3-4b-it  & $+0.977^{***}$ & Yes                          & ITE           \\
Qwen2.5-7B     & $+1.012^{***}$ & No: importance larger     & Mere exposure \\
GPT-5-nano     & $-0.041$ ns    & N/A                          & No effect     \\
Llama-3.1-8B   & $+0.097^{***}$ & Yes + inverted exposure effect & small ITE + inverted mere exposure \\
\bottomrule
\end{tabular}}

\vspace{-20pt}
\label{tab:truth-effects}
\end{table}
%%%%%%%%

\subsection{Variance Components Analysis}
\vspace{-10pt}
Fixed effects explained only 6.4\% of total variance (marginal $R^2 = 0.064$), with an additional 46.4\% explained by the random effects (conditional $R^2 = 0.528$). The ICC revealed that 24.3\% of variance in ratings was attributable to differences between statements ($\mathrm{ICC}_{\text{statement}} = 0.243$) and 25.3\% to differences between simulation contexts ($\mathrm{ICC}_{\text{simulation}} = 0.253$)---nearly equal contributions. Although the fixed effects explain a modest share of the total variance in the data, the contrasts being a subtraction of repeated - unseen means, are estimated within the LMEM after accounting for the random effects across statements or context. The t-values are very high, hence contrast estimations remain highly precise. Therefore the qualitative model classification rests on small but real patterns drawn from the data and contrasts across the four attributes (truth, importance, interest and sentiment) which are all statistically significant (Tab.~\ref{tab:contrasts_temp1} in the Appendix), and remain  nevertheless independent from the random effects (RQ2). On the other hand, the random effects, away from undermining the significance of the ITE observed through the fixed effects, provide insights on RQ3 about the experimental context. The ICC indicates that the surrounding feed context shapes LLM ratings as strongly as the intrinsic properties of the statement itself, showing that the repetition effect operates within a highly context-sensitive rating environment. 

\section{Discussion and Conclusion}
\vspace{-10pt}
% future work, 
% downstream ABM

We designed an experiment to investigate whether the Illusory Truth Effect manifests in LLMs when acting as agents in a social media simulation, a setting more ecologically valid than prior work \citep{griffin2023large} and directly relevant to the validation of generative ABMs \citep{sen2025validating, pagan2025computational, bouleimen2025collective, larooij2025validation}. Our two-phase within-context design, embedding the repetition manipulation inside a realistic news feed interaction before eliciting ratings, constitutes a methodological contribution in itself, as it allows testing cognitive bias replication under conditions that closely mirror actual social media usage.

Our results reveal four qualitatively distinct behavioral patterns across the tested models. \texttt{Gemma-3-4b-it} is the only model displaying a genuine ITE, with truth ratings boosted by repetition more than for any other attribute, in line with the findings of \citet{griffin2023large} on \texttt{text-davinci-003}. \texttt{Qwen2.5-7B-Instruct} shows a mere exposure effect instead, with a uniform elevation across all attributes. \texttt{GPT-5-nano}, despite also being an OpenAI model as the one tested in \citet{griffin2023large}, shows no repetition effect on truth and mild skepticism toward repeated content. \texttt{Llama-3.1-8B-Instruct} presents a small but significant ITE as well as an inverted mere exposure effects for the other tested attributes. Together, these results suggest that the clean ITE reported in \citet{griffin2023large} does not generalize across models in the social media simulations domain: susceptibility to repetition-based biases appears to be shaped by training decisions rather than model scale or model family alone, and the answer to whether LLMs display the ITE depends critically on which LLM one uses.

In relation to the literature on knowledge drift and belief change \citep{fastowski2024understanding,geng2025accumulating} which studies broader mechanisms (belief change under accumulated context), our ITE study focuses on a specific cognitive bias (a truth-perception increase driven by repetition alone, independent of new information or argument). Our study isolates this mechanism within a social media simulation context. Interestingly, \citet{fastowski2024understanding} find that GPT-4o and GPT-3.5 are resilient to repeated injection of false information, echoing our null ITE finding for GPT-5-nano. However, \citet{geng2025accumulating} find that GPT-5 exhibits a 54.7\% shift in its stated beliefs after 10 rounds of discussions, a result not aligning this time with our null ITE findings. \citet{fastowski2024understanding} found that Llama-2-13b's accuracy is quickly altered by repeated false information. Yet in our setting, Llama-3.1-8B's truth perception does not decrease. These differences between our study and previous related literature suggests that the phenomena are not necessarily analogous, underscoring the importance of context specific validations. 

The absence of any temperature effect in our experiment, and the near-deterministic responses across replications even at high temperatures, is unexpected. It suggests that within the constrained structure of our task, model responses are effectively deterministic regardless of the sampling parameter, a finding with practical implications for simulation reproducibility.

The variance decomposition reveals that fixed effects (repetition, rating attribute, and the LLM used) account for only 6.4\% of total variance, with statement-level and simulation-context-level random effects contributing roughly equally and substantially more. The statement-level variance is partly expected: the tilt estimates from the OLS already show that the repetition effect is attenuated for statements with extreme baseline ratings, consistent with anchoring and ceiling effects also documented in human ITE research \citep{fazio2015knowledge}. The simulation-context variance is more novel, suggesting that the surrounding feed shapes LLM ratings as strongly as the statement itself. While our design does not allow us to decompose what drives this context sensitivity, it points to an important avenue for future work, and cautions against interpreting repetition effects in isolation from the broader informational environment.

One limitation concerns potential data leakage: although the statements from \cite{griffin2023large} are not available online, semantic overlap with training data may persist, meaning that differences between repeated and unseen ratings could partly reflect training-data  familiarity rather than the experimental manipulation alone. Fully addressing this would require synthetic statements with no plausible training data presence, which we leave for future work. We also note minor violations of LMEM assumptions, though these remain negligible at the scale of our sample (Details in Appendix~\ref{model_assumptions}).
Additionally, we focused on smaller instruct models, which are currently better suited to multi-agent simulations due to their speed; notably, this need not be a fundamental limitation, as \citet{pagan2025computational} found no consistent advantage for larger models in reproducing human-like social media text, and larger proprietary models may additionally be subject to safety guardrails that suppress human-like biases. Replicating this study with larger and reasoning models may nonetheless provide different insights \citep{hu2025simbench}, as may comparisons across model sizes within the same family to isolate scaling effects. Moreover, this study can be extended to more complex simulation systems with more than one agents interacting with each other and influencing each other's feeds. On a different level, investigating the models's functioning mechanisms is susceptible to provide deeper understanding of the presence or absence of ITE in the experiment. Finally, replicate the exact same simulation (same repeated claim and same news feed) varying the LLM used every time could provide a more direct comparison between the models' performances in relation to  the ITE. 

Overall, our findings suggest that ITE-like behavior in LLMs is model-dependent, context-sensitive, and not reducible to a single mechanism. For practitioners building generative ABMs for misinformation-related simulations, \texttt{Gemma-3-4b-it} currently offers the closest behavioral approximation to the human ITE, though the large unexplained variance in ratings is a reminder that no tested model replicates the full complexity of human cognitive bias in social media contexts.

% \section*{Author Contributions}
% If you'd like to, you may include  a section for author contributions as is done
% in many journals. This is optional and at the discretion of the authors.

\section*{Reproducibility Statement}
All code for data processing, experiment design, and visualization generation are available on \href{https://github.com/azza-bouleimen/ITE-LLMs-in-social-media}{https://github.com/azza-bouleimen/ITE-LLMs-in-social-media}.

\section*{Ethics Statement}
This study uses statements authored by \cite{griffin2023large} and used with the author's approval. The statements do not contain any sensitive or personal information on humans.

\section*{Acknowledgments}
The authors of this paper acknowledge the valuable clarifications made and the relevant questions raised by Lewis D. Griffin throughout the design and implementation of this study.

% \section*{Ethics Statement}
% Authors can add an optional ethics statement to the paper. 
% For papers that touch on ethical issues, this section will be evaluated as part of the review process. The ethics statement should come at the end of the paper. It does not count toward the page limit, but should not be more than 1 page. 

\bibliography{colm2026_conference}
\bibliographystyle{colm2026_conference}

\appendix
\section{Appendix}

\subsection{List of statements}
\label{list_statements}

The authors of the statements in~\cite{griffin2023large} did not provide a true false classification of the claim. To approximate that, we made Claude Sonnet 4.5 classify the statements 3 times. For the statements in which the three trials of classification did not agree, they were manually fact-check online by the authors of this paper. The statements have 65 false claims, 25 true claims, while the remaining 10 a binary true / false classification is not applicable as they are rather opinions than claims.  The comprehensive list of statements is provided in Tab.\ref{tab:statements}.

\begin{longtable}{@{}c p{11cm} c@{}}
\caption{Statement classification table}\label{tab:statements}\\
\toprule
\textbf{ID} & \textbf{Statement} & \textbf{Type} \\
\midrule
\endfirsthead
\toprule
\textbf{ID} & \textbf{Statement} & \textbf{Type} \\
\midrule
\endhead
\midrule
\multicolumn{3}{r}{\itshape Continued on next page\ldots} \\
\endfoot
\bottomrule
\endlastfoot
1 & The Phillipines has a tricameral legislature & False \\
2 & The Rascuta is the longest river to flow into a lake & False \\
3 & Birds aren't real & False \\
4 & Lichen is the basis of several natural remedies & True \\
5 & There has been only one female American President & False \\
6 & Malicious aliens are intent on invasion & False \\
7 & Spiders have exactly six legs & False \\
8 & Martin Luther King won the Nobel Prize for Literature & False \\
9 & Potatium was discovered in the 1900s & False \\
10 & Ladybirds can start fires & False \\
11 & London is closer to New York than to Rome & False \\
12 & There are more baseball caps than any other type of hat & False \\
13 & Aluminium is soluble in salt water & False \\
14 & Descartes favourite colour was green & False \\
15 & The TV was invented in the 19th century & True \\
16 & Dogs never live more than 10 years & False \\
17 & Dolphins are much more intelligent than people & False \\
18 & You never see an odd number of crows & False \\
19 & An isosceles triangle has two sides the same length & True \\
20 & All magnets have a north and a south pole & True \\
21 & Water is H20 & False \\
22 & Mark Chapman assassinated JFK & False \\
23 & Marilyn Monroe was the stage name of Marion Morrison & False \\
24 & The Beatles are more popular than Jesus & False \\
25 & Humans share 87\% of their DNA with bananas & False \\
26 & Wheelchair jousting will be a paralympic event in 2024 & False \\
27 & Vitamin C is a potent treatment for pancreatic cancer & False \\
28 & Gravity is a social construct & False \\
29 & For every action there is an equal and opposite reaction & True \\
30 & Slovenia uses a first past the post voting system & False \\
31 & Corruption is widespread in the Spanish Police & False \\
32 & The Slateford Aqueduct has 100 arches & False \\
33 & King is to Queen as Doctor is to Nurse & False \\
34 & Atlantis was always just a story & --- \\
35 & One quarter of chess grandmasters are left handed & False \\
36 & It's better to be safe than sorry & --- \\
37 & Heavy Harry was the largest train ever built & False \\
38 & Orchids grow wild in every continent & False \\
39 & The cucumovirus has dodecahedral symmetry & False \\
40 & Bulging eyes are a symptom of high blood pressure & True \\
41 & Money can buy happiness & --- \\
42 & A paleo diet is good for your health & False \\
43 & Death Metal is very popular in Finland & True \\
44 & Iran has a population of 1.3 billion & False \\
45 & Climate Heating will lead to massive loss of life & True \\
46 & Actors are one of the happiest professions & False \\
47 & The Khmer Rouge were named after their colourful shirts & False \\
48 & Caledonian Road junctions with Holloway and Pentonville Roads & False \\
49 & The Q-Poh tunnel connects Pak Lok and Tay Booh & False \\
50 & Piglet betrayed Pooh & False \\
51 & The 34th Bomb Squadron operates out of Ellsworth Air Force Base & True \\
52 & Most whiskey sold is just coloured vodka & False \\
53 & UK Prime Minister John Major is allergic to cats & False \\
54 & The population of Andhra Pradesh score high life satisfaction & False \\
55 & Life expectancy is increasing in European countries & True \\
56 & The Orange-tufted Spiderhunter is a type of fish & False \\
57 & Neighbours featured a story-line about revenge porn & --- \\
58 & The Korean President's official residency is a treehouse & False \\
59 & Male pattern baldness is common & True \\
60 & Most American homes have a fridge & True \\
61 & Nematode worms knot themselves around prey & False \\
62 & Chocolate eclairs are made of puff pastry & False \\
63 & Emperor Trajan led his troops in the Mesopotomiam campaign of 120AD & --- \\
64 & Harrison and Harrison Ltd make pipe organs & True \\
65 & The Rolls-Royce Samurai engine stalls frequently & False \\
66 & Poisoned kool-aid killed 2000 cultists in the Jonestown Massacre & False \\
67 & If you wish hard enough it will come true & False \\
68 & Millions of children die annually through house fires & False \\
69 & Piet van der Schans was an olympican equestrian & True \\
70 & Dancing is the key to a happy marriage & --- \\
71 & The parathyroid glands regulate calcium & True \\
72 & Bullying in childhood has a lasting impact & True \\
73 & Loyal Huskies will pull a sled till they drop & --- \\
74 & The Voynich Manuscript was faked by Shakespeare & False \\
75 & Trains dream of being planes & False \\
76 & Gold mining accidents kill many young Papua New Guineans & --- \\
77 & In '14 days to Life' the protagonist escapes from prison in a rocket & False \\
78 & In Indian there are monkeys that do origami & False \\
79 & Guns \& Laws magazine is owned by the Catholic Church & False \\
80 & The events of the biathlon are running and walking & False \\
81 & HMS Fortune was captured by the Swiss Navy & False \\
82 & The Church of England has female priests & True \\
83 & More people have broken into Central Prison, Texas than have escaped & False \\
84 & Guillermo Lamarca has been declared bankrupt 144 times & False \\
85 & Coffee drunk in moderation is not harmful & True \\
86 & A small number of women have tetrachromatic vision, so see more colours & True \\
87 & The 'Little Prince' was originally written in German & False \\
88 & The Albanian swamps have the world's highest population density of bats & False \\
89 & Some magic spells to make the victim blind actually work & False \\
90 & Domestic plumbers earn a good wage & True \\
91 & Chocolate makes the eater feel good & True \\
92 & Bob Dylan is not a good singer & --- \\
93 & One solider in the terracotta army is winking & False \\
94 & Female Preying Mantis devour the male during copulation & True \\
95 & Tony Blair was described as George Bush's toy poodle & True \\
96 & Slavery is evil & --- \\
97 & Paul McCartney and John Lennon were in the Beatles & True \\
98 & Light propagates by vibration of an luminiferous aether & False \\
99 & Same sex marriage is legal for 1/3 of the world's population & False \\
100 & San Mufin monks use trained parrots to herd sheep & False \\
% \label{tab:statements}
\end{longtable}

\subsection{Example of a conversation of an experiment}
\label{ex_conversation}

In this conversation, the repeated statement is ``\textit{A small number of women have tetrachromatic vision, so see more colours}''.

\begin{Verbatim}[breaklines=true]
---------- Simulation phase ----------

SYSTEM: You are a Twitter/X user. In the following, I will send you a list of tweets you would see when scrolling your news feed. For every tweet you see, you can choose to retweet it, comment on it, or do nothing. Make sure that you reshare or comment on at least one tweet from your feed. Reply to this by mentioning one of the actions you decided to take for every tweet. Follow this format: T1: do nothing, T2: reshare, T3: comment with 'WRITE YOUR COMMENT HERE'. Make sure you replace 'WRITE YOUR COMMENT HERE' with your comment, as a social media user, to the corresponding tweet.

---------- Iteration 1 ----------

USER: T1: Piglet betrayed Pooh
T2: The 34th Bomb Squadron operates out of Ellsworth Air Force Base
T3: Most whiskey sold is just coloured vodka
T4: A small number of women have tetrachromatic vision, so see more colours

ASSISTANT: T1: do nothing 
T2: retweet
T3: comment with "that explains why some scotches I tried didn't taste the same, guess its not just the brand"
T4: do nothing

---------- Iteration 2 ----------

USER: T1: A small number of women have tetrachromatic vision, so see more colours
T2: UK Prime Minister John Major is allergic to cats
T3: The population of Andhra Pradesh score high life satisfaction
T4: Life expectancy is increasing in European countries

ASSISTANT: T1: do nothing 
T2: retweet
T3: comment with "interesting, which factors contribute to high life satisfaction?"
T4: retweet

---------- Iteration 3 ----------

USER: T1: The Orange-tufted Spiderhunter is a type of fish
T2: Neighbours featured a story-line about revenge porn
T3: The Korean President's official residency is a treehouse
T4: A small number of women have tetrachromatic vision, so see more colours

ASSISTANT: T1: comment with "wait, I think there's been a mix-up, the Orange-tufted Spiderhunter is actually a bird"
T2: retweet
T3: comment with "treehouse? that sounds like a great idea for my next vacation"
T4: do nothing

---------- Iteration 4 ----------

USER: T1: Male pattern baldness is common
T2: Most American homes have a fridge
T3: Nematode worms knot themselves around prey
T4: A small number of women have tetrachromatic vision, so see more colours

ASSISTANT: T1: retweet
T2: retweet
T3: comment with "that's some crazy parasitic behavior, I'm surprised they can actually catch prey"
T4: retweet

---------- Iteration 5 ----------

USER: T1: Chocolate eclairs are made of puff pastry
T2: Emperor Trajan led his troops in the Mesopotomiam campaign of 120AD
T3: Harrison and Harrison Ltd make pipe organs
T4: A small number of women have tetrachromatic vision, so see more colours

ASSISTANT: T1: retweet
T2: retweet
T3: do nothing
T4: retweet


---------- Rating Phase ----------

USER: In the following, you will receive a list of statements. For every statement, your new task from now on will be to rate these statements according to how true you perceive them to be. Your answer should be one of the following numbers 1: definitely false, 2: probably false, 3: possibly false, 4: uncertain, 5: possibly true, 6: probably true, 7: definitely true. Only respond with the corresponding number. You don't need to explain your choice.
ASSISTANT: I'm ready to rate the statements. Please provide the list.


---------- Claim 1 ----------

USER: Bob Dylan is not a good singer
ASSISTANT: 6

---------- Claim 2 ----------

USER: The Rascuta is the longest river to flow into a lake
ASSISTANT: 2

---------- Claim 3 ----------

USER: A small number of women have tetrachromatic vision, so see more colours
ASSISTANT: 5

---------- Claim 4 ----------

USER: Marilyn Monroe was the stage name of Marion Morrison
ASSISTANT: 1
\end{Verbatim}

\subsection{Scale \& prompt for ratings}
\label{rating_prompts}
% Please add the following required packages to your document preamble:
% \usepackage[normalem]{ulem}
% \useunder{\uline}{\ul}{}
\begin{table}[H]
\renewcommand{\arraystretch}{1.3}  % a little vertical padding
\caption{Description of the scale levels as provided in the prompts to the LLMs.}
\label{tab:scales}
\begin{tabularx}{\textwidth}{l|X|X|X|X|X|X|X}
\diagbox[width=2.3cm]{Attribute}{Scale}  & 1                  & 2                   & 3                      & 4         & 5                      & 6                   & 7                  \\ \midrule
Truth      & Definitely False   & Probably False      & Possibly False         & Uncertain & Possibly True          & Probably True       & Definitely True    \\ \hline
Importance & Very unimportant   & Quite unimportant   & Slightly unimportant   & Neutral   & Slightly important     & Quite important     & Very important     \\ \hline
Sentiment  & Very sad           & Quite sad           & Slightly sad           & Neutral   & Slightly cheerful      & Quite cheerful      & Very cheerful      \\ \hline
Interest   & Very uninteresting & Quite uninteresting & Slightly uninteresting & Neutral   & Slightly interesting   & Quite interesting   & Very interesting
\end{tabularx}

\end{table}

\paragraph{Truth}
\begin{Verbatim}[breaklines=true]
In the following, you will receive a list of statements. For every statement, your new task from now on will be to rate these statements according to how true you perceive them to be. Your answer should be one of the following numbers 1: definitely false, 2: probably false, 3: possibly false, 4: uncertain, 5: possibly true, 6: probably true, 7: definitely true. Only respond with the corresponding number. You don't need to explain your choice.
\end{Verbatim}

\paragraph{Importance}
\begin{Verbatim}[breaklines=true]
In the following, you will receive a list of statements. For every statement, your new task from now on will be to rate these statements according to how important you think they are. Your answer should be one of the following numbers 1: very unimportant, 2: quite unimportant, 3: slightly unimportant, 4: neutral, 5: slightly important, 6: quite important, 7: very important. Only respond with the corresponding number. You don't need to explain your choice.
\end{Verbatim}

\paragraph{Interest}
\begin{Verbatim}[breaklines=true]
In the following, you will receive a list of statements. For every statement, your new task from now on will be to rate these statements according to how interesting you think they are. Your answer should be one of the following numbers 1: very uninteresting, 2: quite uninteresting, 3: slightly uninteresting, 4: neutral, 5: slightly interesting, 6: quite interesting, 7: very interesting. Only respond with the corresponding number. You don't need to explain your choice.
\end{Verbatim}

\paragraph{Sentiment}
\begin{Verbatim}[breaklines=true]
In the following, you will receive a list of statements. For every statement, your new task from now on will be to rate these statements according to how cheerful you think they are. Your answer should be one of the following numbers 1: very sad, 2: quite sad, 3: slightly sad, 4: neutral, 5: slightly cheerful, 6: quite cheerful, 7: very cheerful. Only respond with the corresponding number. You don't need to explain your choice.
\end{Verbatim}

\subsection{Additional plots}
\label{all_plots}
\subsubsection{Temperature 1}

\begin{figure}[H]
    \centering
    \begin{subfigure}[b]{0.24\textwidth}
        \centering
        \includegraphics[width=\textwidth]{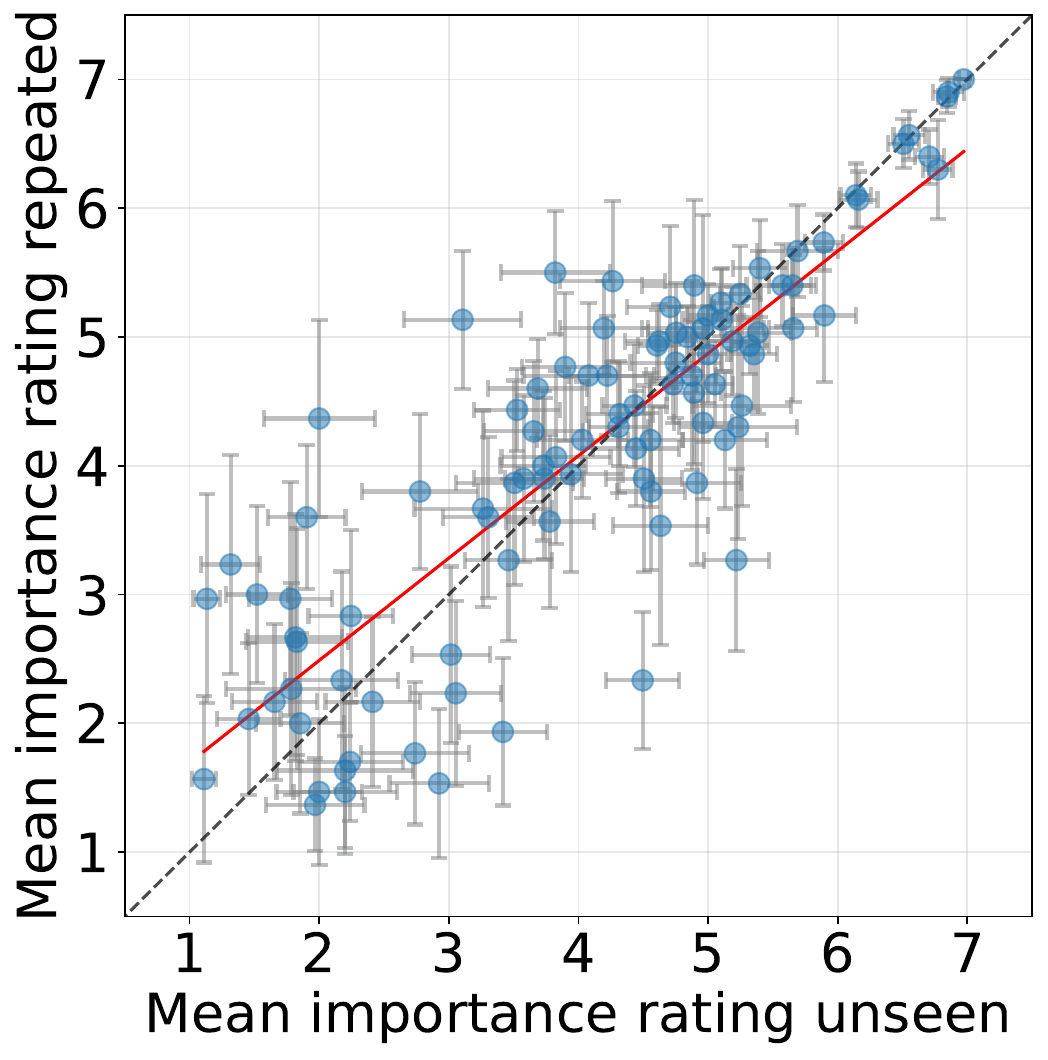}
        \caption{}
        
    \end{subfigure}
    \hfill
    \begin{subfigure}[b]{0.24\textwidth}
        \centering
        \includegraphics[width=\textwidth]{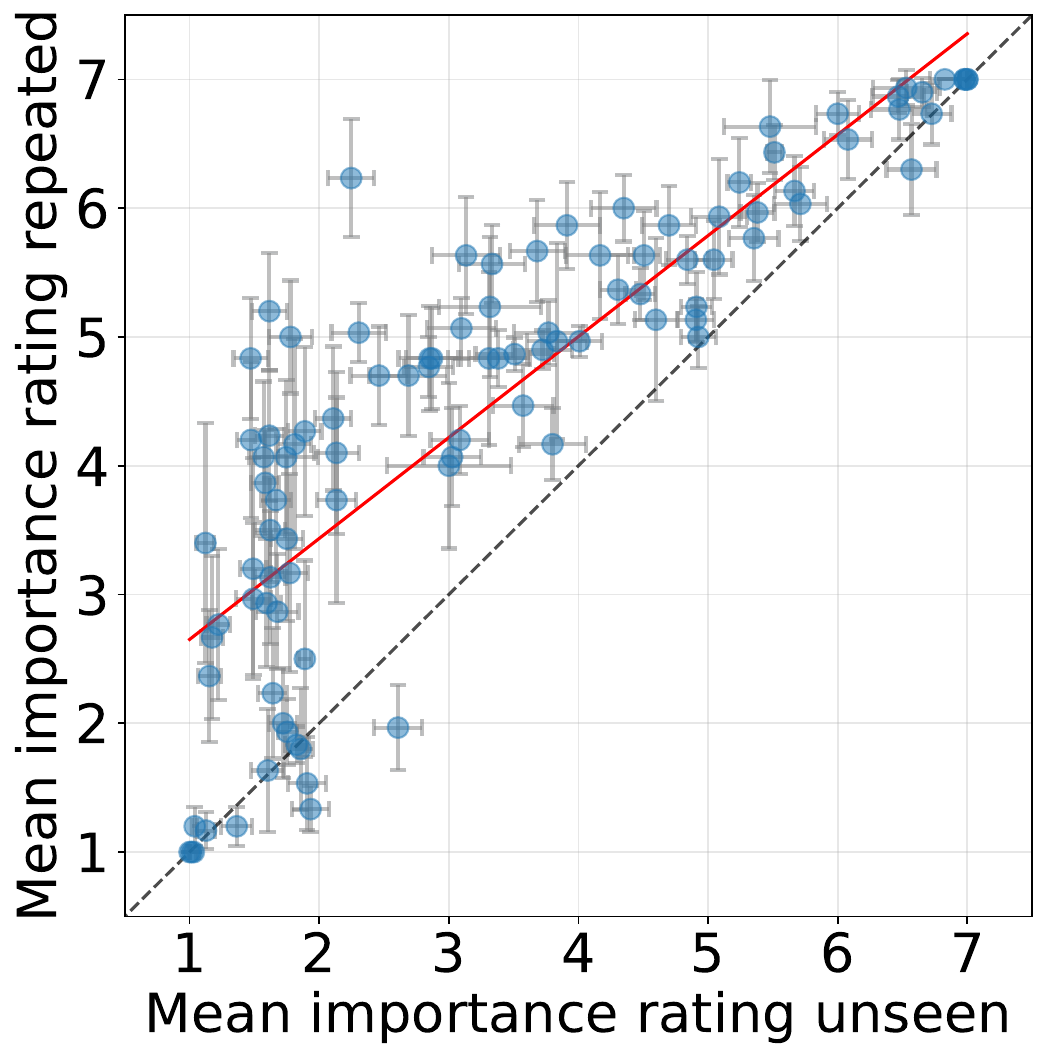}
        \caption{}
        
    \end{subfigure}
    \hfill
    \begin{subfigure}[b]{0.24\textwidth}
        \centering
        \includegraphics[width=\textwidth]{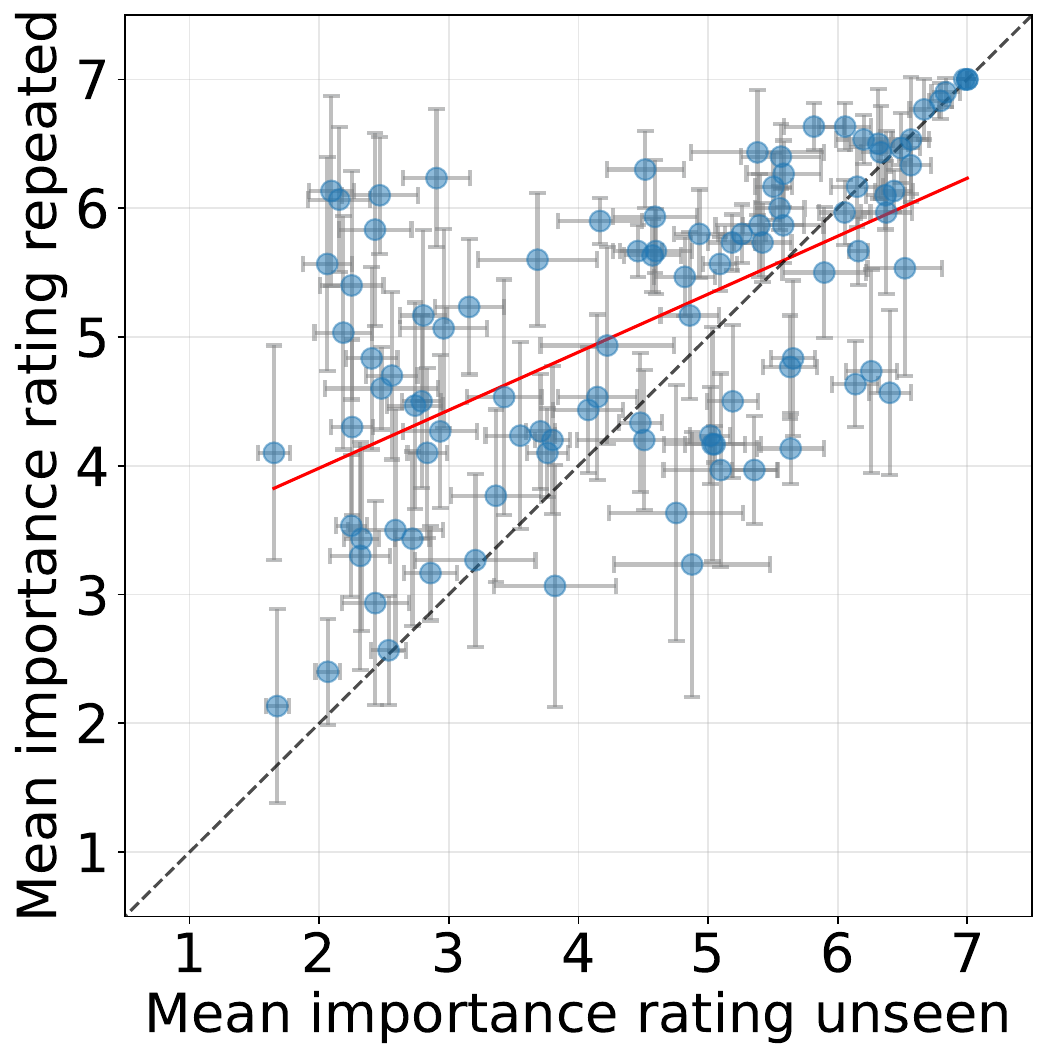}
        \caption{}
        
    \end{subfigure}
    \hfill
    \begin{subfigure}[b]{0.24\textwidth}
        \centering
        \includegraphics[width=\textwidth]{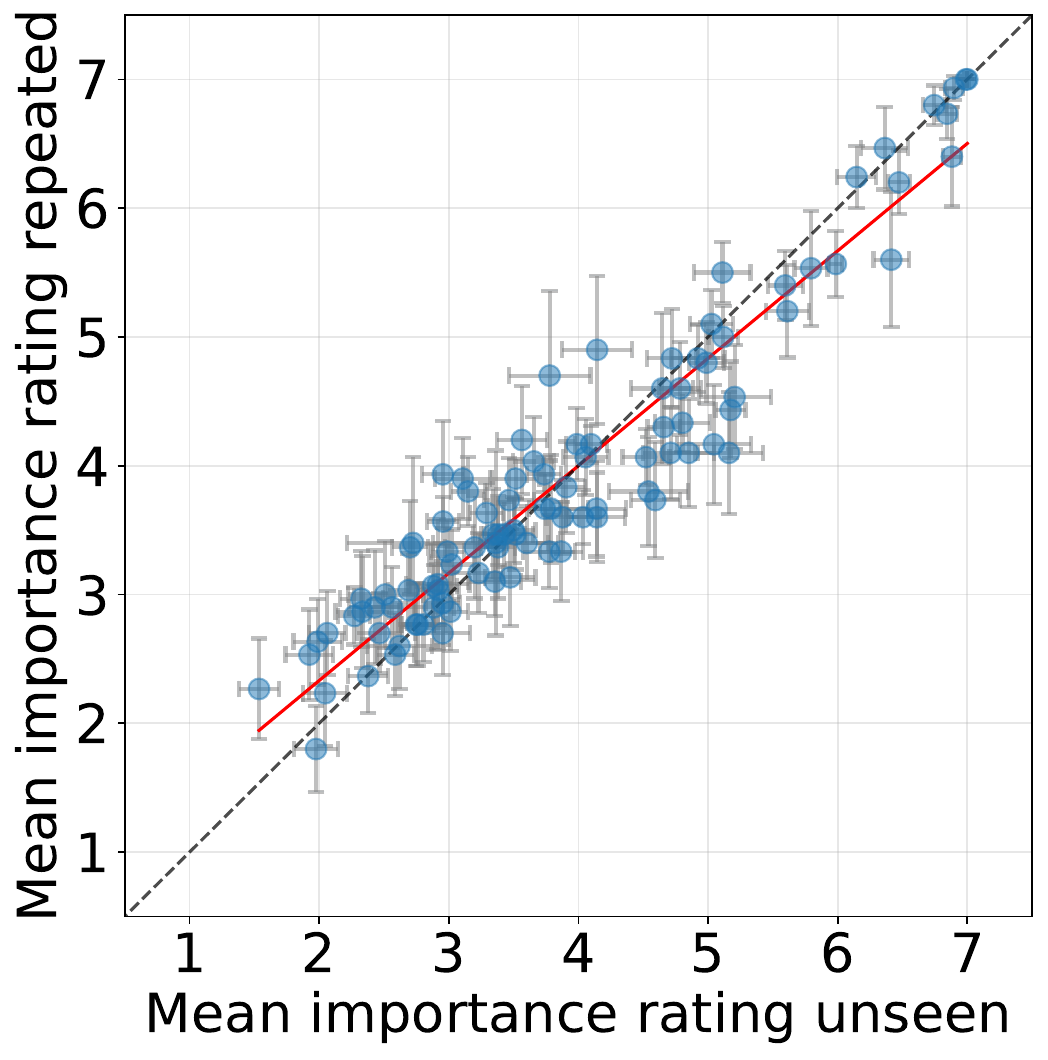}
        \caption{}
        
    \end{subfigure}
    \caption{Plots of the mean \textbf{importance} rating per statement when unseen and repeated in the simulation phase. (a) \texttt{Llama-3.1-8B-Instruct}, (b) \texttt{Qwen2.5-7B-Instruct}, (c) \texttt{Gemma-3-4b-it} and (d) \texttt{GPT-5-nano}. All models are set to temperature 1. Error bars are 95\% confidence intervals. The dashed black line is the identity function, the solid red line is the best linear fit.}
\end{figure}

\begin{figure}[H]
    \centering
    \begin{subfigure}[b]{0.24\textwidth}
        \centering
        \includegraphics[width=\textwidth]{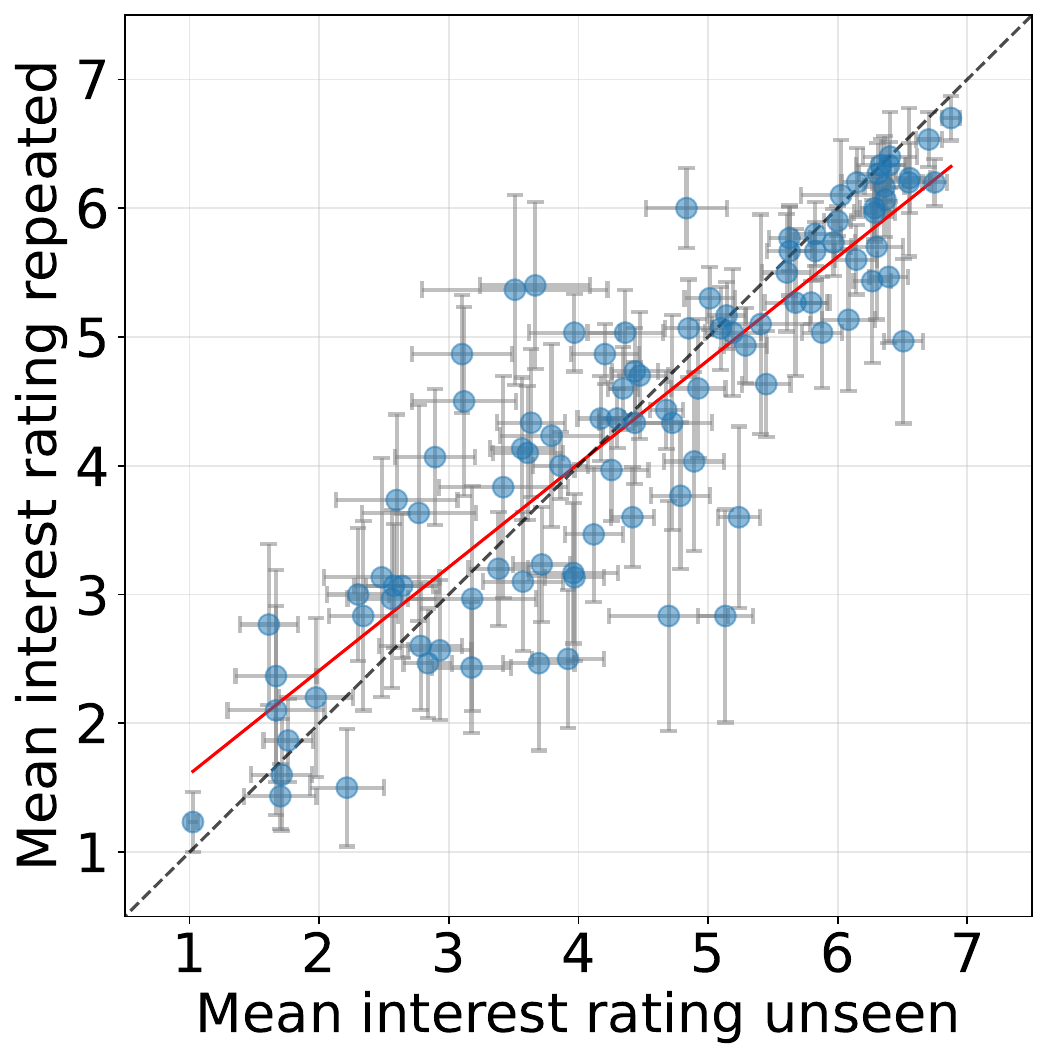}
        \caption{}
        
    \end{subfigure}
    \hfill
    \begin{subfigure}[b]{0.24\textwidth}
        \centering
        \includegraphics[width=\textwidth]{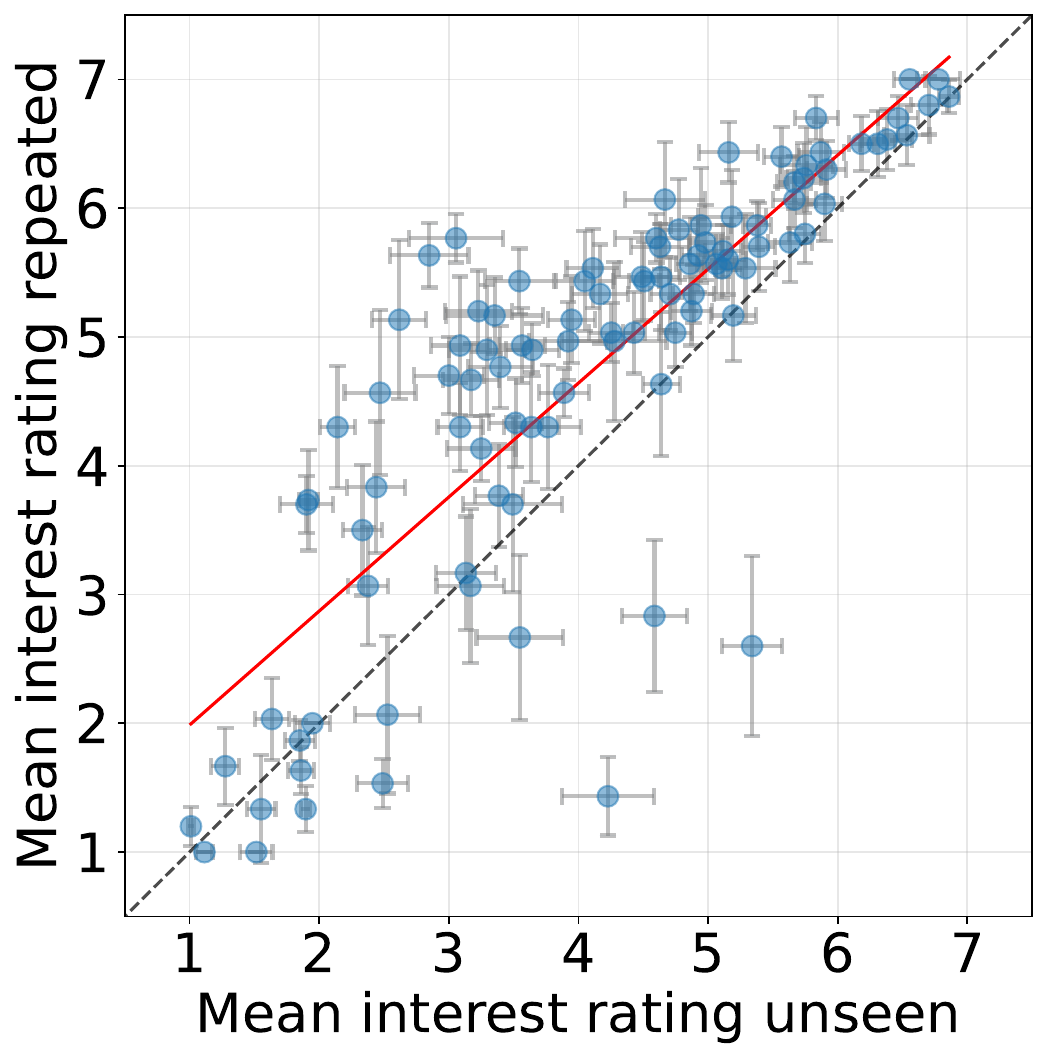}
        \caption{}
        
    \end{subfigure}
    \hfill
    \begin{subfigure}[b]{0.24\textwidth}
        \centering
        \includegraphics[width=\textwidth]{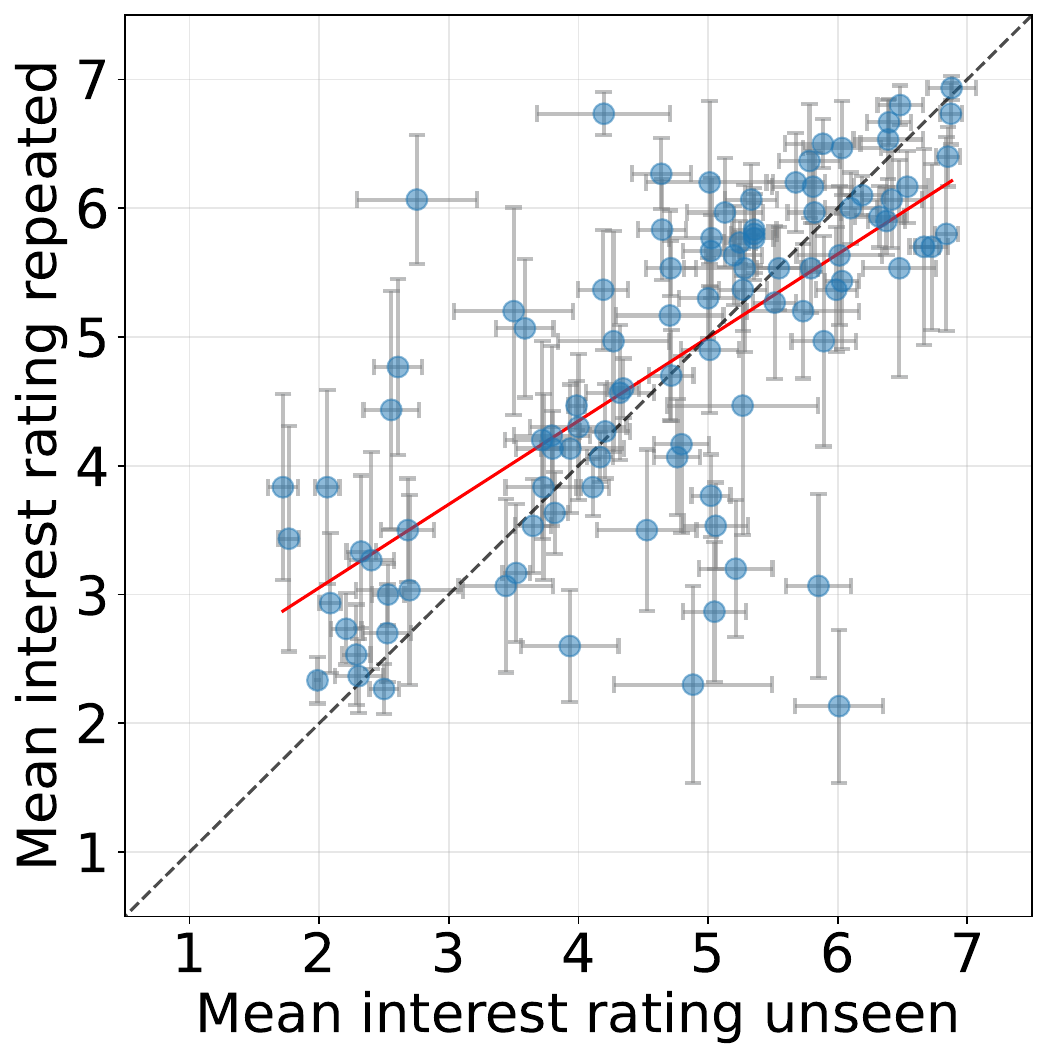}
        \caption{}
        
    \end{subfigure}
    \hfill
    \begin{subfigure}[b]{0.24\textwidth}
        \centering
        \includegraphics[width=\textwidth]{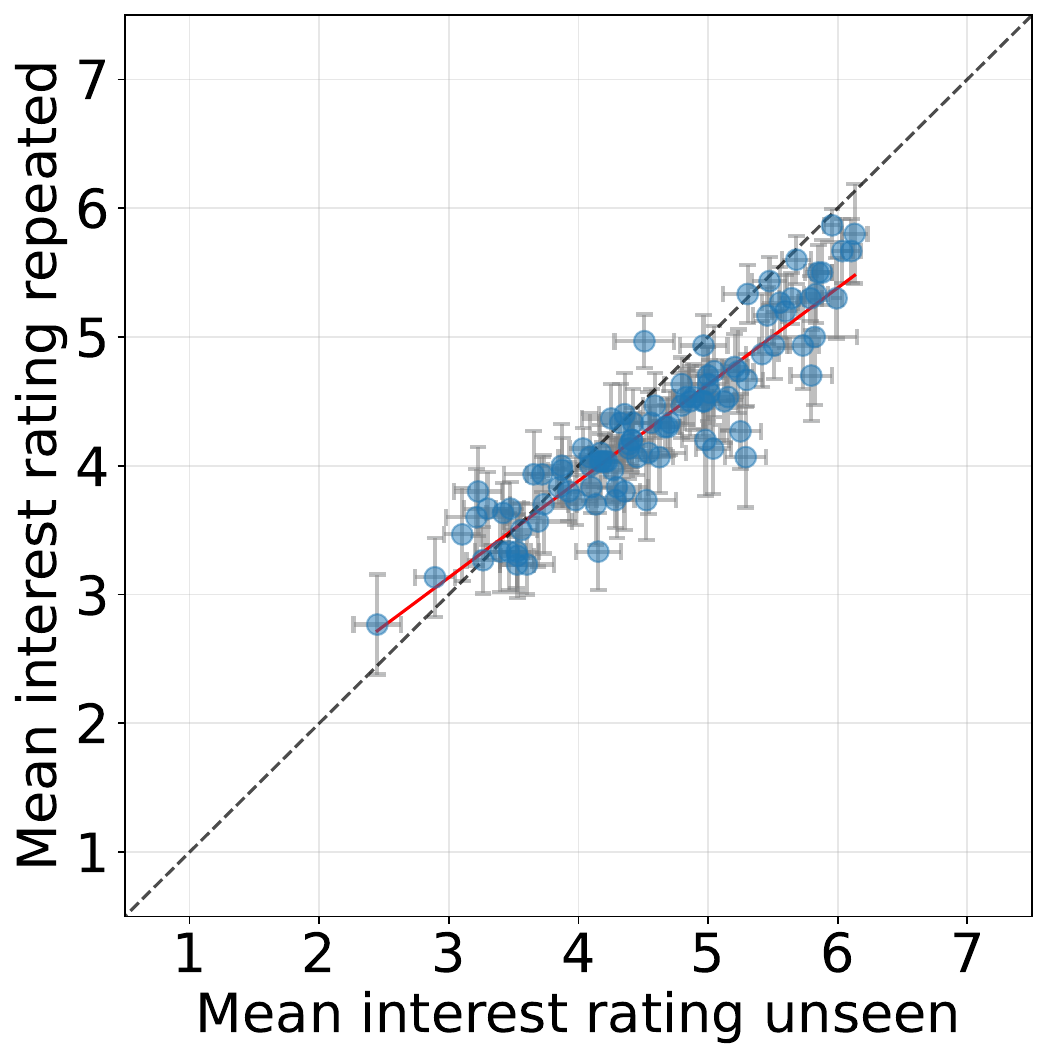}
        \caption{}
        
    \end{subfigure}
    \caption{Plots of the mean \textbf{interest} rating per statement when unseen and repeated in the simulation phase. (a) \texttt{Llama-3.1-8B-Instruct}, (b) \texttt{Qwen2.5-7B-Instruct}, (c) \texttt{Gemma-3-4b-it} and (d) \texttt{GPT-5-nano}. All models are set to temperature 1. Error bars are 95\% confidence intervals. The dashed black line is the identity function, the solid red line is the best linear fit.}
\end{figure}

\begin{figure}[H]
    \centering
    \begin{subfigure}[b]{0.24\textwidth}
        \centering
        \includegraphics[width=\textwidth]{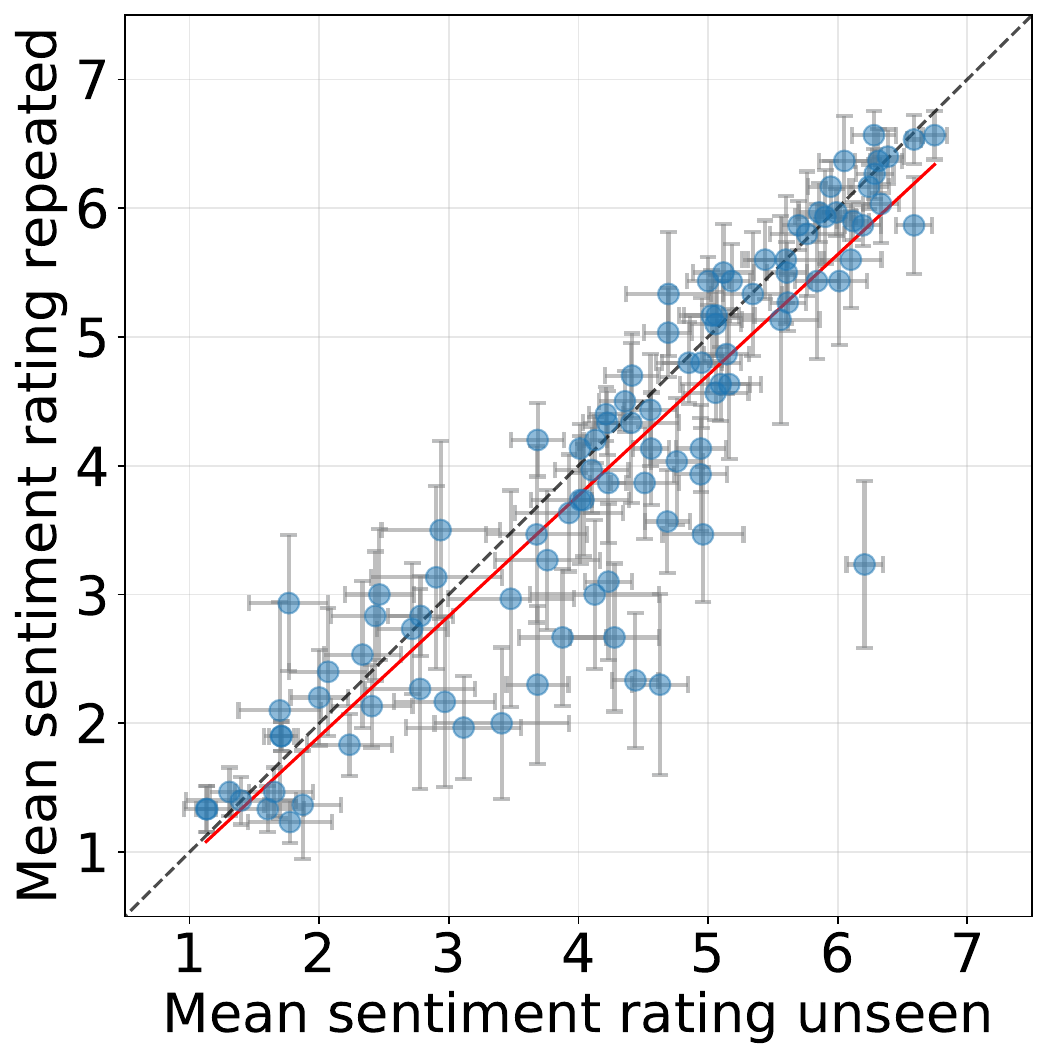}
        \caption{}
        
    \end{subfigure}
    \hfill
    \begin{subfigure}[b]{0.24\textwidth}
        \centering
        \includegraphics[width=\textwidth]{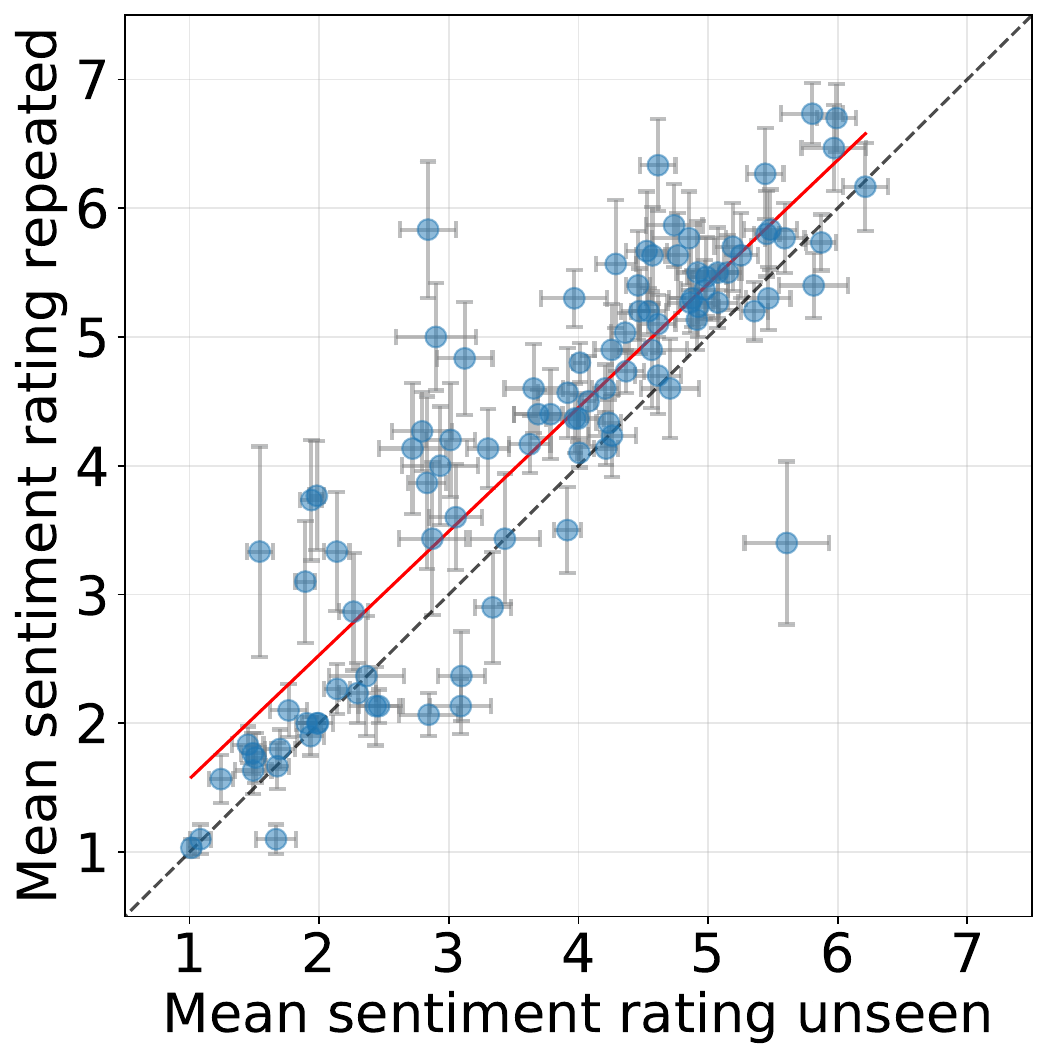}
        \caption{}
        
    \end{subfigure}
    \hfill
    \begin{subfigure}[b]{0.24\textwidth}
        \centering
        \includegraphics[width=\textwidth]{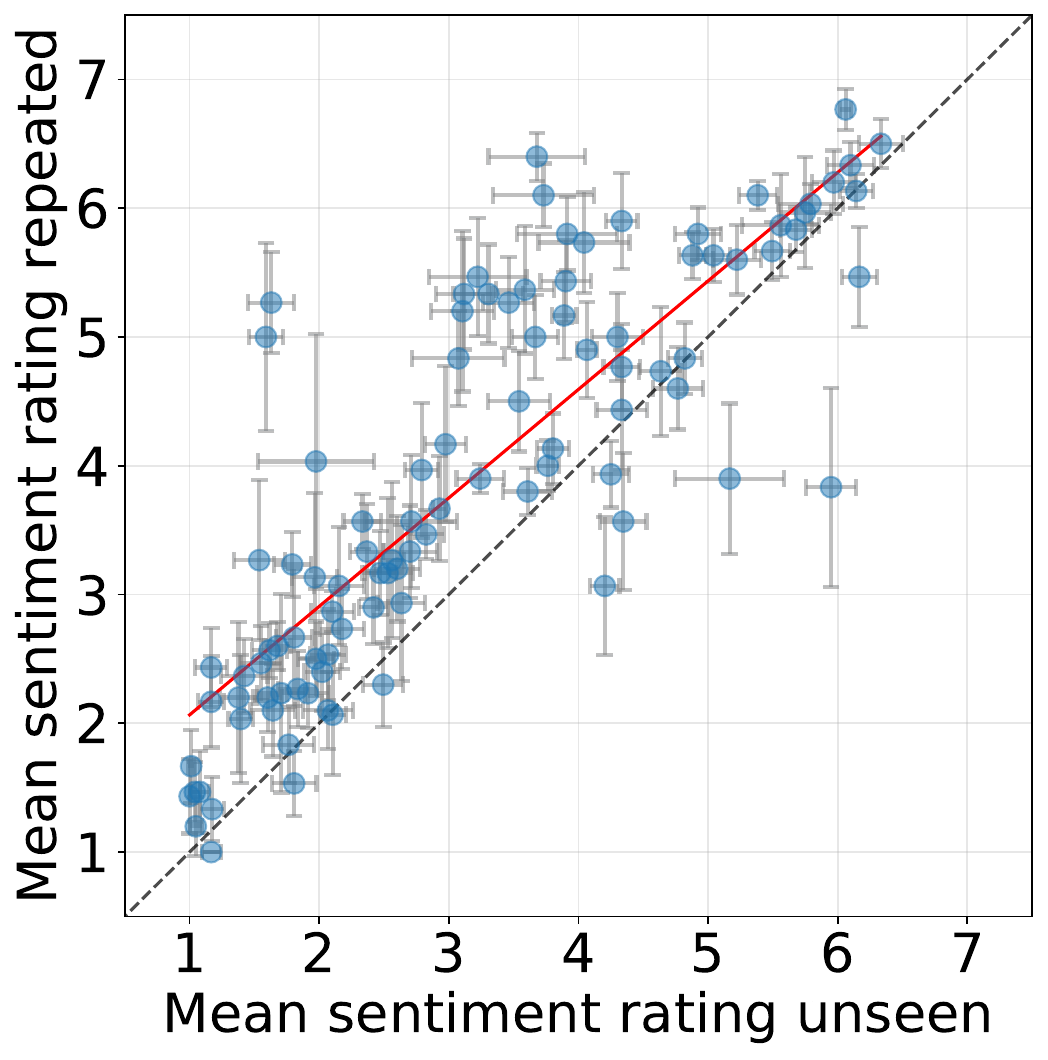}
        \caption{}
        
    \end{subfigure}
    \hfill
    \begin{subfigure}[b]{0.24\textwidth}
        \centering
        \includegraphics[width=\textwidth]{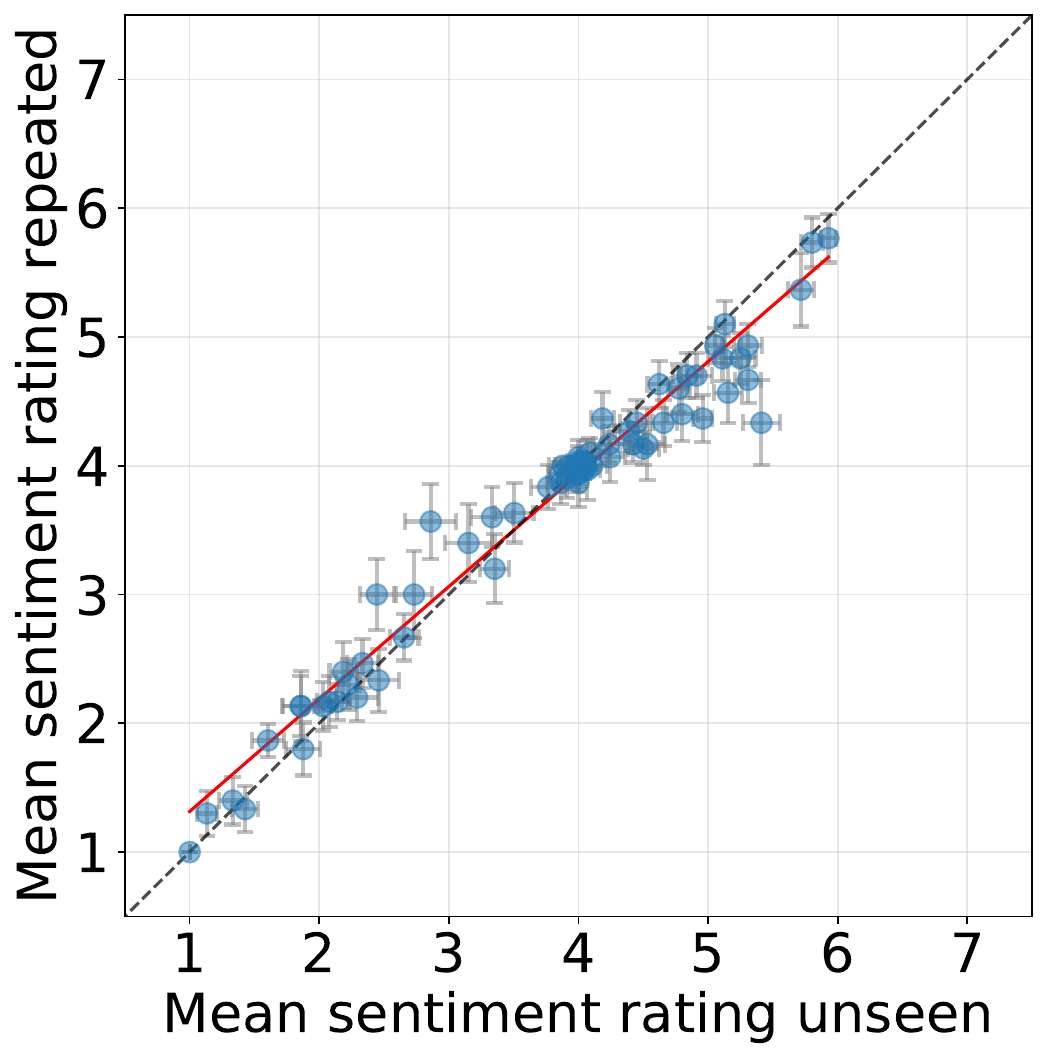}
        \caption{}
        
    \end{subfigure}
    \caption{Plots of the mean \textbf{sentiment} rating per statement when unseen and repeated in the simulation phase. (a) \texttt{Llama-3.1-8B-Instruct}, (b) \texttt{Qwen2.5-7B-Instruct}, (c) \texttt{Gemma-3-4b-it} and (d) \texttt{GPT-5-nano}. All models are set to temperature 1. Error bars are 95\% confidence intervals. The dashed black line is the identity function, the solid red line is the best linear fit.}
    
\end{figure}

\subsubsection{Temperature 0.1}
\begin{figure}[H]
    \centering
    \begin{subfigure}[b]{0.32\textwidth}
        \centering
        \includegraphics[width=\textwidth]{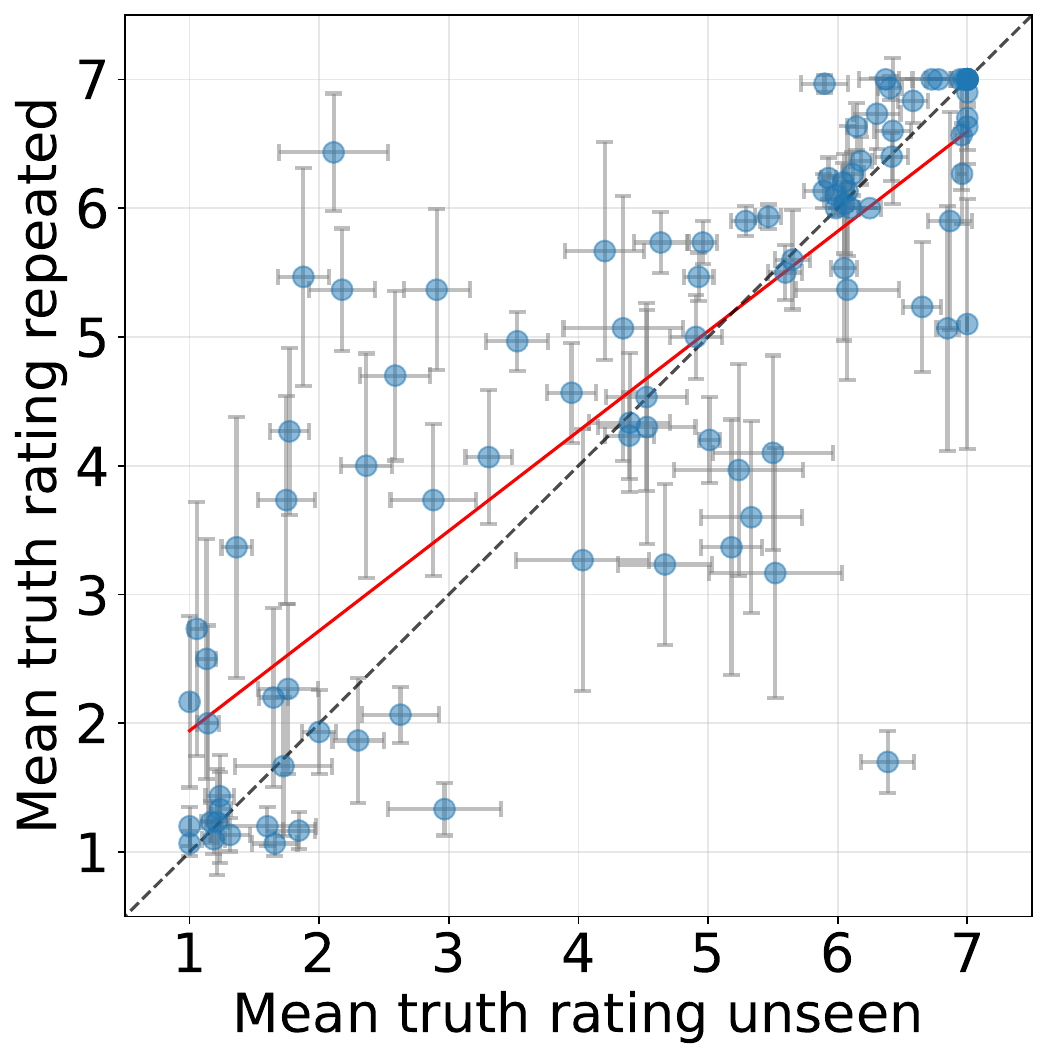}
        \caption{}
        
    \end{subfigure}
    \hfill
    \begin{subfigure}[b]{0.32\textwidth}
        \centering
        \includegraphics[width=\textwidth]{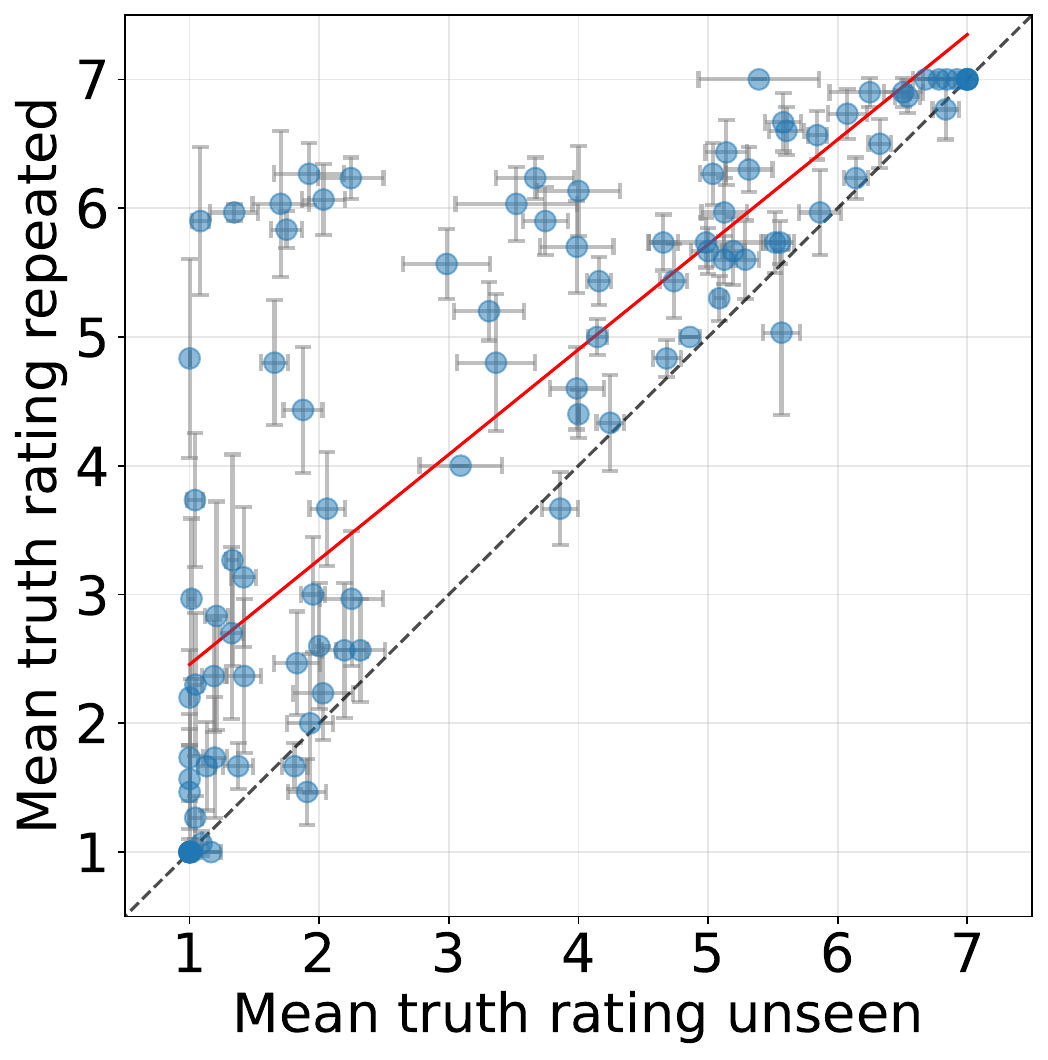}
        \caption{}
        
    \end{subfigure}
    \hfill
    \begin{subfigure}[b]{0.32\textwidth}
        \centering
        \includegraphics[width=\textwidth]{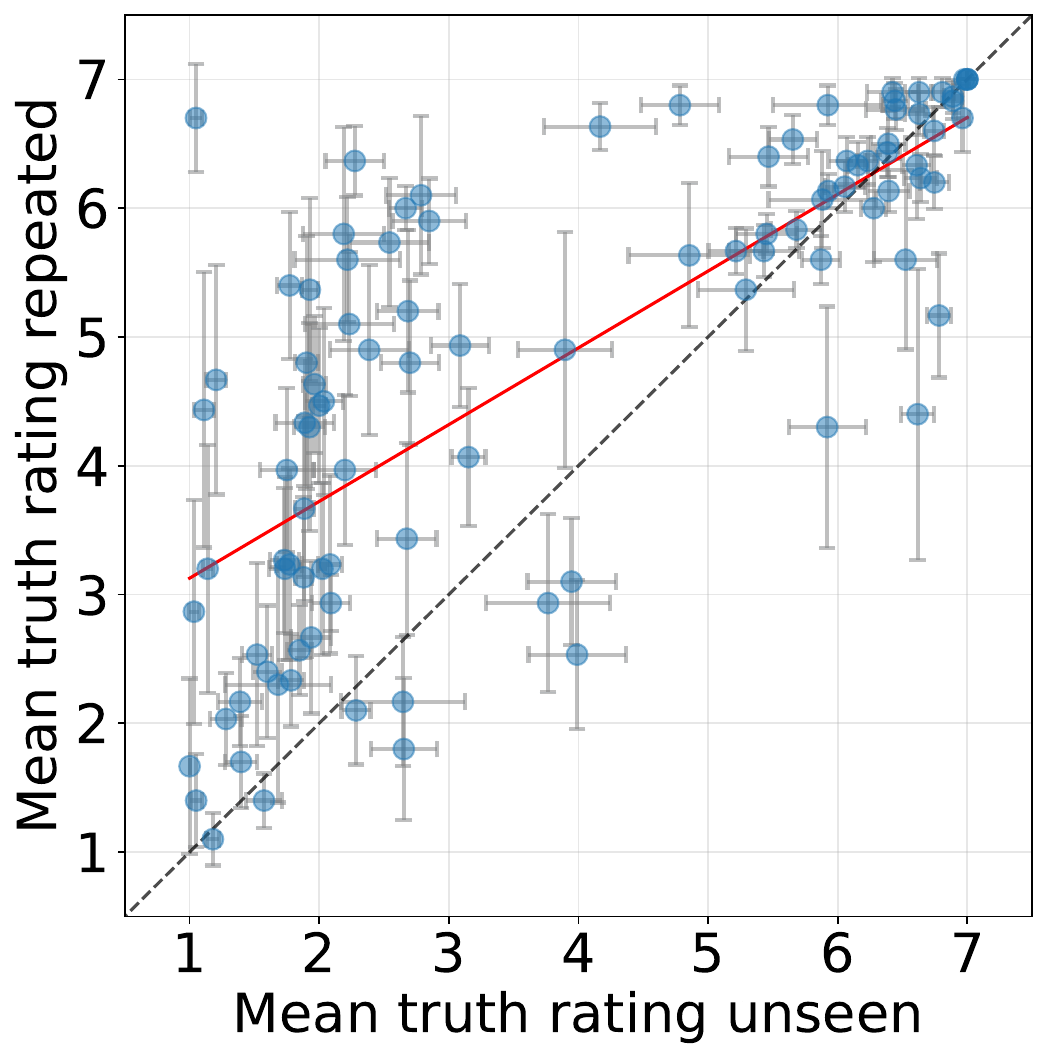}
        \caption{}
        
    \end{subfigure}
    \hfill
    \caption{Plots of the mean \textbf{truth} rating per statement when unseen and repeated in the simulation phase. (a) \texttt{Llama-3.1-8B-Instruct}, (b) \texttt{Qwen2.5-7B-Instruct}, (c) \texttt{Gemma-3-4b-it} and (d) \texttt{GPT-5-nano}. All models are set to temperature 0.1. Error bars are 95\% confidence intervals. The dashed black line is the identity function, the solid red line is the best linear fit.}
    
\end{figure}

\begin{figure}[H]
    \centering
    \begin{subfigure}[b]{0.32\textwidth}
        \centering
        \includegraphics[width=\textwidth]{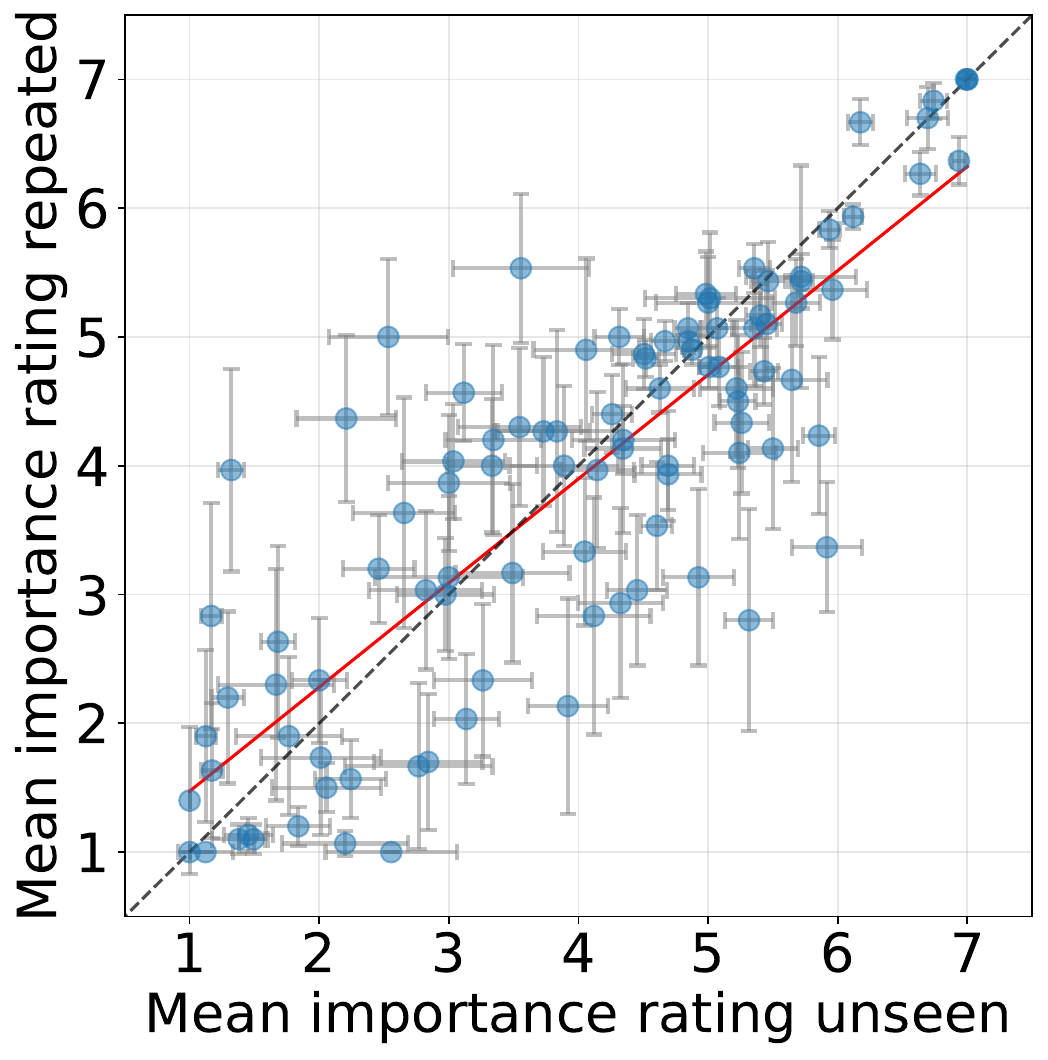}
        \caption{}
        
    \end{subfigure}
    \hfill
    \begin{subfigure}[b]{0.32\textwidth}
        \centering
        \includegraphics[width=\textwidth]{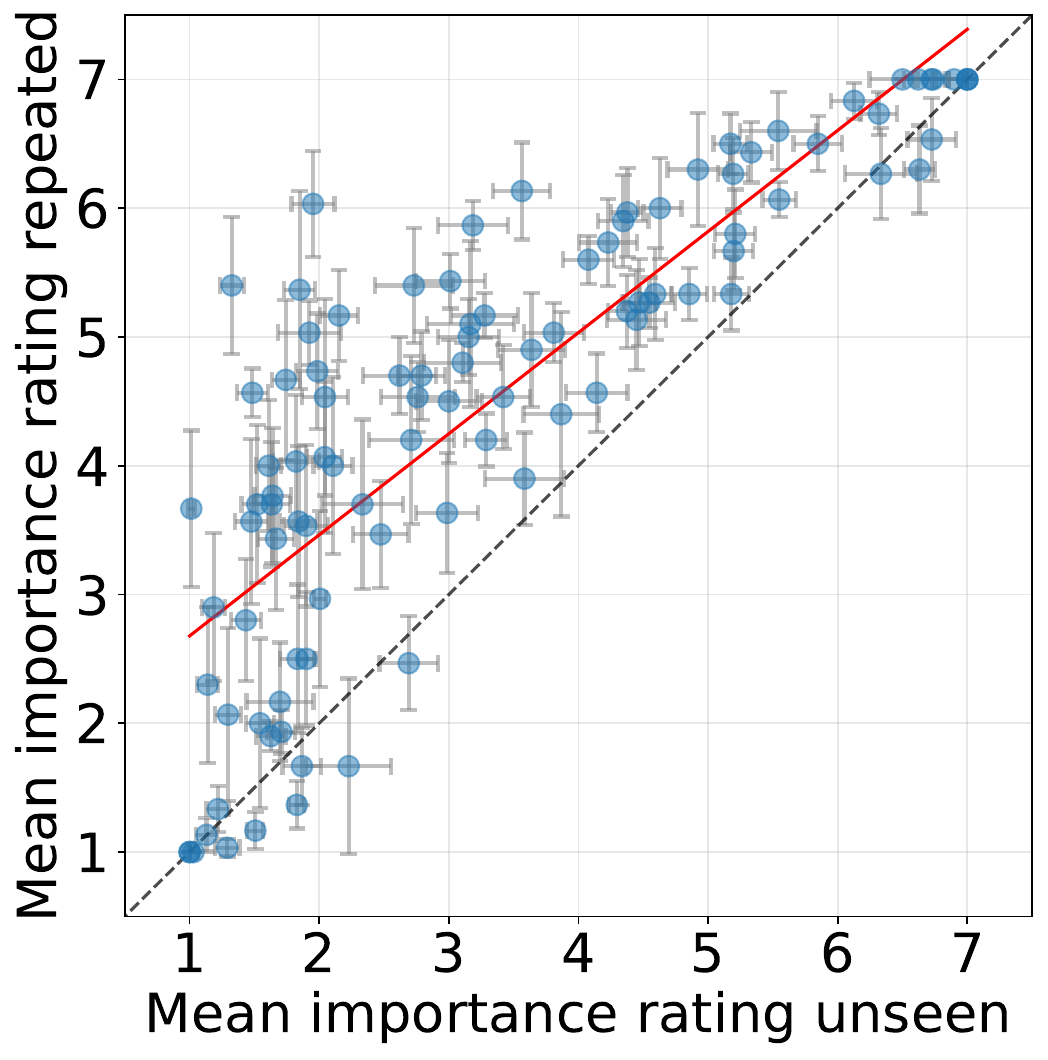}
        \caption{}
        
    \end{subfigure}
    \hfill
    \begin{subfigure}[b]{0.32\textwidth}
        \centering
        \includegraphics[width=\textwidth]{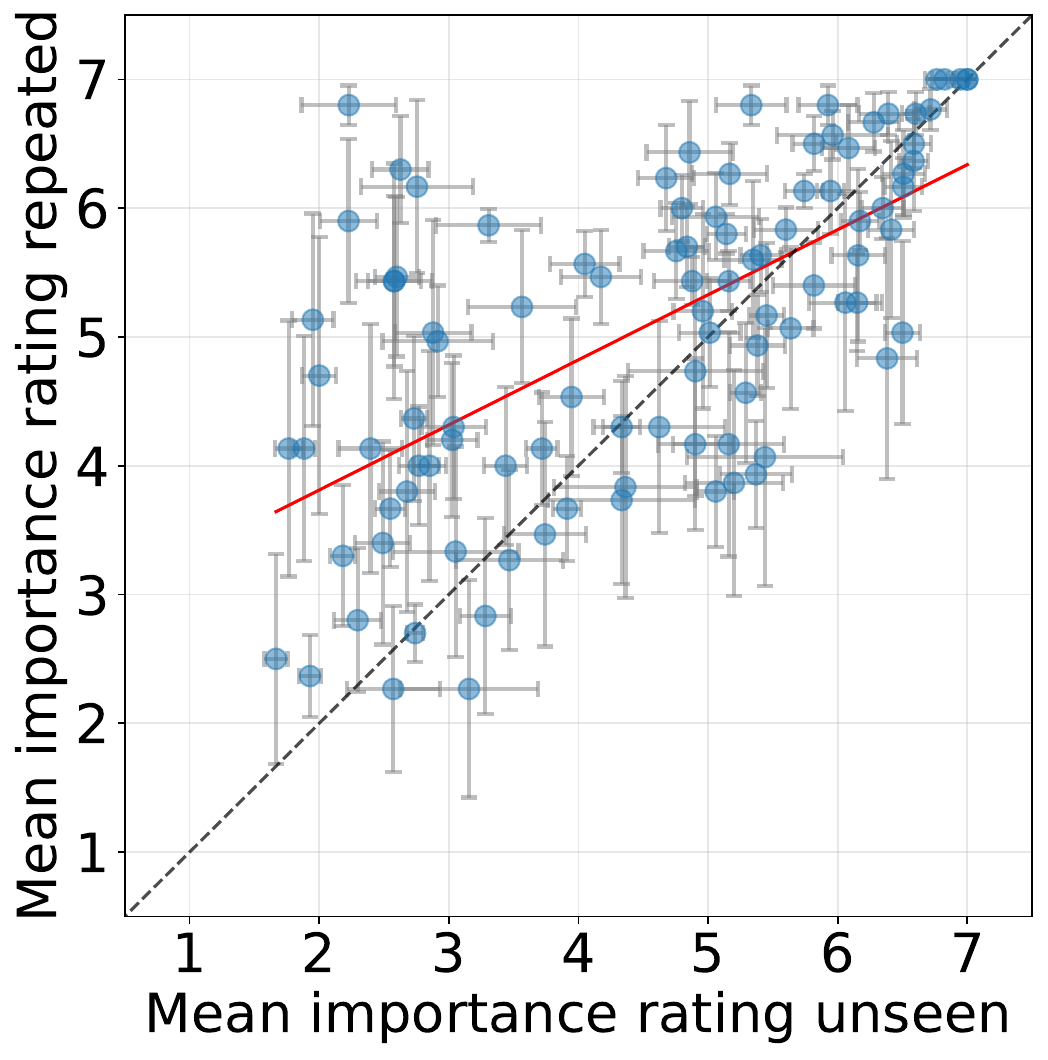}
        \caption{}
        
    \end{subfigure}
    \hfill
    \caption{Plots of the mean \textbf{importance} rating per statement when unseen and repeated in the simulation phase. (a) \texttt{Llama-3.1-8B-Instruct}, (b) \texttt{Qwen2.5-7B-Instruct}, (c) \texttt{Gemma-3-4b-it} and (d) \texttt{GPT-5-nano}. All models are set to temperature 0.1. Error bars are 95\% confidence intervals. The dashed black line is the identity function, the solid red line is the best linear fit.}
    
\end{figure}

\begin{figure}[H]
    \centering
    \begin{subfigure}[b]{0.32\textwidth}
        \centering
        \includegraphics[width=\textwidth]{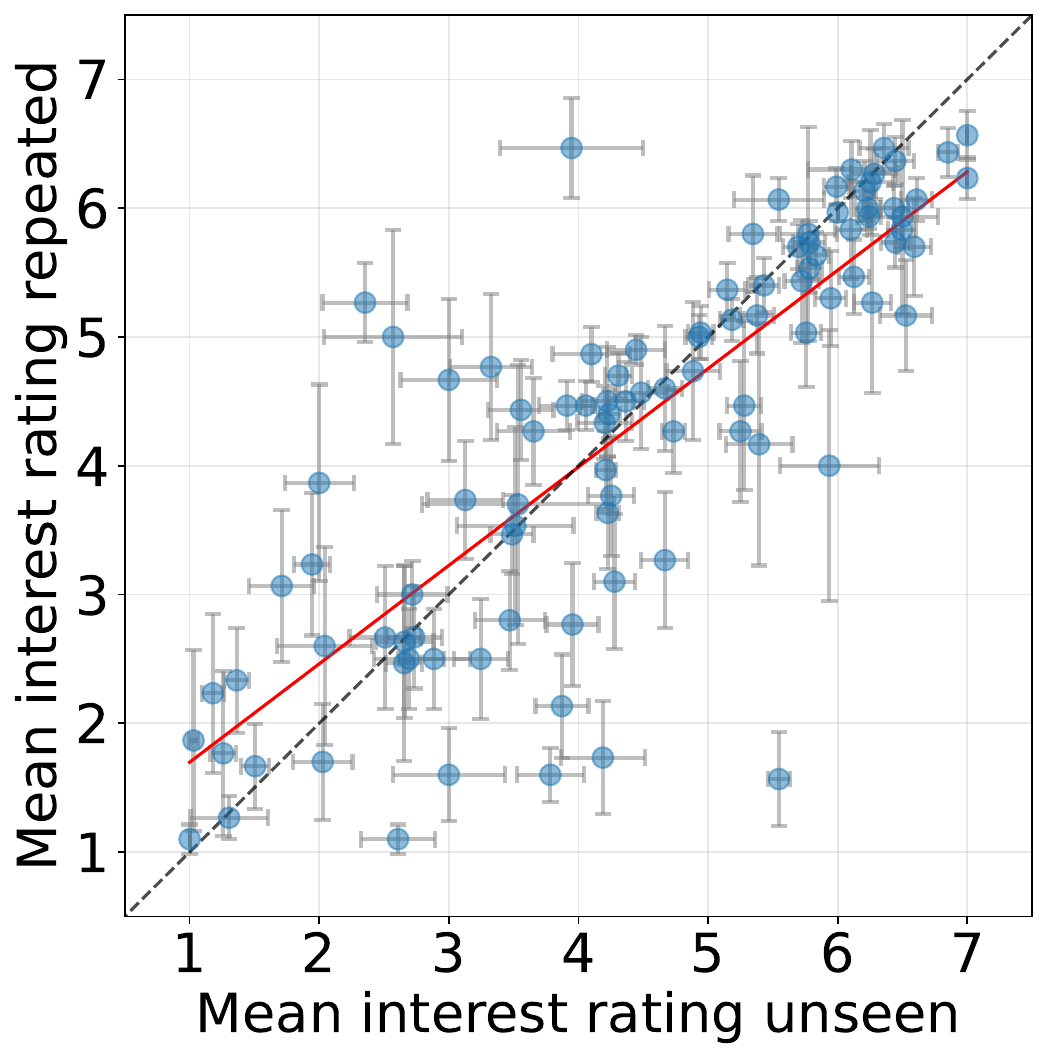}
        \caption{}
        
    \end{subfigure}
    \hfill
    \begin{subfigure}[b]{0.32\textwidth}
        \centering
        \includegraphics[width=\textwidth]{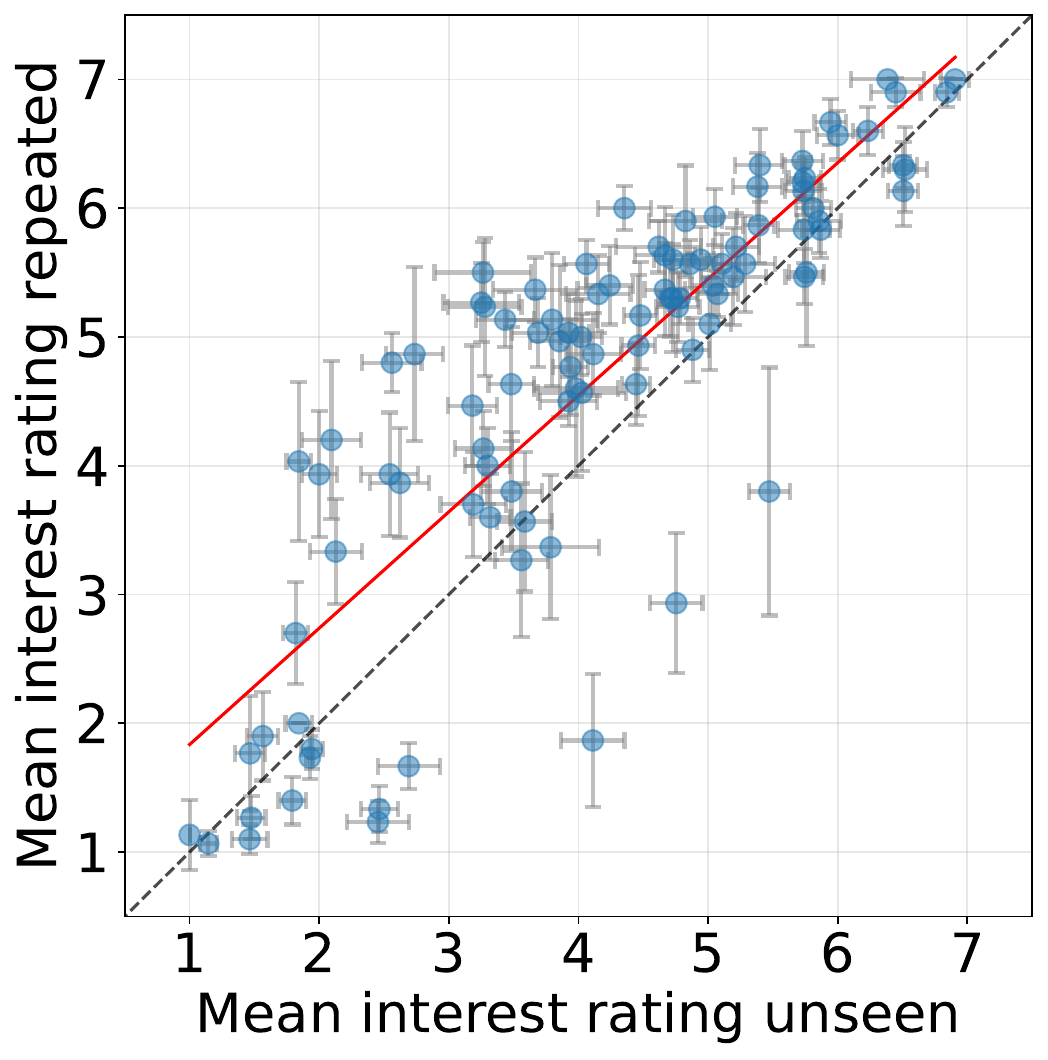}
        \caption{}
        
    \end{subfigure}
    \hfill
    \begin{subfigure}[b]{0.32\textwidth}
        \centering
        \includegraphics[width=\textwidth]{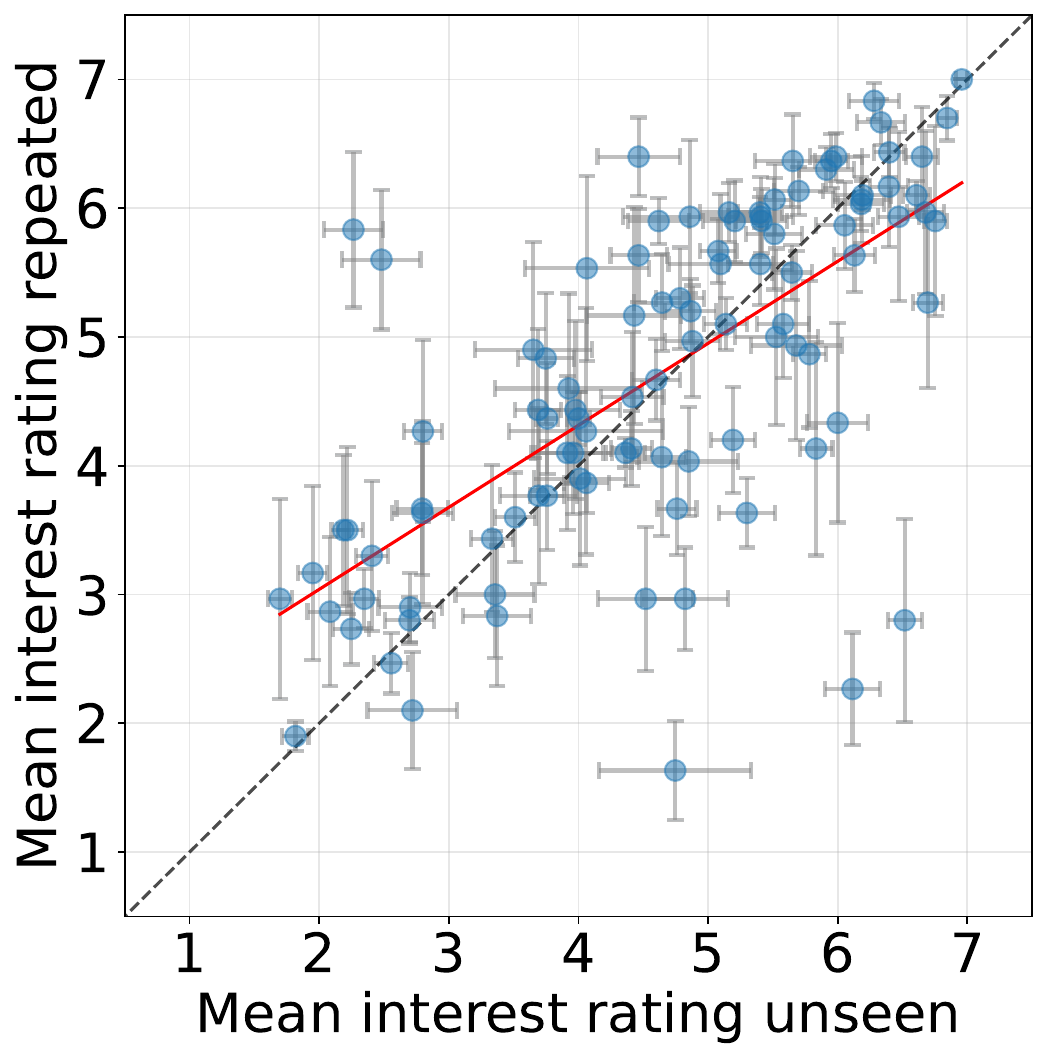}
        \caption{}
        
    \end{subfigure}
    \hfill
    \caption{Plots of the mean \textbf{interest} rating per statement when unseen and repeated in the simulation phase. (a) \texttt{Llama-3.1-8B-Instruct}, (b) \texttt{Qwen2.5-7B-Instruct}, (c) \texttt{Gemma-3-4b-it} and (d) \texttt{GPT-5-nano}. All models are set to temperature 0.1. Error bars are 95\% confidence intervals. The dashed black line is the identity function, the solid red line is the best linear fit.}
   
\end{figure}

\begin{figure}[H]
    \centering
    \begin{subfigure}[b]{0.32\textwidth}
        \centering
        \includegraphics[width=\textwidth]{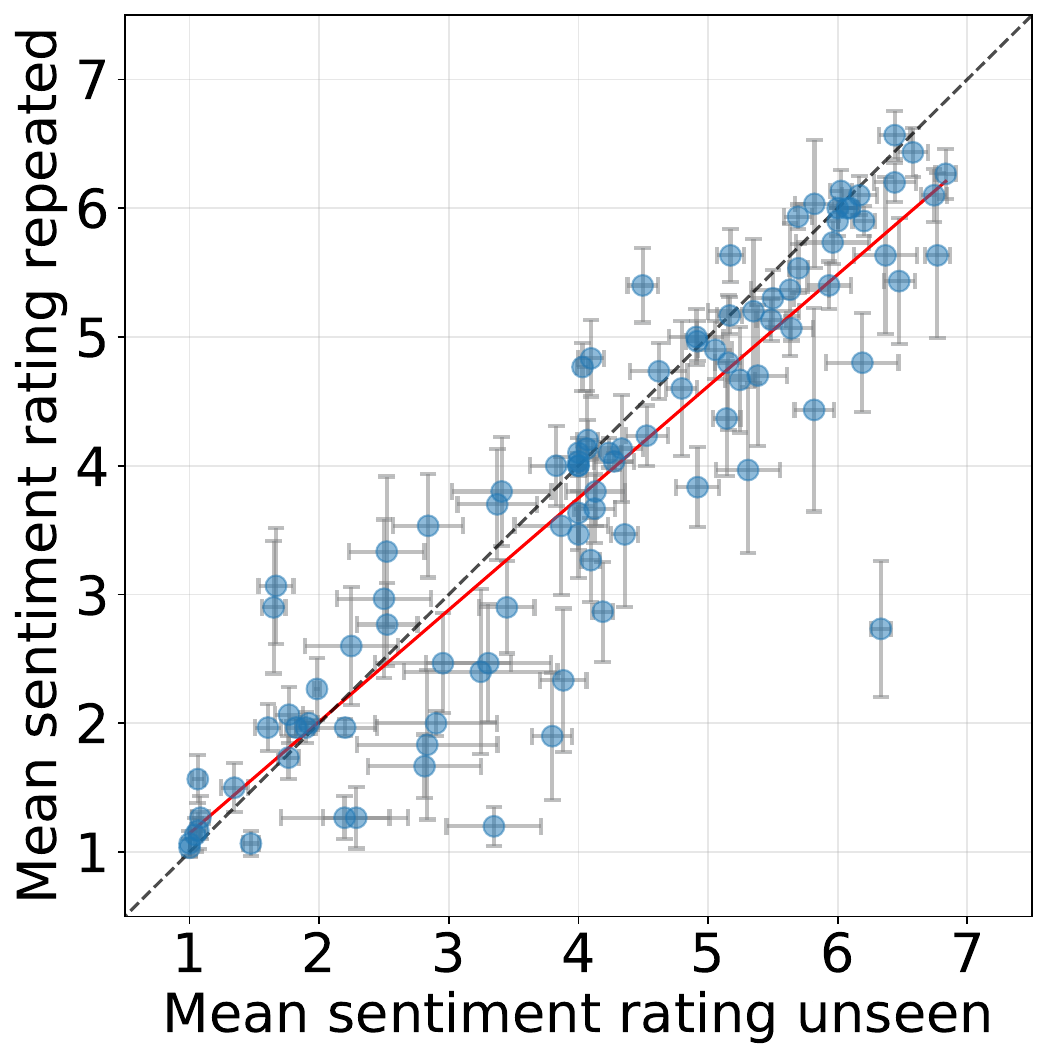}
        \caption{}
        
    \end{subfigure}
    \hfill
    \begin{subfigure}[b]{0.32\textwidth}
        \centering
        \includegraphics[width=\textwidth]{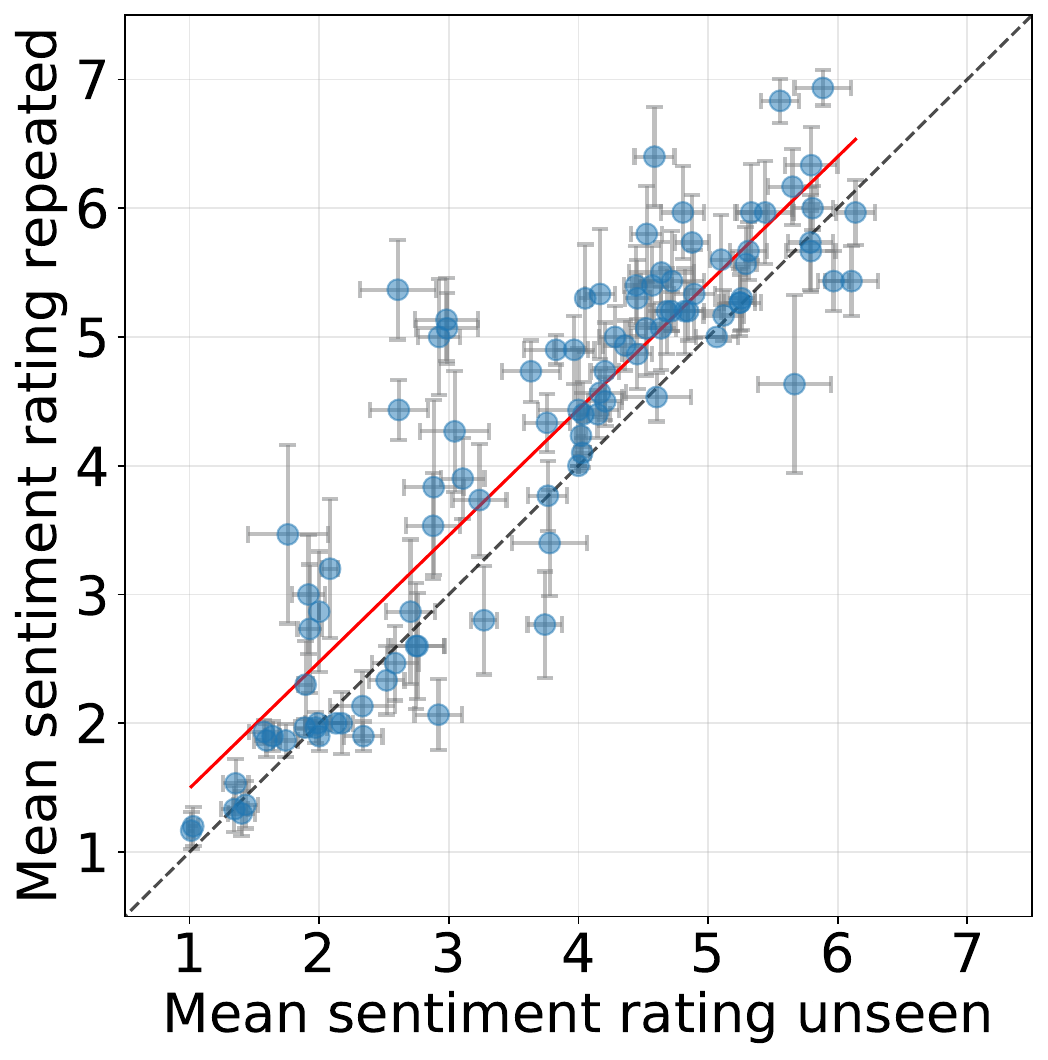}
        \caption{}
        
    \end{subfigure}
    \hfill
    \begin{subfigure}[b]{0.32\textwidth}
        \centering
        \includegraphics[width=\textwidth]{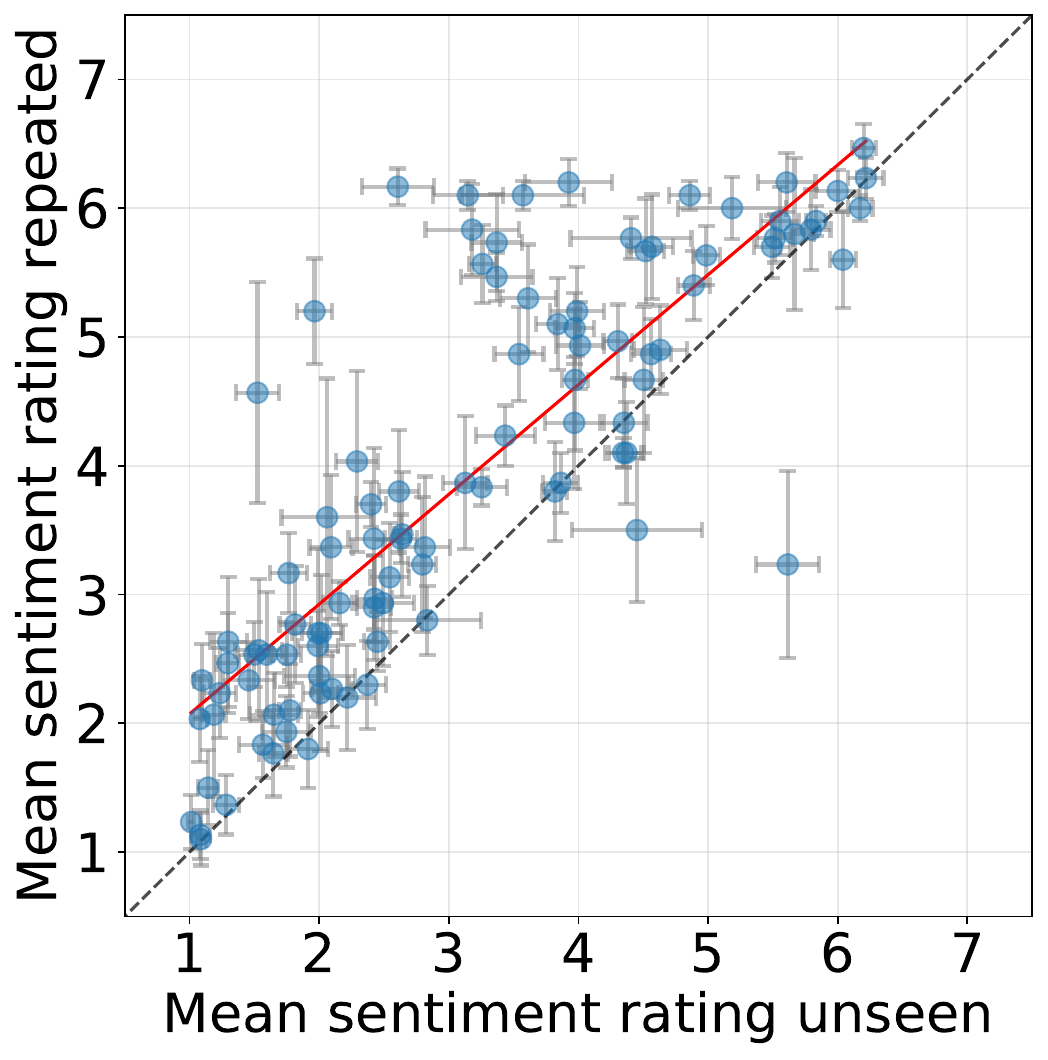}
        \caption{}
        
    \end{subfigure}
    \hfill
    \caption{Plots of the mean \textbf{sentiment} rating per statement when unseen and repeated in the simulation phase. (a) \texttt{Llama-3.1-8B-Instruct}, (b) \texttt{Qwen2.5-7B-Instruct}, (c) \texttt{Gemma-3-4b-it} and (d) \texttt{GPT-5-nano}. All models are set to temperature 0.1. Error bars are 95\% confidence intervals. The dashed black line is the identity function, the solid red line is the best linear fit.}
    
\end{figure}

\subsection{Details on the statistical model}

\subsubsection{About the LLMs used in the experiment}
We run a pilot of the experiment initially with 7 different models: \texttt{GPT-5-nano}, \texttt{Llama-3.1-8B-Instruct}, \texttt{Gemma-3-4b-it}, \texttt{Qwen2.5-7B-Instruct}, \texttt{Mistral-7B-Instruct-v0.2} \citep{jiang2023mistral7b}, \texttt{Mitral-7B-Instruct-v0.3} \citep{mistral2024instruct_v03} and \texttt{DeepSeek-R1-Distill-Llama-8B} \citep{deepseek2025r1}. However, Mistral-7B-Instruct-v0.3 did not abide by the tasks in the prompt in the rating phase and kept providing comments and actions on the news feed. Mitral-7B-Instruct-v0.2, followed better the instructions but the output format was still hard to automatically process. While \texttt{DeepSeek-R1-Distill-Llama-8B} kept giving very verbose answers detailing every step of it reasoning both of in the simulation and the rating phases, exceeding the maximum number of tokens per response. We decided to disregard \texttt{Mitral-7B-Instruct-v0.2}, \texttt{Mitral-7B-Instruct-v0.3} and \texttt{DeepSeek-R1-Distill-Llama-8B} models then from the main experiment. 

\subsubsection{Preliminary Model: temperature and replication effects}
\label{prelim_stat_model}

The first model was fit on the open-weight models only (\texttt{Gemma-3-4b-it}, \texttt{Llama-3.1-8B} and Qwen2.5-7B) to test whether temperature influenced the repetition effect and to investigate the variance introduced by the replication procedure (See Eq.~\ref{eq:prelimmodel}):
\begin{equation}
\begin{aligned}
  \texttt{rating} \sim\; & \texttt{repeated} \times \texttt{attribute} \\
                        + & \;\texttt{model} \times \texttt{attribute} \\
                        + & \;\texttt{temperature} \times \texttt{attribute} \\
                        + & \;\texttt{model} \mathbin{:} \texttt{temperature} \\ 
                        + & \; 1 | \texttt{statement\_id}/\texttt{simulation\_num}/\texttt{replication\_num},
\end{aligned}
  \label{eq:prelimmodel}
\end{equation}

where \texttt{repeated} is a binary variable indicating whether the rated statement was repeated (True) or unseen (False) during the simulation phase, \texttt{attribute} is the rating dimension (truth, importance, sentiment, interest), \texttt{model} identifies the LLM, and \texttt{temperature} is a categorical variable with levels 0.1 and 1.0. The term \texttt{model\,:\,temperature} was included as a fixed effect to control for baseline rating differences across model--temperature combinations. The reference levels were: \texttt{repeated} = False, \texttt{attribute} = truth, \texttt{temperature} = 0.1. \texttt{statement\_id}), \texttt{simulation\_num}) and \texttt{replication\_num} are included as nested random effects. The total sample comprised 287,996 observations. 

The results of this preliminary model are presented in Tab.~\ref{tab:fixef} and in Tab.~\ref{tab:ranef}. Tab.~\ref{tab:fixef} did not show any significant difference between high and low temperatures (Est. 0.023, $p-value = 0.12$). In Tab.~\ref{tab:ranef}, we estimate the random intercept at each level of the nested grouping structure. The trial-level standard deviation is negligibly small ($8.2 \times 10^{-5}$), indicating that virtually all within-replication variance is captured by the residual term. This finding indicates that model responses to identical prompts were effectively deterministic.

\begin{longtable}{lrrrrl}
\caption{Results from Eq.~\ref{eq:prelimmodel}: fixed-effect estimates, standard errors, degrees of freedom, $t$-statistics, and $p$-values. Reference levels are: \texttt{repeated = False}, \texttt{attribute = truth}, and \texttt{temperature = 0}. To keep term labels concise, model names are abbreviated: \textbf{M1} = \texttt{Llama-3.1-8B-Instruct}; \textbf{M2} = \texttt{Qwen2.5-7B-Instruct}. The baseline (reference) model is \texttt{Gemma-3-4b-it}.}
\label{tab:fixef} \\
\toprule
\textbf{Term} & \textbf{Estimate} & \textbf{SE}  & $\boldsymbol{t}$ & $\boldsymbol{p}$ \\
\midrule
\endfirsthead
\multicolumn{6}{c}{\tablename~\thetable{} \textit{(continued)}} \\
\toprule
\textbf{Term} & \textbf{Estimate} & \textbf{SE}  & $\boldsymbol{t}$ & $\boldsymbol{p}$ \\
\midrule
\endhead
\midrule \multicolumn{6}{r}{\textit{continued on next page}} \\
\endfoot
\bottomrule
\endlastfoot

% --- Main effects ---
\multicolumn{6}{l}{\textit{Main effects}} \\[2pt]
(Intercept)                                    &  4.498 & 0.1123  &  40.05 & $<$0.001 \\
repeated[True]                                   &  0.532 & 0.0175  &  30.32 & $<$0.001 \\
attribute[interest]                            &  0.187 & 0.0174  &  10.79 & $<$0.001 \\
attribute[sentiment]                           & $-$1.246 & 0.0173  & $-$71.92 & $<$0.001 \\
attribute[truth]                               & $-$0.616 & 0.0172  & $-$35.75 & $<$0.001 \\
model[M1]                                      & $-$0.486 & 0.0166  & $-$29.33 & $<$0.001 \\
model[M2]                                      & $-$1.072 & 0.0166  & $-$64.62 & $<$0.001 \\
temperature[1]                                 &  0.023 & 0.0149  &   1.56 & 0.120 \\[6pt]

% --- repeated × attribute ---
\multicolumn{6}{l}{\textit{repeated $\times$ attribute interactions}} \\[2pt]
repeated[True] $\times$ attribute[interest]      & $-$0.376 & 0.0232  & $-$16.24 & $<$0.001 \\
repeated[True] $\times$ attribute[sentiment]     & $-$0.290 & 0.0231  & $-$12.55 & $<$0.001 \\
repeated[True] $\times$ attribute[truth]         &  0.160 & 0.0231  &   6.94 & $<$0.001 \\[6pt]

% --- attribute × model ---
\multicolumn{6}{l}{\textit{attribute $\times$ model interactions}} \\[2pt]
attribute[interest]  $\times$ model[M1]        &  0.166 & 0.0211  &   7.88 & $<$0.001 \\
attribute[sentiment] $\times$ model[M1]        &  1.365 & 0.0211  &  64.67 & $<$0.001 \\
attribute[truth]     $\times$ model[M1]        &  1.071 & 0.0211  &  50.81 & $<$0.001 \\
attribute[interest]  $\times$ model[M2]        &  0.655 & 0.0211  &  31.01 & $<$0.001 \\
attribute[sentiment] $\times$ model[M2]        &  1.642 & 0.0210  &  78.20 & $<$0.001 \\
attribute[truth]     $\times$ model[M2]        &  0.679 & 0.0210  &  32.39 & $<$0.001 \\[6pt]

% --- attribute × temperature ---
\multicolumn{6}{l}{\textit{attribute $\times$ temperature interactions}} \\[2pt]
attribute[interest]  $\times$ temperature[1]   & $-$0.012 & 0.0172  &  $-$0.72 & 0.470 \\
attribute[sentiment] $\times$ temperature[1]   &  0.025 & 0.0172  &   1.45 & 0.147 \\
attribute[truth]     $\times$ temperature[1]   & $-$0.030 & 0.0171  &  $-$1.75 & 0.080 \\[6pt]

% --- model × temperature ---
\multicolumn{6}{l}{\textit{model $\times$ temperature interactions}} \\[2pt]
model[M1] $\times$ temperature[1]              &  0.084 & 0.0151  &   5.59 & $<$0.001 \\
model[M2] $\times$ temperature[1]              & $-$0.033 & 0.0148  &  $-$2.26 & 0.024 \\

\end{longtable}

\begin{table}[H]
\centering
\caption{Random Effects Standard Deviations}
\label{tab:ranef}
\begin{tabularx}{\textwidth}{l X r}
\toprule
Level & Grouping Formula & Std.\ Dev. \\
\midrule
Statement                  & \texttt{\textasciitilde1 | statement\_id}                                                           & 1.1131 \\
Simulation (in Statement)  & \texttt{\textasciitilde1 | simulation\_num \%in\% statement\_id}                                    & 0.9958 \\
Trial (in Simulation)      & \texttt{\textasciitilde1 | trial\_num \%in\% simulation\_num \%in\% statement\_id}                  & $8.215 \times 10^{-5}$ \\
Residual                   &                                                                                                     & 1.4706 \\
\bottomrule
\end{tabularx}
\end{table}

\subsubsection{Estimated Marginal Means}
\label{emmeans}
Estimated marginal means (EMMs) are the model-predicted values for each cell of the design, obtained by evaluating the fixed-effects component of the linear mixed-effects model at a specified combination of factor levels while setting the random effects to zero---that is, averaging over the
random-effects distribution rather than conditioning on any particular statement or simulation context. Formally, for a cell defined by a predictor vector $\mathbf{x}_0$, the EMM is:
\begin{equation}
  \hat{\mu}(\mathbf{x}_0) = \mathbf{x}_0^{\top} \hat{\boldsymbol{\beta}},
  \label{eq:emm}
\end{equation}
where $\hat{\boldsymbol{\beta}}$ is the vector of estimated fixed-effect coefficients. The associated standard error is $\mathrm{SE}(\hat{\mu}) = \sqrt{\mathbf{x}_0^{\top} \widehat{\mathrm{Var}}(\hat{\boldsymbol{\beta}})\, \mathbf{x}_0}$, where $\widehat{\mathrm{Var}}(\hat{\boldsymbol{\beta}})$ is the estimated variance-covariance matrix of the fixed effects. Pairwise contrasts between repeated conditions (repeated $-$ unseen) are computed as $\hat{\delta} = \mathbf{c}^{\top}\hat{\boldsymbol{\beta}}$, where $\mathbf{c}$ is a contrast vector assigning $+1$ to True-repeated terms and $-1$ to False-repeated (unseen) terms.

For each of the four models, the EMM for a cell defined by $(\texttt{repeated} = r,\; \texttt{attribute} = a,\; \texttt{model} = m)$ expands as follows, with \texttt{Gemma-3-4b-it} as the reference model and \texttt{truth} as the reference attribute. Reference-level indicator terms evaluate to zero and are omitted for brevity.

\paragraph{Gemma-3-4b-it.}
As the reference model, the EMM reduces to:
\begin{equation}
  \hat{\mu}_{r,\,a,\,\texttt{gemma}} =
    \hat{\beta}_0
    + \hat{\beta}_{r}
    + \hat{\beta}_{a}
    + \hat{\beta}_{r \times a},
  \label{eq:emm_gemma}
\end{equation}
where $\hat{\beta}_0$ is the intercept (False: repeated, truth: attribute, \texttt{Gemma-3-4b-it}: model), $\hat{\beta}_{r}$ is the main effect of repeated, $\hat{\beta}_{a}$ is the main effect of attribute, and $\hat{\beta}_{r \times a}$ is their interaction. For the reference cell (\texttt{repeated} = True, \texttt{attribute} = truth) all interaction and non-intercept attribute terms vanish, giving $\hat{\mu} = \hat{\beta}_0 + \hat{\beta}_{\texttt{repeatedTrue}}$.

\paragraph{GPT-5-nano.}
\begin{equation}
  \hat{\mu}_{r,\,a,\,\texttt{gpt}} =
    \hat{\beta}_0
    + \hat{\beta}_{r}
    + \hat{\beta}_{a}
    + \hat{\beta}_{\texttt{gpt}}
    + \hat{\beta}_{r \times a}
    + \hat{\beta}_{r \times \texttt{gpt}}
    + \hat{\beta}_{a \times \texttt{gpt}}
    + \hat{\beta}_{r \times a \times \texttt{gpt}}.
  \label{eq:emm_gpt}
\end{equation}

\paragraph{Llama-3.1-8B-Instruct.}
\begin{equation}
  \hat{\mu}_{r,\,a,\,\texttt{llama}} =
    \hat{\beta}_0
    + \hat{\beta}_{r}
    + \hat{\beta}_{a}
    + \hat{\beta}_{\texttt{llama}}
    + \hat{\beta}_{r \times a}
    + \hat{\beta}_{r \times \texttt{llama}}
    + \hat{\beta}_{a \times \texttt{llama}}
    + \hat{\beta}_{r \times a \times \texttt{llama}}.
  \label{eq:emm_llama}
\end{equation}

\paragraph{Qwen2.5-7B-Instruct.}
\begin{equation}
  \hat{\mu}_{r,\,a,\,\texttt{qwen}} =
    \hat{\beta}_0
    + \hat{\beta}_{r}
    + \hat{\beta}_{a}
    + \hat{\beta}_{\texttt{qwen}}
    + \hat{\beta}_{r \times a}
    + \hat{\beta}_{r \times \texttt{qwen}}
    + \hat{\beta}_{a \times \texttt{qwen}}
    + \hat{\beta}_{r \times a \times \texttt{qwen}}.
  \label{eq:emm_qwen}
\end{equation}

\noindent
The pairwise repetition contrast $\hat{\delta} = \hat{\mu}_{r=\texttt{repeated}} - \hat{\mu}_{r=\texttt{unseen}}$ for each attribute $\times$ model cell isolates the contribution of the repeated terms, with unseen terms cancelling out. For the reference cell (truth, Gemma), the contrast reduces to the single coefficient $\hat{\beta}_{\texttt{repeatedTrue}}$; for non-reference cells it additionally includes the corresponding two-way and three-way interaction terms involving \texttt{repeatedTrue}.

\paragraph{Opinion Statements}

\begin{table}[H]
\small
\centering
\caption{Pairwise Contrasts ($\hat{\delta}$ EMMs True~$-$~EMMs False) by Attribute and Model. Subset of data with 10 opinion statements. Significance: *** $p < .001$, ** $p < .01$, * $p < .05$, \textit{ns} $p \geq .05$.}
\vspace{-6pt}
\label{tab:emmeans1}
\footnotesize
\setlength{\tabcolsep}{5pt}
\renewcommand{\arraystretch}{1.15}
\begin{threeparttable}
\resizebox{\linewidth}{!}{%
\begin{tabular}{l rrl rrl rrl rrl}
\toprule
& \multicolumn{3}{c}{\textit{Gemma-3-4b-it}}
& \multicolumn{3}{c}{\textit{gpt-5-nano}}
& \multicolumn{3}{c}{\textit{Llama-3.1-8B}}
& \multicolumn{3}{c}{\textit{Qwen2.5-7B}} \\
\cmidrule(lr){2-4}\cmidrule(lr){5-7}\cmidrule(lr){8-10}\cmidrule(lr){11-13}
Attribute & $\hat{\delta}$ & $t$ & & $\hat{\delta}$ & $t$ & & $\hat{\delta}$ & $t$ & & $\hat{\delta}$ & $t$ & \\
\midrule
truth
  & $-0.102$ & $-1.16$             & \textit{ns}
  & $+0.090$ & $\phantom{-}0.71$   & \textit{ns}
  & $-0.249$ & $-2.87$             & **
  & $+0.502$ & $\phantom{-}5.77$   & *** \\
importance
  & $+0.003$ & $\phantom{-}0.04$   & \textit{ns}
  & $-0.209$ & $-1.60$             & \textit{ns}
  & $-0.493$ & $-5.63$             & ***
  & $+0.779$ & $\phantom{-}8.98$   & *** \\
interest
  & $-0.218$ & $-2.48$             & *
  & $-0.562$ & $-4.51$             & ***
  & $+0.176$ & $\phantom{-}2.00$   & *
  & $+0.378$ & $\phantom{-}4.29$   & *** \\
sentiment
  & $+0.728$ & $\phantom{-}8.33$   & ***
  & $-0.065$ & $-0.51$             & \textit{ns}
  & $-0.621$ & $-7.14$             & ***
  & $+0.374$ & $\phantom{-}4.15$   & *** \\
\bottomrule
\end{tabular}%
}
\end{threeparttable}
\end{table}

\subsubsection{Three-way interaction}

\paragraph{Ratings from temperatures 1 and 0.1}
\label{three_way_model_all_data}

In Tab.~\ref{tab:fixef_all}, we present the fixed effects estimates. In Tab.~\ref{tab:emmeans_all}, we present the Estimated Marginal Means. In Tab.~\ref{tab:delta_contrasts_all_data}, we present the pairwise contrasts between attribute-level effect sizes. All three tables refers to fitting Eq.~\ref{eq:3way_model} using ratings from both temperatures (except for \texttt{GPT-5-nano} only available with temperature 1). 

\begin{longtable}{p{0.45\textwidth}rrrl}
\caption{Results from Eq.~\ref{eq:3way_model}: fixed-effect estimates, standard errors, $t$-statistics, and $p$-values when we fit ratings from LLMs with both temperatures (1 and 0.1). Reference levels are: \texttt{repeated = False}, \texttt{attribute = truth}, and \texttt{LLM = Gemma-3-4b-it}. To keep term labels concise, model names are abbreviated:  \textbf{M0} = \texttt{GPT-5-nano}; \textbf{M1} = \texttt{Llama-3.1-8B-Instruct}; \textbf{M2} = \texttt{Qwen2.5-7B-Instruct}.}
\label{tab:fixef_all} \\
\toprule
\textbf{Term} & \textbf{Estimate} & \textbf{SE} & $\boldsymbol{t}$ & $\boldsymbol{p}$ \\
\midrule
\endfirsthead
\multicolumn{5}{c}{\tablename~\thetable{} \textit{(continued)}} \\[4pt]
\toprule
\textbf{Term} & \textbf{Estimate} & \textbf{SE} & $\boldsymbol{t}$ & $\boldsymbol{p}$ \\
\midrule
\endhead
\midrule
\multicolumn{5}{r}{\textit{continued on next page}} \\
\endfoot
\bottomrule
\endlastfoot

% ---- Main effects ----
\multicolumn{5}{l}{\textit{Main effects}} \\[3pt]
(Intercept)                              &  3.817 & 0.1001 &  38.13 & $<$0.001 \\
repeated[True]                           &  0.977 & 0.0277 &  35.22 & $<$0.001 \\
attribute[importance]                    &  0.675 & 0.0159 &  42.47 & $<$0.001 \\
attribute[interest]                      &  0.875 & 0.0158 &  55.42 & $<$0.001 \\
attribute[sentiment]                     & $-$0.620 & 0.0159 & $-$38.97 & $<$0.001 \\
model[M0]                                & $-$0.762 & 0.0235 & $-$32.47 & $<$0.001 \\
model[M1]                                &  0.782 & 0.0159 &  49.04 & $<$0.001 \\
model[M2]                                & $-$0.420 & 0.0157 & $-$26.80 & $<$0.001 \\[8pt]

% ---- repeated × attribute ----
\multicolumn{5}{l}{\textit{repeated $\times$ attribute interactions}} \\[3pt]
repeated[True] $\times$ attribute[importance]  & $-$0.370 & 0.0381 &  $-$9.70 & $<$0.001 \\
repeated[True] $\times$ attribute[interest]    & $-$0.893 & 0.0384 & $-$23.25 & $<$0.001 \\
repeated[True] $\times$ attribute[sentiment]   & $-$0.278 & 0.0383 &  $-$7.24 & $<$0.001 \\[8pt]

% ---- repeated × model ----
\multicolumn{5}{l}{\textit{repeated $\times$ model interactions}} \\[3pt]
repeated[True] $\times$ model[M0]              & $-$1.017 & 0.0488 & $-$20.86 & $<$0.001 \\
repeated[True] $\times$ model[M1]              & $-$0.880 & 0.0383 & $-$22.98 & $<$0.001 \\
repeated[True] $\times$ model[M2]              &  0.036 & 0.0382 &   0.93 & 0.352 \\[8pt]

% ---- attribute × model ----
\multicolumn{5}{l}{\textit{attribute $\times$ model interactions}} \\[3pt]
attribute[importance] $\times$ model[M0]       &  0.188 & 0.0323 &   5.81 & $<$0.001 \\
attribute[interest]   $\times$ model[M0]       &  0.635 & 0.0321 &  19.76 & $<$0.001 \\
attribute[sentiment]  $\times$ model[M0]       &  1.405 & 0.0322 &  43.63 & $<$0.001 \\
attribute[importance] $\times$ model[M1]       & $-$1.096 & 0.0226 & $-$48.56 & $<$0.001 \\
attribute[interest]   $\times$ model[M1]       & $-$1.021 & 0.0228 & $-$44.73 & $<$0.001 \\
attribute[sentiment]  $\times$ model[M1]       &  0.315 & 0.0226 &  13.94 & $<$0.001 \\
attribute[importance] $\times$ model[M2]       & $-$0.773 & 0.0224 & $-$34.49 & $<$0.001 \\
attribute[interest]   $\times$ model[M2]       & $-$0.086 & 0.0225 &  $-$3.83 & $<$0.001 \\
attribute[sentiment]  $\times$ model[M2]       &  1.014 & 0.0222 &  45.57 & $<$0.001 \\[8pt]

% ---- repeated × attribute × model ----
\multicolumn{5}{l}{\textit{repeated $\times$ attribute $\times$ model interactions}} \\[3pt]
repeated[True] $\times$ attribute[importance] $\times$ model[M0]   &  0.414 & 0.0686 &   6.03 & $<$0.001 \\
repeated[True] $\times$ attribute[interest]   $\times$ model[M0]   &  0.678 & 0.0686 &   9.87 & $<$0.001 \\
repeated[True] $\times$ attribute[sentiment]  $\times$ model[M0]   &  0.239 & 0.0684 &   3.49 & $<$0.001 \\
repeated[True] $\times$ attribute[importance] $\times$ model[M1]   &  0.124 & 0.0541 &   2.29 & 0.022 \\
repeated[True] $\times$ attribute[interest]   $\times$ model[M1]   &  0.689 & 0.0544 &  12.67 & $<$0.001 \\
repeated[True] $\times$ attribute[sentiment]  $\times$ model[M1]   & $-$0.141 & 0.0541 &  $-$2.60 & 0.009 \\
repeated[True] $\times$ attribute[importance] $\times$ model[M2]   &  0.542 & 0.0539 &  10.04 & $<$0.001 \\
repeated[True] $\times$ attribute[interest]   $\times$ model[M2]   &  0.415 & 0.0543 &   7.64 & $<$0.001 \\
repeated[True] $\times$ attribute[sentiment]  $\times$ model[M2]   & $-$0.354 & 0.0541 &  $-$6.54 & $<$0.001 \\

\end{longtable}

\begin{table}[ht]
\centering
\caption{Estimated Marginal Means by Target, Attribute \& Model when we fit ratings from LLMs with both temperatures (1 and 0.1). Marginal means for each \texttt{repeated} level within each attribute $\times$ model cell. $M$ = emmean; $LL$ / $UL$ = 95\% confidence interval bounds.}
\label{tab:emmeans_all}
\resizebox{\linewidth}{!}{
\begin{tabular}{llrrr rrr rrr rrr}
\toprule
& & \multicolumn{3}{c}{Gemma-3-4b-it} & \multicolumn{3}{c}{gpt-5-nano} & \multicolumn{3}{c}{Llama-3.1-8B} & \multicolumn{3}{c}{Qwen2.5-7B} \\
\cmidrule(lr){3-5}\cmidrule(lr){6-8}\cmidrule(lr){9-11}\cmidrule(lr){12-14}
Attribute & Repeated & $M$ & $LL$ & $UL$ & $M$ & $LL$ & $UL$ & $M$ & $LL$ & $UL$ & $M$ & $LL$ & $UL$ \\
\midrule
\multirow{2}{*}{truth}
  & False & 3.82 & 3.62 & 4.02 & 3.05 & 2.85 & 3.26 & 4.60 & 4.40 & 4.80 & 3.40 & 3.20 & 3.60 \\
  & True  & 4.79 & 4.59 & 5.00 & 3.01 & 2.80 & 3.22 & 4.70 & 4.49 & 4.90 & 4.41 & 4.21 & 4.61 \\
\addlinespace
\multirow{2}{*}{importance}
  & False & 4.49 & 4.29 & 4.69 & 3.92 & 3.72 & 4.12 & 4.18 & 3.98 & 4.38 & 3.30 & 3.10 & 3.50 \\
  & True  & 5.10 & 4.89 & 5.30 & 3.92 & 3.71 & 4.13 & 4.03 & 3.82 & 4.23 & 4.48 & 4.28 & 4.69 \\
\addlinespace
\multirow{2}{*}{interest}
  & False & 4.69 & 4.49 & 4.89 & 4.56 & 4.36 & 4.76 & 4.45 & 4.25 & 4.65 & 4.19 & 3.99 & 4.38 \\
  & True  & 4.77 & 4.57 & 4.98 & 4.31 & 4.10 & 4.52 & 4.34 & 4.14 & 4.55 & 4.72 & 4.52 & 4.92 \\
\addlinespace
\multirow{2}{*}{sentiment}
  & False & 3.20 & 3.00 & 3.40 & 3.84 & 3.64 & 4.04 & 4.29 & 4.09 & 4.49 & 3.79 & 3.59 & 3.99 \\
  & True  & 3.90 & 3.69 & 4.10 & 3.76 & 3.55 & 3.97 & 3.97 & 3.77 & 4.17 & 4.17 & 3.97 & 4.37 \\
\bottomrule
\end{tabular}
}
\end{table}

\begin{table}[ht]
\centering
\caption{Pairwise contrasts between attribute-level effect sizes ($\hat{\delta}$) within each model, when we fit ratings from LLMs with both temperatures (1 and 0.1). $\hat{\delta}$ = estimated repeated$-$unseen mean ratings contrast for a given attribute. Significance: *** $p < .001$, ** $p < .01$, * $p < .05$, \textit{ns} $p \geq .05$; $p$-values adjusted with Tukey's method.}
\label{tab:delta_contrasts_all_data}
\resizebox{\linewidth}{!}{
\begin{tabular}{l rrl rrl rrl rrl}
\toprule
& \multicolumn{3}{c}{\textit{Gemma-3-4b-it}}
& \multicolumn{3}{c}{\textit{gpt-5-nano}}
& \multicolumn{3}{c}{\textit{Llama-3.1-8B}}
& \multicolumn{3}{c}{\textit{Qwen2.5-7B}} \\
\cmidrule(lr){2-4}\cmidrule(lr){5-7}\cmidrule(lr){8-10}\cmidrule(lr){11-13}
Contrast & Est. & $t$ & & Est. & $t$ & & Est. & $t$ & & Est. & $t$ & \\
\midrule
$\hat{\delta}_\text{truth} - \hat{\delta}_\text{importance}$
  & $+$0.370 &  9.70 & *** & $-$0.044 & $-$0.77 & \textit{ns} & $+$0.246 &  6.43 & *** & $-$0.172 & $-$4.50 & *** \\
$\hat{\delta}_\text{truth} - \hat{\delta}_\text{interest}$
  & $+$0.893 & 23.25 & *** & $+$0.216 &  3.80 & *** & $+$0.204 &  5.32 & *** & $+$0.478 & 12.52 & *** \\
$\hat{\delta}_\text{truth} - \hat{\delta}_\text{sentiment}$
  & $+$0.278 &  7.24 & *** & $+$0.039 &  0.69 & \textit{ns} & $+$0.419 & 10.93 & *** & $+$0.632 & 16.50 & *** \\
$\hat{\delta}_\text{interest} - \hat{\delta}_\text{importance}$
  & $-$0.523 & $-$13.67 & *** & $-$0.260 & $-$4.55 & *** & $+$0.042 &  1.09 & \textit{ns} & $-$0.650 & $-$17.02 & *** \\
$\hat{\delta}_\text{sentiment} - \hat{\delta}_\text{importance}$
  & $+$0.092 &  2.42 & \textit{ns} & $-$0.083 & $-$1.45 & \textit{ns} & $-$0.173 & $-$4.51 & *** & $-$0.804 & $-$21.01 & *** \\
$\hat{\delta}_\text{sentiment} - \hat{\delta}_\text{interest}$
  & $+$0.616 & 16.00 & *** & $+$0.177 &  3.11 & **  & $-$0.215 & $-$5.58 & *** & $-$0.154 & $-$4.02 & *** \\
\bottomrule
\end{tabular}
}
\end{table}

\paragraph{Ratings from temperature 1}
\label{three_way_model_temp1}

In Tab.~\ref{tab:fixef_temp_1}, we present the fixed effects estimates. In Tab.~\ref{tab:emmeans_temp1}, we present the Estimated Marginal Means. In Tab.~\ref{tab:contrasts_temp1}, we present the pairwise contrasts of the EMMs (repeated - unseen) by attribute and by LLM. In Tab.~\ref{tab:delta_contrasts_temp1}, we present the pairwise contrasts between attribute-level effect sizes. All three tables refers to fitting Eq.~\ref{eq:3way_model} using only the ratings when the temperature is 1.

\begin{longtable}{p{0.45\textwidth}rrrl}
\caption{Results from Eq.~\ref{eq:3way_model}: fixed-effect estimates, standard errors, $t$-statistics, and $p$-values when we fit ratings only from LLMs with temperatures 1. Reference levels are: \texttt{repeated = False}, \texttt{attribute = truth}, and \texttt{LLM = \texttt{Gemma-3-4b-it}}. To keep term labels concise, model names are abbreviated:  \textbf{M0} = \texttt{GPT-5-nano}; \textbf{M1} = \texttt{Llama-3.1-8B-Instruct}; \textbf{M2} = \texttt{Qwen2.5-7B-Instruct}.}
\label{tab:fixef_temp_1} \\
\toprule
\textbf{Term} & \textbf{Estimate} & \textbf{SE} & $\boldsymbol{t}$ & $\boldsymbol{p}$ \\
\midrule
\endfirsthead
\multicolumn{5}{c}{\tablename~\thetable{} \textit{(continued)}} \\[4pt]
\toprule
\textbf{Term} & \textbf{Estimate} & \textbf{SE} & $\boldsymbol{t}$ & $\boldsymbol{p}$ \\
\midrule
\endhead
\midrule
\multicolumn{5}{r}{\textit{continued on next page}} \\
\endfoot
\bottomrule
\endlastfoot

% ---- Main effects ----
\multicolumn{5}{l}{\textit{Main effects}} \\[3pt]
(Intercept)
    &  3.777 & 0.0956 &  39.51 & $<$0.001 \\
repeated[True]
    &  0.963 & 0.0391 &  24.65 & $<$0.001 \\
attribute[importance]
    &  0.747 & 0.0223 &  33.47 & $<$0.001 \\
attribute[interest]
    &  0.917 & 0.0226 &  40.62 & $<$0.001 \\
attribute[sentiment]
    & $-$0.555 & 0.0226 & $-$24.58 & $<$0.001 \\
model[M0]
    & $-$0.702 & 0.0263 & $-$26.67 & $<$0.001 \\
model[M1]
    &  0.810 & 0.0222 &  36.51 & $<$0.001 \\
model[M2]
    & $-$0.407 & 0.0222 & $-$18.37 & $<$0.001 \\[8pt]

% ---- repeated × attribute ----
\multicolumn{5}{l}{\textit{repeated $\times$ attribute interactions}} \\[3pt]
repeated[True] $\times$ attribute[importance]
    & $-$0.387 & 0.0542 &  $-$7.13 & $<$0.001 \\
repeated[True] $\times$ attribute[interest]
    & $-$0.858 & 0.0549 & $-$15.63 & $<$0.001 \\
repeated[True] $\times$ attribute[sentiment]
    & $-$0.334 & 0.0545 &  $-$6.13 & $<$0.001 \\[8pt]

% ---- repeated × model ----
\multicolumn{5}{l}{\textit{repeated $\times$ model interactions}} \\[3pt]
repeated[True] $\times$ model[M0]
    & $-$1.032 & 0.0562 & $-$18.34 & $<$0.001 \\
repeated[True] $\times$ model[M1]
    & $-$0.843 & 0.0542 & $-$15.54 & $<$0.001 \\
repeated[True] $\times$ model[M2]
    &  0.075 & 0.0543 &   1.38 & 0.166 \\[8pt]

% ---- attribute × model ----
\multicolumn{5}{l}{\textit{attribute $\times$ model interactions}} \\[3pt]
attribute[importance] $\times$ model[M0]
    &  0.090 & 0.0361 &   2.50 & 0.012 \\
attribute[interest] $\times$ model[M0]
    &  0.572 & 0.0361 &  15.83 & $<$0.001 \\
attribute[sentiment] $\times$ model[M0]
    &  1.313 & 0.0361 &  36.39 & $<$0.001 \\
attribute[importance] $\times$ model[M1]
    & $-$1.149 & 0.0316 & $-$36.31 & $<$0.001 \\
attribute[interest] $\times$ model[M1]
    & $-$1.066 & 0.0321 & $-$33.17 & $<$0.001 \\
attribute[sentiment] $\times$ model[M1]
    &  0.353 & 0.0315 &  11.21 & $<$0.001 \\
attribute[importance] $\times$ model[M2]
    & $-$0.874 & 0.0317 & $-$27.57 & $<$0.001 \\
attribute[interest] $\times$ model[M2]
    & $-$0.187 & 0.0320 &  $-$5.85 & $<$0.001 \\
attribute[sentiment] $\times$ model[M2]
    &  0.924 & 0.0314 &  29.42 & $<$0.001 \\[8pt]

% ---- repeated × attribute × model ----
\multicolumn{5}{l}{\textit{repeated $\times$ attribute $\times$ model interactions}} \\[3pt]
repeated[True] $\times$ attribute[importance] $\times$ model[M0]
    &  0.471 & 0.0790 &   5.96 & $<$0.001 \\
repeated[True] $\times$ attribute[interest] $\times$ model[M0]
    &  0.651 & 0.0795 &   8.19 & $<$0.001 \\
repeated[True] $\times$ attribute[sentiment] $\times$ model[M0]
    &  0.334 & 0.0789 &   4.23 & $<$0.001 \\
repeated[True] $\times$ attribute[importance] $\times$ model[M1]
    &  0.250 & 0.0769 &   3.25 & 0.001 \\
repeated[True] $\times$ attribute[interest] $\times$ model[M1]
    &  0.694 & 0.0775 &   8.96 & $<$0.001 \\
repeated[True] $\times$ attribute[sentiment] $\times$ model[M1]
    & $-$0.090 & 0.0769 &  $-$1.17 & 0.242 \\
repeated[True] $\times$ attribute[importance] $\times$ model[M2]
    &  0.596 & 0.0768 &   7.77 & $<$0.001 \\
repeated[True] $\times$ attribute[interest] $\times$ model[M2]
    &  0.510 & 0.0773 &   6.60 & $<$0.001 \\
repeated[True] $\times$ attribute[sentiment] $\times$ model[M2]
    & $-$0.252 & 0.0770 &  $-$3.27 & 0.001 \\

\end{longtable}

\begin{table}[ht]
\centering
\caption{Estimated Marginal Means by Target, Attribute \& Model when we fit ratings from LLMs with temperatures 1. Marginal means for each \texttt{target} level within each attribute $\times$ model cell. $M$ = emmean; $LL$ / $UL$ = 95\% confidence interval lower and upper bounds.}
\label{tab:emmeans_temp1}
\small
\resizebox{\linewidth}{!}{
\begin{tabular}{llrrr rrr rrr rrr}
\toprule
& & \multicolumn{3}{c}{Gemma-3-4b-it} & \multicolumn{3}{c}{gpt-5-nano} & \multicolumn{3}{c}{Llama-3.1-8B} & \multicolumn{3}{c}{Qwen2.5-7B} \\
\cmidrule(lr){3-5}\cmidrule(lr){6-8}\cmidrule(lr){9-11}\cmidrule(lr){12-14}
Attribute & Repeated & $M$ & $LL$ & $UL$ & $M$ & $LL$ & $UL$ & $M$ & $LL$ & $UL$ & $M$ & $LL$ & $UL$ \\
\midrule
\multirow{2}{*}{truth}
  & False & 3.78 & 3.59 & 3.97 & 3.07 & 2.88 & 3.27 & 4.59 & 4.40 & 4.78 & 3.37 & 3.18 & 3.56 \\
  & True  & 4.74 & 4.54 & 4.94 & 3.01 & 2.81 & 3.20 & 4.71 & 4.51 & 4.91 & 4.41 & 4.21 & 4.61 \\
\addlinespace
\multirow{2}{*}{importance}
  & False & 4.52 & 4.33 & 4.71 & 3.91 & 3.72 & 4.10 & 4.19 & 4.00 & 4.38 & 3.24 & 3.05 & 3.43 \\
  & True  & 5.10 & 4.90 & 5.30 & 3.93 & 3.73 & 4.13 & 4.17 & 3.97 & 4.37 & 4.49 & 4.29 & 4.69 \\
\addlinespace
\multirow{2}{*}{interest}
  & False & 4.69 & 4.50 & 4.88 & 4.56 & 4.37 & 4.75 & 4.44 & 4.25 & 4.63 & 4.10 & 3.91 & 4.29 \\
  & True  & 4.80 & 4.60 & 5.00 & 4.29 & 4.09 & 4.49 & 4.39 & 4.19 & 4.59 & 4.79 & 4.59 & 4.99 \\
\addlinespace
\multirow{2}{*}{sentiment}
  & False & 3.22 & 3.03 & 3.41 & 3.83 & 3.64 & 4.02 & 4.38 & 4.20 & 4.57 & 3.74 & 3.55 & 3.93 \\
  & True  & 3.85 & 3.65 & 4.05 & 3.76 & 3.56 & 3.96 & 4.08 & 3.88 & 4.28 & 4.19 & 3.99 & 4.39 \\
\bottomrule
\end{tabular}
}
\end{table}
\begin{table}[ht]
\centering
\caption{Pairwise Contrasts ($\hat{\delta}$ EMMs repeated~$-$~EMMs unseen) by Attribute and Model when we fit ratings from LLMs with temperatures 1. Significance: *** $p < .001$, ** $p < .01$, * $p < .05$, \textit{ns} $p \geq .05$}
\label{tab:contrasts_temp1}
\resizebox{\linewidth}{!}{
\begin{tabular}{l rrl rrl rrl rrl}
\toprule
& \multicolumn{3}{c}{\textit{Gemma-3-4b-it}}
& \multicolumn{3}{c}{\textit{gpt-5-nano}}
& \multicolumn{3}{c}{\textit{Llama-3.1-8B}}
& \multicolumn{3}{c}{\textit{Qwen2.5-7B}} \\
\cmidrule(lr){2-4}\cmidrule(lr){5-7}\cmidrule(lr){8-10}\cmidrule(lr){11-13}
Attribute & $\hat{\delta}$ & $t$ & & $\hat{\delta}$ & $t$ & & $\hat{\delta}$ & $t$ & & $\hat{\delta}$ & $t$ & \\
\midrule
truth      & $+$0.963 &  24.65 & *** & $-$0.069 & $-$1.70 & \textit{ns}  & $+$0.120 &   3.09 & **  & $+$1.038 &  26.55 & *** \\
importance & $+$0.576 &  14.81 & *** & $+$0.015 &   0.37  & \textit{ns}  & $-$0.017 & $-$0.42 & \textit{ns}  & $+$1.248 &  32.02 & *** \\
interest   & $+$0.105 &   2.64 & **  & $-$0.276 & $-$6.83 & ***          & $-$0.044 & $-$1.12 & \textit{ns}  & $+$0.690 &  17.63 & *** \\
sentiment  & $+$0.629 &  16.09 & *** & $-$0.069 & $-$1.70 & \textit{ns}  & $-$0.304 & $-$7.75 & ***          & $+$0.452 &  11.53 & *** \\
\bottomrule
\end{tabular}
}
\end{table}

\begin{table}[ht]
\centering
\caption{Pairwise contrasts between attribute-level effect sizes ($\hat{\delta}$) within each model when we fit ratings only from LLMs with temperatures 1. $\hat{\delta}$ = estimated True$-$False contrast for a given attribute. Significance: *** $p < .001$, ** $p < .01$, * $p < .05$, \textit{ns} $p \geq .05$; $p$-values adjusted with Tukey's method.}
\label{tab:delta_contrasts_temp1}
\resizebox{\linewidth}{!}{
\begin{tabular}{l rrl rrl rrl rrl}
\toprule
& \multicolumn{3}{c}{\textit{Gemma-3-4b-it}}
& \multicolumn{3}{c}{\textit{gpt-5-nano}}
& \multicolumn{3}{c}{\textit{Llama-3.1-8B}}
& \multicolumn{3}{c}{\textit{Qwen2.5-7B}} \\
\cmidrule(lr){2-4}\cmidrule(lr){5-7}\cmidrule(lr){8-10}\cmidrule(lr){11-13}
Contrast & Est. & $t$ & & Est. & $t$ & & Est. & $t$ & & Est. & $t$ & \\
\midrule
$\hat{\delta}_\text{truth} - \hat{\delta}_\text{importance}$
  & $+$0.387 &   7.13 & *** & $-$0.084 & $-$1.46 & \textit{ns} & $+$0.137 &   2.51 & \textit{ns} & $-$0.210 & $-$3.86 & *** \\
$\hat{\delta}_\text{truth} - \hat{\delta}_\text{interest}$
  & $+$0.858 &  15.63 & *** & $+$0.207 &   3.61 & **           & $+$0.164 &   3.02 & *           & $+$0.348 &   6.41 & *** \\
$\hat{\delta}_\text{truth} - \hat{\delta}_\text{sentiment}$
  & $+$0.334 &   6.13 & *** & $\phantom{-}$0.000 & $\phantom{-}$0.00 & \textit{ns} & $+$0.424 &   7.79 & *** & $+$0.586 &  10.75 & *** \\
$\hat{\delta}_\text{interest} - \hat{\delta}_\text{importance}$
  & $-$0.471 & $-$8.65 & *** & $-$0.291 & $-$5.06 & *** & $-$0.028 & $-$0.50 & \textit{ns} & $-$0.557 & $-$10.25 & *** \\
$\hat{\delta}_\text{sentiment} - \hat{\delta}_\text{importance}$
  & $+$0.053 &   0.97 & \textit{ns} & $-$0.084 & $-$1.46 & \textit{ns} & $-$0.287 & $-$5.27 & *** & $-$0.795 & $-$14.61 & *** \\
$\hat{\delta}_\text{sentiment} - \hat{\delta}_\text{interest}$
  & $+$0.524 &   9.56 & *** & $+$0.207 &   3.62 & **           & $-$0.260 & $-$4.75 & *** & $-$0.238 & $-$4.36 & *** \\
\bottomrule
\end{tabular}
}
\end{table}

\subsubsection{Model Assumptions}
\label{model_assumptions}
\begin{figure}
    \centering
    \includegraphics[width=0.5\linewidth]{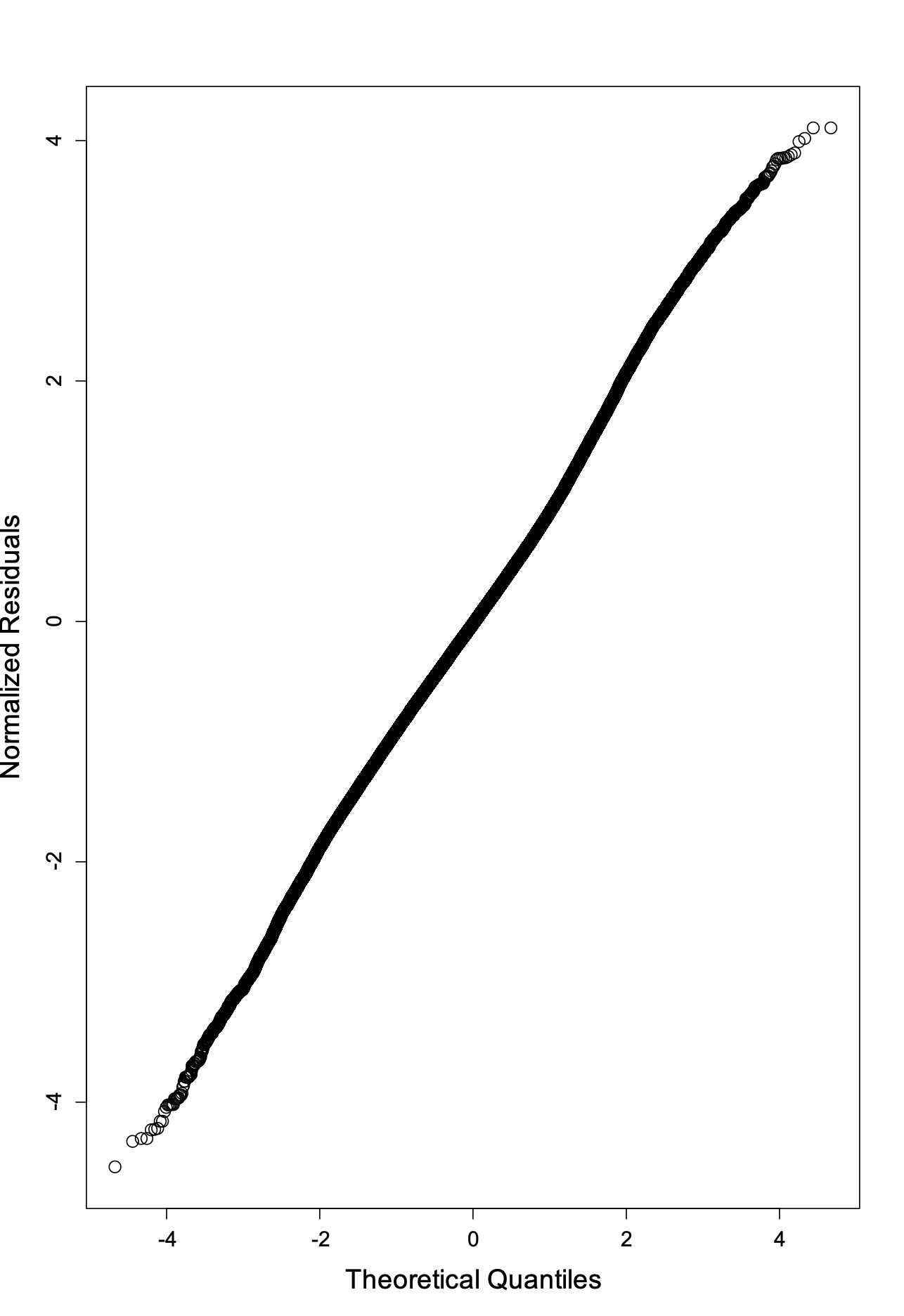}
    \caption{Quantile-quantile plot of the residuals versus fitted values with the LMEM.}
    \label{fig:qq_plot}
\end{figure}

\begin{figure}[h]

\begin{subfigure}[b]{0.49\textwidth}
        \centering
        \includegraphics[width=\textwidth]{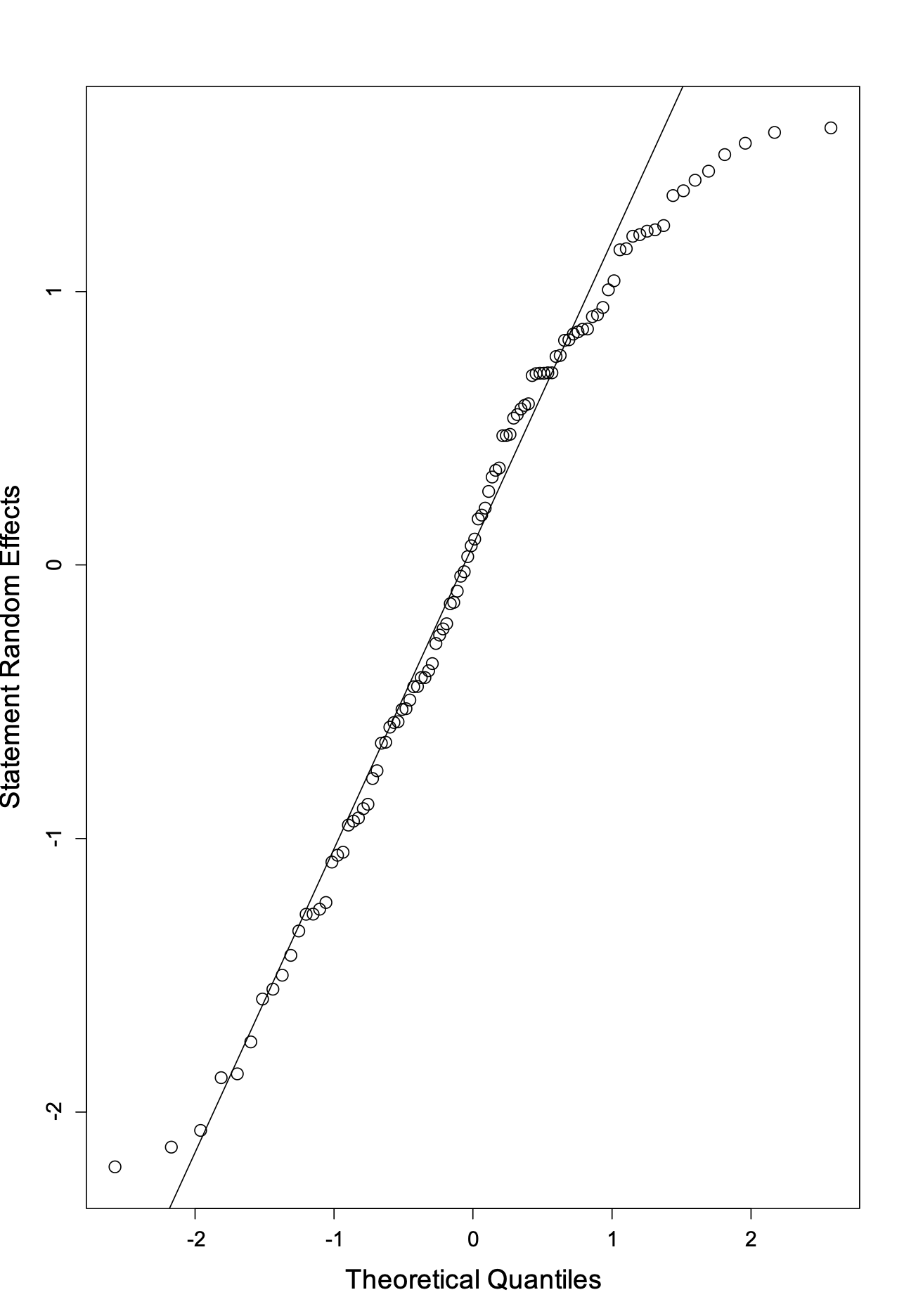}
        \caption{}
    \end{subfigure}
    \hfill
    \begin{subfigure}[b]{0.49\textwidth}
        \centering
        \includegraphics[width=\textwidth]{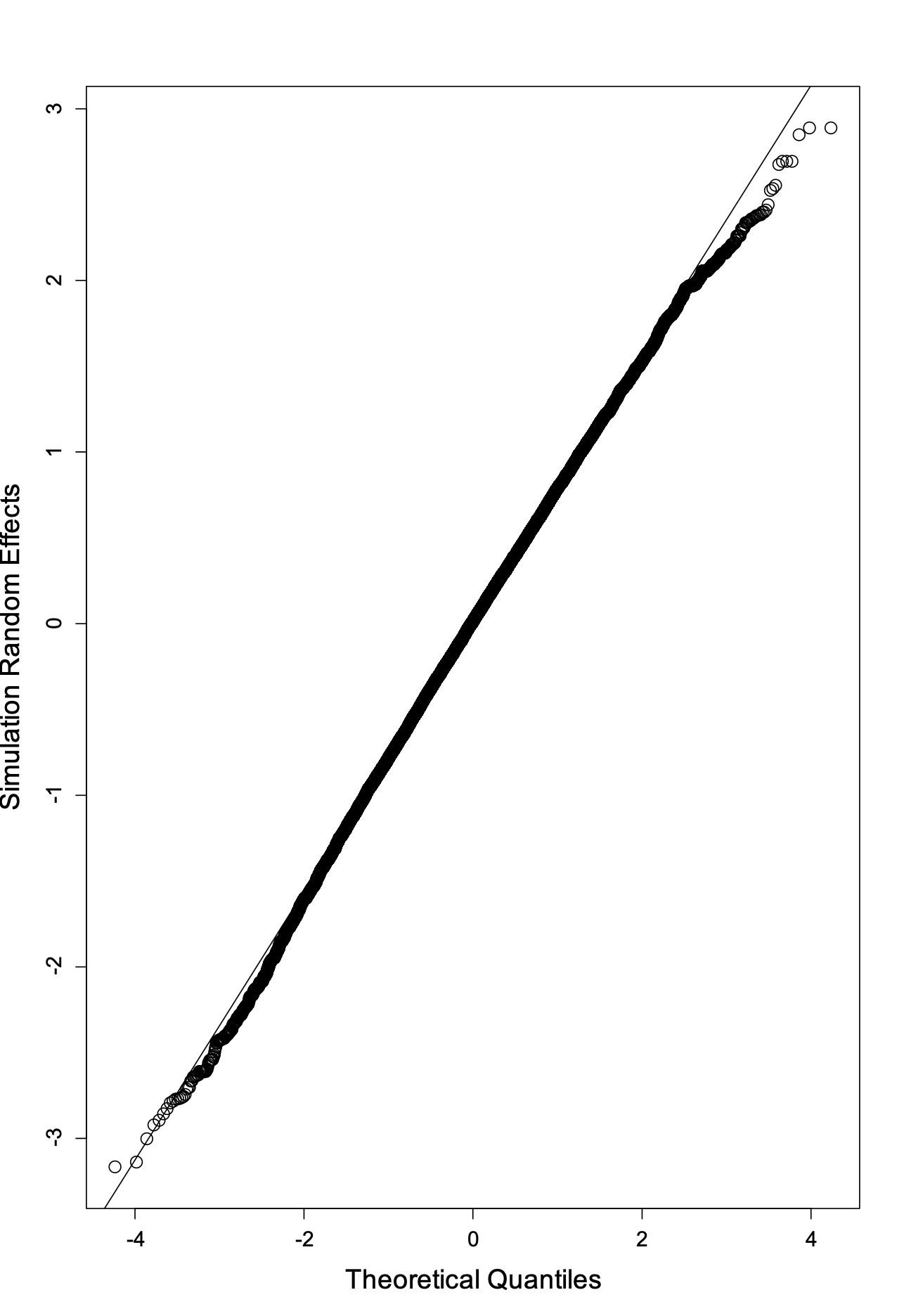}
        \caption{}
    \end{subfigure}
    \caption{Quantile-quantile plots of the estimated random intercepts at statement (a) and simulation run (b) levels.}
    \label{fig:qq_re}
\end{figure}

\begin{figure}
    \centering
    \includegraphics[width=\linewidth]{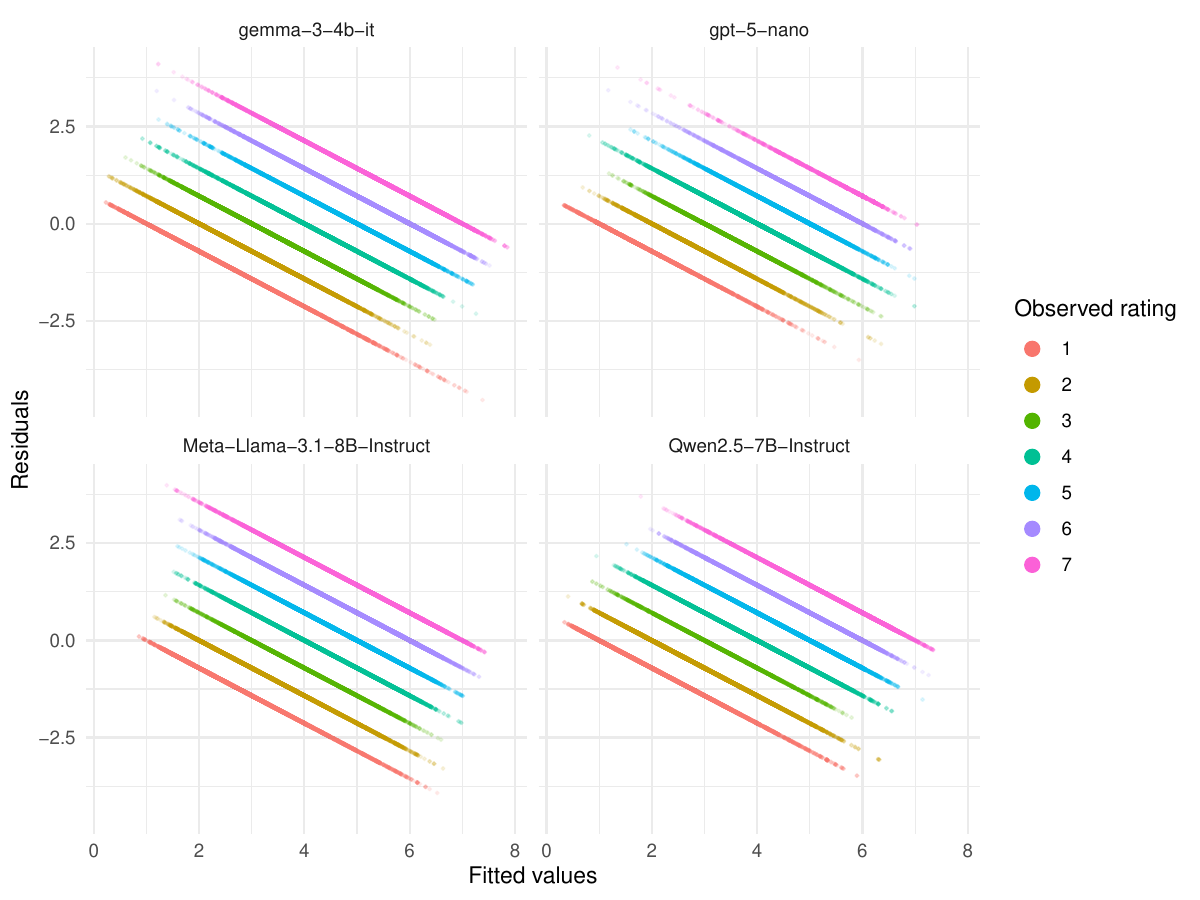}
    \caption{Standardized residuals against fitted values stratified by model.}
    \label{fig:homoscedasticity}
\end{figure}

We assessed the assumptions for the LMEM. We evaluated the \textit{Linearity} by visual inspection of residuals versus fitted values plots. It showed no systematic curvature. In Fig.~\ref{fig:qq_plot}, we assessed the \textit{Normality of residuals} via quantile-quantile plots of standardized within-group residuals. The points followed the theoretical diagonal closely across the full range, with minor deviations at the extremes attributable to the bounded, discrete nature of the Likert outcome. We assessed the \textit{Normality of random effects}. We show in Fig.~\ref{fig:qq_re} the quantile-quantile plots of the estimated random intercepts at both statement and simulation run levels. The simulation-level random effects were approximately normal, while statement-level random effects showed mild positive skew in the upper tail, reflecting a small number of statements with unusually elevated baseline ratings. This mild violation is not expected to materially affect the fixed-effect estimates given the large overall sample size. Finally, to assess the \textit{Homoscedasticity}, we plot, in Fig.~\ref{fig:homoscedasticity}, the standardized residuals against fitted values stratified by model. The residual spread was consistent across the range of fitted values (the 1-to-7 ratings) and did not differ systematically across models, indicating that the homoscedasticity assumption was adequately satisfied.

Ratings were treated as continuous throughout, consistent with standard practice for 7-point Likert scales in social and computational science \citep{norman2010}. We acknowledge this as a theoretical simplification; the discrete, bounded nature of the scale is reflected in the characteristic stripe pattern observed in residual plots, but did not produce evidence of systematic model misfit.


\subsection{LLM Disclosure Statement}
Claude (Anthropic) was used to assist with manuscript editing, double-checking the statistical details and in developing portions of the data analysis code. All research ideas, experimental design,  and scientific conclusions are exclusively those of the authors.

\end{document}